**Artificial entrepreneurial cognition: Locating and causally steering an opportunity recognition dial inside large language models (LLMs)**

**Christian Fisch[a], Angela Altmeier[a], Martin Obschonka[b*], Michal Kosinski[c], Pin Ni[a]**

[a] *Interdisciplinary Centre for Security, Reliability and Trust (SnT), University of Luxembourg, Luxembourg, 1855, Luxembourg*

[b] *Amsterdam Business School*, *University of Amsterdam, Amsterdam, 1018 TV, The Netherlands*

[c] *Stanford Graduate School of Business, Stanford University, Stanford, CA 94305, USA*

* Corresponding author. Amsterdam Business School, University of Amsterdam, Plantage Muidergracht 12, 1018 TV Amsterdam, The Netherlands. m.obschonka@uva.nl

*E-mails:* christian.fisch@uni.lu (C. Fisch), angela.altmeier@uni.lu (A. Altmeier), m.obschonka@uva.nl (M. Obschonka), michalk@stanford.edu (M. Kosinski), ni.pin@uni.lu (P. Ni)

**Artificial entrepreneurial cognition: Locating and causally steering an opportunity recognition dial inside large language models (LLMs)**

**Abstract:** Entrepreneurial cognition is a foundation of entrepreneurship research. Yet the growing involvement of large language models (LLMs) in entrepreneurial work extends the cognition question beyond human actors to systems whose internal representations remain largely unexplored. We introduce *artificial entrepreneurial cognition*, the functional organisation of entrepreneurship-relevant representations and computations inside artificial intelligence (AI) systems. We bring mechanistic interpretability into entrepreneurship research through representation engineering. Focusing on opportunity recognition (OR), we construct 636 matched OR-present and OR-absent scenario pairs and recover an OR direction in Llama 3.1 8B-Instruct. Rather than infer the construct from outputs, we intervene directly on this direction, steering the model up and down along what we call the *opportunity recognition dial*, and its opportunity judgments shift with it. To our knowledge, this is the first causal intervention on an internal representation of an entrepreneurship construct inside an LLM. Held-out tests, lexical and topical controls, behavioural ablation, and geometric comparisons show that the direction is recoverable, consequential, and distinct from the opportunity evaluation and exploitation directions, although steering it also shifts judgments about these neighbouring stages. Recovery, signed steering, and geometric separation hold across four additional LLMs spanning different scales and families. These results give the contested distinction between opportunity recognition and evaluation a concrete representational form inside AI systems. More broadly, they establish internal representations as a new object of entrepreneurship inquiry and show how entrepreneurship theory can guide their identification, causal manipulation, and interpretation.

**Data and code availability:** The analysis code and data used in this study are publicly available at https://osf.io/ks2ae/overview?view_only=be3522b774bf45888cd6080ae75c0684. Our detailed technical appendix spells out the implementation, parameter choices, and diagnostics.

## 1. Introduction

For decades, research on entrepreneurial cognition has sought to understand "how entrepreneurs think and 'why' they do some of the things they do" (Kuratko et al., 2021, p. 1683). This work has examined the knowledge structures and judgments that shape how entrepreneurs perceive, interpret, and respond to entrepreneurial situations (e.g., Baron, 2006; Grégoire et al., 2010; Mitchell et al., 2002). Recently, however, the cognition question has begun to extend beyond human actors.

Artificial intelligence (AI) systems, most notably large language models (LLMs), are increasingly used to screen venture ideas, evaluate pitches, generate business models, and support founders, potentially influencing which possibilities receive entrepreneurial attention and capital (e.g., Csaszar et al., 2024; Obschonka et al., 2025). In a survey of 1,051 founders who started businesses in 2025, 60% reported using AI to help launch them (Gusto, 2026). Yet entrepreneurship research currently lacks an established framework for examining how entrepreneurship-relevant distinctions are represented and used inside the AI systems now participating in entrepreneurial work. Addressing this gap connects entrepreneurship research to broader efforts to understand the internal workings of increasingly capable AI systems (e.g., Bereska and Gavves, 2024; Horvitz and West, 2026; Mueller et al., 2026).

To help open this new black box, we introduce *artificial entrepreneurial cognition*, defined as the functional organisation of entrepreneurship-relevant representations and computations inside AI systems, with LLMs as the focal instantiation examined here.[1] This definition retains entrepreneurial cognition's traditional focus on the internal structures that support entrepreneurship-relevant assessments and judgments (Mitchell et al., 2002), while shifting the object of inquiry from human knowledge structures to representations and computations inside

[1] We do not claim that an LLM thinks like an entrepreneur or possesses an entrepreneurial mind. Entrepreneurship-relevant distinctions may nonetheless have identifiable internal counterparts that contribute to the system's responses, and OR is one such distinction. The OR dial, investigated here, is therefore not artificial entrepreneurial cognition in its entirety, but the first empirical handle on one of its constituent structures.

AI systems. We use cognition in a functional, not phenomenal, sense. The term refers to what such systems compute and represent internally and does not imply consciousness, subjective understanding, intention, or entrepreneurial agency (cf. Shanahan et al., 2023). This creates a bounded theoretical problem for entrepreneurship research: if LLMs now participate in entrepreneurial work, do entrepreneurship-relevant constructs have identifiable and causally consequential counterparts inside them?

Existing entrepreneurship research cannot answer this question because it has largely examined LLMs from the outside. One stream treats AI systems, including LLMs, as tools that extend human entrepreneurial capacity (e.g., Chalmers et al., 2021; Townsend and Hunt, 2019). A second, more recent stream treats them as simulators that reproduce entrepreneurial profiles when prompted, or as instruments for eliciting entrepreneurial judgments and elaborating theory (e.g., Dimov, 2026; Obschonka and Fisch, 2025). Both streams have advanced the field, but both characterise what LLMs produce rather than what they represent internally. The result is a construct-attribution problem. An LLM may produce a plausible opportunity judgment because an internal structure related to opportunity recognition (OR) contributes to its processing, or because it reproduces surface regularities associated with entrepreneurial language. Outputs alone cannot distinguish these explanations. When an LLM evaluates venture pitches, for example, apparent accuracy may reflect exposure to a venture's digital traces, so a seemingly valid opportunity judgment does not have to rest on the construct it appears to express (e.g., Kleinert and Urbig, 2026).

This problem matters in practice because founders and investors may act on LLM judgments without being able to inspect why a system treats one idea as promising and another as unremarkable and therefore struggle to distinguish genuine insight from an unsupported but persuasive assessment. The same opacity constrains entrepreneurship research: the field has asked "how do we deal with the often-cited black-box nature of AI, which can make it difficult for humans (or machines) to evaluate the quality of its outputs and processes?” (Obschonka et al., 2025, p. 632). Without access to the internal basis of these judgments, artificial entrepreneurial cognition remains behaviourally visible but theoretically ungrounded.

Yet this construct-attribution problem is more tractable in an open-weight LLM[2] than in human cognition. Entrepreneurship research has developed several ways to gain closer access to human cognition, including the study of knowledge structures and verbal protocols designed to reconstruct cognitive processes (e.g., Grégoire et al., 2010; Mitchell et al., 2002). These methods can elicit or infer cognitive structures and processes from participants’ reports, choices, and task performance. They do not, however, ordinarily allow researchers to locate a construct-specific internal representation, vary its activation at a chosen strength, remove it, and observe the resulting change within the same ongoing computation. Neuroimaging can identify neural activity associated with entrepreneurial cognition (e.g., Ooms et al., 2024), but it likewise does not permit this kind of direct, graded, and reversible intervention. In an open-weight LLM, by contrast, internal states can be recorded and perturbed while the model performs a task. Figure 1 captures this shift in empirical access, which opens a new route for entrepreneurship research: theory-defined internal structures can be located, manipulated, and tested causally, where earlier they could only be inferred from their external manifestations.

*- Please insert Figure 1 about here -*

We use this access to examine OR as a foundational case of artificial entrepreneurial cognition. OR sits at the core of the individual–opportunity nexus (e.g., Eckhardt and Shane, 2003; Shane and Venkataraman, 2000) and has long resisted direct empirical access as an internal cognitive process (e.g., Dimov, 2011). It also presents a demanding construct-validation problem: recognition is hard to separate from evaluation and from exploitation, and favourability is often bundled into the opportunity construct itself (e.g., Davidsson, 2015; McMullen and Shepherd, 2006). This contested boundary makes OR a theoretically diagnostic case that allows us to test whether an entrepreneurship-relevant internal structure can be located and whether it is distinguishable from the evaluative and action-oriented judgments surrounding it.

[2] Open-weight LLMs make their learned parameters available for download, enabling local deployment and direct access to internal activations. They are a widely used and powerful segment of contemporary LLMs, not a niche class of research models. They include prominent model families such as Llama, Qwen, Gemma, Mistral, and DeepSeek. Open-weight models are not necessarily fully open source because the underlying training data and code may remain unavailable (Longpre et al., 2025).

Accordingly, we ask whether an OR-related internal representation can be located and causally manipulated in an LLM, and how this representation is organised relative to evaluation (in the form of feasibility and desirability) and exploitation. We examine this question within a bounded, discovery-style design in which a potentially value-generating structure is present in the described situation and the model provides a third-person judgment. We therefore do not adjudicate the broader discovery–creation debate (e.g., Alvarez and Barney, 2007; Kirzner, 1997), claim that the model recognises opportunities in the human cognitive sense, or attribute entrepreneurial agency to it.

To answer this question, we draw on the linear representation hypothesis from mechanistic interpretability (MI), which holds that models frequently encode human-interpretable concepts as directions in their high-dimensional activation spaces (e.g., Elhage et al., 2022; Park et al., 2024). Indeed, extracting and characterising general concept representations in large-scale AI models has been described as a "fundamental open problem" (Beaglehole et al., 2026, p. 787). Representation engineering builds on this hypothesis by providing methods for locating such directions and shifting activations along them during inference (for a survey, see Wehner et al., 2025). Using 636 matched OR-present and OR-absent scenario pairs, we identify an OR direction in the open-weight model Llama 3.1 8B-Instruct and test it in three steps. First, we examine whether the direction can be recovered across varied scenarios without relying on obvious lexical cues. Second, we test its causal operativity by steering it up and down and ablating it, an intervention that removes the direction from the model's internal activations.

Third, we examine its representational integration by comparing its geometry with the feasibility, desirability, and exploitation directions and by testing whether intervention on the OR direction shifts judgments concerning these adjacent constructs. Because the direction can be located and adjusted at a chosen strength, we call it the *opportunity recognition dial*.[3] To our knowledge, this is the first causal intervention on an internal representation of an entrepreneurship construct inside an LLM.

We report three principal findings. First, a stable OR direction can be recovered from the model's internal states across held-out scenarios and survives tests against lexical, topical, and estimation artefacts. Second, the direction is causally operative. Turning the dial upward *and* downward shifts opportunity judgments in the predicted directions, while ablating it weakens the OR readout. Random-direction and off-construct comparisons reduce concerns that these effects are generic consequences of perturbing the model. Third, the OR direction is geometrically distinguishable from opportunity feasibility, desirability, and exploitation, yet intervention on it shifts judgments concerning all three. Recognition is therefore distinct at the level of representational structure but connected to evaluation and action at the level of expressed judgment. Importantly, the central empirical signatures supporting these findings extend beyond our primary model and appear across four additional LLMs spanning different model scales and families. The OR component of artificial entrepreneurial cognition therefore has a reproducible representational counterpart across the LLM systems we examine, which strengthens the standing of artificial entrepreneurial cognition as an important object of entrepreneurship research.

[3] The term *dial* refers to the particular reproducible direction identified through our contrastive design and manipulated in the analyses that follow. We do not claim that it is the model's unique, complete, or compositionally pure representation of OR. OR-related information may be distributed across multiple directions, features, or a higher-dimensional subspace. The recovered direction provides one empirically identifiable and causally testable handle on that broader representational structure.

Our study makes three contributions. First, we develop a structural theory of artificial entrepreneurial cognition in LLMs and provide its first empirical test. In doing so, we extend research on entrepreneurial cognition (e.g., Baron, 2006; Grégoire et al., 2010; Mitchell et al., 2002) to AI systems and add an internal, representational perspective to the emerging work on AI in entrepreneurship (e.g., Chalmers et al., 2021; Lévesque et al., 2022; Obschonka et al., 2025; Shepherd and Majchrzak, 2022). The theory specifies that an entrepreneurship construct has a meaningful counterpart inside an AI system when a construct-related structure is recoverable across varied inputs, causally operative under intervention, and systematically organised relative to theoretically adjacent constructs. Importantly, entrepreneurship theory is not merely the application context for an AI method. Representation engineering provides access to the model, but entrepreneurship theory determines what should count as OR and how an OR-related structure should differ from evaluation, exploitation, commerciality, affect, and other adjacent constructs.

Second, we contribute to research on entrepreneurial opportunities and their recognition (e.g., Ardichvili et al., 2003; Shane and Venkataraman, 2000; Short et al., 2010) by making the recognition–evaluation boundary experimentally tractable in an AI system. In the model we study, recognition is geometrically distinct from feasibility, desirability, and exploitation, yet causally coupled to all three. This pattern gives the distinction between recognition and evaluation (e.g., McMullen and Shepherd, 2006; Wood and McKelvie, 2015) a concrete representational form while also reproducing, at the level of model output, the bundling that has concerned opportunity scholars (e.g., Davidsson, 2015; Hansen et al., 2011). A distinction the field has debated conceptually (e.g., Dimov, 2011; Ramoglou and Tsang, 2016) thereby becomes experimentally accessible in a system where it can be located, separated, and perturbed. We report this as a property of the AI system and a demonstration of how entrepreneurship theory can be used to interrogate internal representations in LLMs, not as evidence about how human entrepreneurs recognise and evaluate opportunities.

Third, we introduce representation engineering as a promising empirical approach for entrepreneurship research. Representation engineering is a top-down approach within MI, a programme in AI that seeks to reverse-engineer model computations into human-understandable components (Bereska and Gavves, 2024). Whereas bottom-up MI work often examines individual neurons, features, or circuits, representation engineering operates at the level of directions in activation space that can be read, compared, and steered during inference (e.g., Zou et al., 2023). Its appeal for entrepreneurship research lies in moving the point of causal intervention inside the system under study. Whether a construct causes or merely covaries with an outcome remains difficult to establish where endogeneity is pervasive and experiments are costly and confined to external conditions (e.g., Anderson et al., 2022; Åstebro and Hoos, 2021; Tzabbar, 2026). Representation engineering offers a complementary standard of evidence. Researchers can locate a construct-related representation inside the model, increase, decrease, or ablate it, and observe the resulting change in model output (Figure 1). We embed this approach in a validation logic supplied by entrepreneurship theory. The combination offers a reusable template for studying other dimensions of artificial entrepreneurial cognition.

## 2. Theoretical framework: A structural theory of artificial entrepreneurial cognition

### 2.1 From observable AI behaviour to internal structure

Research on AI in entrepreneurship has largely examined AI systems through their observable behaviour. At least two perspectives can be distinguished: the tool perspective treats AI systems as instruments that augment human entrepreneurial capacity and support venture creation, judgment, and action (e.g., Chalmers et al., 2021; Shepherd and Majchrzak, 2022; Townsend and Hunt, 2019). The simulator perspective, more recent and centred on generative AI and LLMs, examines the judgments, profiles, and strategies that these models produce under different prompts and experimental conditions, from generating and evaluating venture strategies to supporting opportunity search and assessing early-stage ventures (e.g., Csaszar et al., 2024; Kleinert and Urbig, 2026; Obschonka and Fisch, 2025). These perspectives are neither exhaustive nor mutually exclusive, but both focus primarily on observable outputs and set aside the internal structures that contribute to them.

Moving from observable outputs to internal structure raises a different evidentiary question. A model may produce an output that corresponds to a theoretical construct without relying on an internal structure specific to that construct, so output correspondence alone cannot establish an internal counterpart. As Figure 1 illustrates, open-weight LLMs provide an additional route of empirical access: candidate internal structures can be located and directly manipulated. Accordingly, a construct-related internal structure must be recoverable, causally operative, and systematically organised relative to theoretically adjacent constructs. The following sections explain the methodological basis for examining such structures and develop these requirements into a structural theory of artificial entrepreneurial cognition.

### 2.2 Mechanistic interpretability, representation engineering, and the linear representation hypothesis

MI is a prominent and rapidly advancing research programme that aims to make the internal computations of neural networks scientifically tractable (e.g., Bereska and Gavves, 2024; Mueller et al., 2026).[4] A recent study in *Science* illustrates the reach of the broader effort to understand and manipulate internal model representations by extracting representations for hundreds of concepts and using them to steer multiple model types (Beaglehole et al., 2026).

[4] Bereska and Gavves (2024) convey the importance of MI by analogy to psychology's major turn from behaviourism to cognitive neuroscience: lacking tools for introspection, early psychology treated the mind as a black box and studied only observable behaviour, until internal processes became tractable.

Whereas behavioural research treats a model as an opaque input–output system, MI seeks to identify internal features, representations, and computational pathways and, where possible, to test their causal contribution to model behaviour (e.g., Arditi et al., 2024; Lindsey et al., 2025; Templeton et al., 2024). MI includes bottom-up and top-down approaches. Bottom-up work decomposes networks into fine-grained features or circuits, for example through dictionary learning or circuit tracing (e.g., Bricken et al., 2023; Lindsey et al., 2025; Templeton et al., 2024). Top-down work begins with a high-level concept and examines how it is represented and used within the model without first reconstructing the underlying circuitry. Our study follows this second route through representation engineering, which provides methods for locating and intervening on concept-related patterns in a model's internal activations (Zou et al., 2023).

Representation engineering builds on the linear representation hypothesis, which proposes that neural networks often encode high-level concepts as directions in high-dimensional activation space (Elhage et al., 2022; Park et al., 2024). Activations are the internal numerical states generated as a model processes an input, and a direction captures a pattern of variation across many activation dimensions, in contrast to the activity of a single neuron or isolated feature. Candidate directions can be estimated by comparing activations generated by inputs that differ systematically on a target concept. Once identified, a direction can be tested through probing, compared geometrically with other directions, and manipulated during inference (Bereska and Gavves, 2024; Zou et al., 2023). Prior research has recovered directions associated with truth and falsehood, sentiment, refusal, and persona-related tendencies (Arditi et al., 2024; Chen et al., 2025; Marks and Tegmark, 2024; Tigges et al., 2024), and steering or ablating such directions can produce theoretically predicted changes in model behaviour (Rimsky et al., 2024; Turner et al., 2023).

This approach draws an important evidentiary distinction. Information may be decodable without being functionally involved (e.g., Belinkov, 2022), and part of that decodability can come from the probe itself (e.g., Hewitt and Liang, 2019). Probe accuracy therefore establishes information availability, not causal operativity. The latter requires intervention, such as adding, subtracting, or ablating a candidate direction and examining whether model responses change as predicted. Appropriate controls remain necessary because arbitrary internal perturbations may also affect behaviour. For entrepreneurship research, representation engineering provides internal access, while entrepreneurship theory supplies the construct definition, boundaries, and expected relationships needed to interpret what is found. Jointly, they make it possible to test whether a theory-defined entrepreneurship-relevant distinction has a recoverable, causally operative, and systematically organised internal counterpart. This provides the methodological basis for the structural theory developed next.

### 2.3 A structural theory of artificial entrepreneurial cognition

We propose a structural theory of artificial entrepreneurial cognition. The theory holds that entrepreneurship-relevant distinctions can have identifiable internal counterparts in AI systems. These counterparts concern the organisation of internal representations and computations that contribute to an AI system's judgments. The theory is structural rather than developmental, which means that it specifies what such counterparts consist of, how they are organised, and what evidence establishes them. However, it does not explain how a particular structure arose during training.

The theory extends work that treats LLMs as legitimate objects of construct-based inquiry (e.g., Cao and Kosinski, 2024; Hagendorff et al., 2023; Kosinski, 2024; Pellert et al., 2024). While the general proposition that concepts can be represented as directions in activation space is established in MI research, what is new is the claim that entrepreneurship constructs can have artificial, model-internal counterparts, together with an account of what this claim entails and how it can be tested. Because entrepreneurship constructs are theoretically differentiated yet related, their artificial counterparts should be distinguishable but systematically connected.

The theory concerns AI systems, specifically LLMs. It does not propose that LLMs reproduce the cognitive architecture of human entrepreneurs, nor does it imply consciousness, subjective experience, intention, or entrepreneurial agency. We use cognition in a functional sense: the psychological vocabulary carries no attribution of mental states (cf. Pellert et al., 2024; Shanahan et al., 2023; Shiffrin and Mitchell, 2023), and the claim concerns the distinctions a system represents and the computational structures that contribute to its responses (Park et al., 2024; Zou et al., 2023). The theory therefore does not depend on whether LLMs possess human-like cognition. Structured conceptual representations can emerge from language prediction alone and align with human behavioural judgments (Xu et al., 2025), but such convergence does not establish a shared cognitive architecture (Li et al., 2026) or justify treating LLMs as surrogates for human participants. The theory also does not assume that a recovered structure was produced by exposure to entrepreneurial discourse. Whether such a structure emerged through pretraining, instruction tuning, post-training, or other learning processes remains a separate empirical question.

We use representational structure to denote a reproducible direction in activation space that provides an empirical handle on a theoretically specified construct. A recovered direction need not be the model's only or complete representation of that construct. The claim is deliberately narrow: construct-related information may be distributed across several directions, features, or a higher-dimensional subspace.

Consistent with standards of construct clarity, the identity of a direction cannot be inferred from its label alone (Suddaby, 2010). The theory therefore specifies three properties that must be established empirically. The first is representational existence. A candidate construct-related structure has representational existence when it can be recovered across varied inputs and cannot be explained by obvious surface cues. The second is causal operativity. A construct-related direction is causally operative when directly manipulating it changes the system's relevant responses in a directional and theoretically interpretable manner. The third is representational integration. A direction is representationally integrated when it occupies a theoretically meaningful position relative to adjacent constructs, involving both differentiation and connection. Two directions can be geometrically distinguishable while intervention on one changes responses concerning the other. Together, these properties specify when the evidence supports a theoretically meaningful internal counterpart of an entrepreneurship construct. They distinguish such a counterpart from a surface correlation, an inert decodable signal, or an indiscriminate direction that alters many outputs simultaneously.

### 2.4 Opportunity recognition as the focal theoretical case

We examine OR as the focal test of the structural theory of artificial entrepreneurial cognition. OR is especially suitable because it is central to entrepreneurship research while remaining difficult to separate from adjacent constructs. Cognitive accounts describe OR as the perception of a meaningful pattern (Baron, 2006), or the structural alignment of a technology with market conditions (Grégoire et al., 2010). Process accounts place recognition near the beginning of an opportunity process that continues into evaluation, development, and exploitation (Ardichvili et al., 2003).

At the same time, the conceptual boundaries of OR remain contested. McMullen and Shepherd (2006) distinguish third-person recognition that an opportunity exists from the subsequent first-person evaluation of whether it is feasible and desirable for a particular actor. Davidsson (2015) adds that favourability is frequently bundled into the opportunity construct, making recognition difficult to separate from evaluation. These perspectives set a high bar for construct validation. If an OR direction is recovered inside an LLM, it must be distinguishable from evaluation (i.e., feasibility and desirability) and exploitation.

Our operationalisation follows a bounded, discovery-style account. We define OR as *the cognitive act of noticing a generalisable structure in a scenario that could serve as the basis for a potential value-generating activity, independent of evaluation or action*. The scenarios present the relevant structure as existing in the described situation and ask the model to render a third-person judgment. This places the study within a discovery-style framing (Alvarez and Barney, 2007) and at the opportunity-as-happening level of analysis (Dimov, 2011). It does not require that the opportunity is retrospectively confirmed through a successful outcome (Ramoglou and Tsang, 2016), nor does it test whether the model personally evaluates or acts on the opportunity.

The following subsections apply the structural theory to OR by translating its three properties into testable propositions that form a cumulative logic. We first ask whether an OR direction can be recovered, then whether it is causally operative, and finally how it is organised relative to adjacent constructs. Figure 2 summarises the structural theory and its application to OR.

*- Please insert Figure 2 about here -*

#### 2.4.1 Representational existence

The first theoretical requirement concerns whether OR-related information has a stable internal counterpart. A model could produce an opportunity-related response, or a probe could recover the OR label, by relying on surface words such as "market," "customer," "business," or "opportunity." Shortcut learning of this kind is common in machine-learning systems (Geirhos et al., 2020).

Entrepreneurship theory provides a stronger criterion. Baron (2006) describes recognition as pattern perception, and Grégoire et al. (2010) show that recognition depends on the alignment of structural relations between a new means and a market need, with surface similarity contributing little. An OR direction should therefore be recoverable when direct lexical cues are removed and should generalise across changes in wording, topic, and scenario form.

Representational existence also requires safeguards against misidentifying the recovered direction. This is a discriminant-validity requirement familiar from construct validation more generally (Campbell and Fiske, 1959). Internal-state analyses and appropriate controls are therefore needed to determine whether the recovered information reflects the intended OR contrast or an incidental correlate (e.g., Alain and Bengio, 2017; Belinkov, 2022; Hewitt and Liang, 2019).

> ***Proposition 1 (Representational existence).*** *An entrepreneurship construct has representational existence inside an LLM when it is recoverable as a stable, linear direction that generalises beyond the surface wording of individual inputs. For OR, the OR direction should separate OR-present from OR-absent scenarios across held-out and altered inputs.*

### 2.4.2 Causal operativity

Recovering information about a construct from a model's activations does not establish that the model uses this information when producing a response. A probe may identify a direction that encodes OR-related information even if that direction plays no part in the computation generating the model's output. Establishing causal operativity therefore requires intervention on the recovered direction and evidence that the intervention changes the model's behaviour in theoretically predicted ways (Belinkov, 2022; Mueller et al., 2026).

The logic rests on how a representational direction enters the model's computation. At a given layer, the model's activation state contains multiple components that are passed to subsequent layers. If the recovered OR direction contributes to the processing that produces an OR assessment, changing the activation along that direction should change the strength of the OR-related signal available to downstream computation. Adding activation along the direction should therefore increase the influence of OR-related information, whereas subtracting activation should reduce it. Ablating the direction provides a complementary test by removing that component from the activation state. Adding the direction shows that it is sufficient to move the readout, whereas removing it shows whether the model needs the direction to produce its usual response. If the direction is functionally involved, OR-related assessments should decline relative to an otherwise comparable control intervention.

This directional logic is important because an intervention may alter model outputs simply by disrupting processing. A behavioural change alone is therefore insufficient. Evidence for causal operativity requires interventions that leave the model otherwise coherent and a signed pattern in which OR-up strengthens OR-related assessments and OR-down weakens them, each benchmarked against random directions of comparable magnitude, while ablation reduces them relative to matched control directions (e.g., Arditi et al., 2024; Rimsky et al., 2024; Turner et al., 2023).

> ***Proposition 2 (Causal operativity).*** *A representational direction related to an entrepreneurship construct is causally operative inside an LLM when intervention on it produces theoretically predicted changes in the model output. For OR, increasing activation along the OR direction should strengthen OR-related assessments, decreasing it should weaken them, and removing the direction should reduce them relative to matched controls.*

### 2.4.3 Representational integration

The third property concerns relationships among construct-related directions. A representational architecture does not have to be a collection of isolated modules. Entrepreneurship constructs occupy distinguishable but connected positions within theory, and the internal organisation of an LLM may reflect both differentiation and connection.

For OR, opportunity theory provides specific expectations. As noted above, McMullen and Shepherd (2006) distinguish recognition from the subsequent evaluation of feasibility and desirability. Feasibility and desirability have also played central roles in research on entrepreneurial evaluation and intention (Shapero and Sokol, 1982; Wood and McKelvie, 2015), while exploitation concerns action on a recognised and evaluated possibility. At the same time, Davidsson's (2015) critique suggests that recognition and favourability may become bundled when opportunities are described or assessed.

Representational integration therefore generates two distinct predictions. First, an OR direction should be geometrically distinguishable from directions associated with evaluation and exploitation. It should also be separable, both geometrically and under intervention, from theoretically adjacent correlates, such as positive affect (Baron, 2008) or the presence of a commercial setting. Second, because opportunity theory connects these processes, intervention on the OR direction may nevertheless shift responses concerning evaluation and exploitation. Geometric separation and causal coupling are not contradictory. A direction can occupy a distinct position in activation space while remaining connected to related processes at the level of output. Evidence of both differentiation and coupling indicates a structured architecture in which construct-related directions are distinct but not fully modular.

***Proposition 3 (Representational integration).*** *Representational directions related to entrepreneurship constructs are integrated inside an LLM when they occupy distinguishable positions in activation space while remaining causally connected at the level of model* *output. For OR, the OR direction should be geometrically distinguishable from evaluation and exploitation, while intervention on it should produce coordinated changes in evaluation and exploitation assessments.*

## 3. Method and results

A common workflow in representation engineering is to construct inputs that differ on a target concept, record the model's internal activations, estimate a candidate representational direction, and validate that direction through predictive, causal, and geometric tests (e.g., Arditi et al., 2024; Chen et al., 2025; Marks and Tegmark, 2024). We adapt this workflow to OR and organise the analysis in three cumulative parts, each corresponding to one theoretical proposition and its evidentiary requirements (Table 1).

Part 1 (Section 3.1) tests Proposition 1, representational existence, by examining whether OR-related information can be recovered from the internal activations of Llama 3.1 8B-Instruct. We construct matched contrastive scenarios that differ in whether OR is present but remain as similar as possible in incidental features. We then estimate a direction in the model's activation space that separates OR-present from OR-absent scenarios. Successful recovery on held-out scenarios indicates that OR-related information is linearly decodable from the model's internal states. This evidence is correlational and does not establish that the model uses the recovered direction when producing a response (Belinkov, 2022).

Part 2 (Section 3.2) tests Proposition 2, causal operativity, by intervening directly on the recovered OR direction. We use activation steering to add or subtract the direction during inference and assess whether the model's opportunity judgments shift as theoretically predicted. We complement these steering tests with directional ablation, in which the direction is removed from the model's internal activations. If steering shifts the model's judgments in the predicted direction and ablation weakens them relative to matched controls, the direction is causally operative rather than merely decodable.

Part 3 (Section 3.3) tests Proposition 3, representational integration, by examining how the OR direction relates to feasibility, desirability, and exploitation. We first test whether steering the OR direction shifts the model's evaluation and exploitation assessments. We then estimate separate directions for these constructs using independently constructed batteries and compare their geometry with that of the OR direction. This tests whether OR is representationally distinct from adjacent constructs.

*- Please insert Table 1 about here -*

## 3.1 Part 1: Locating the OR direction

### 3.1.1 Scenario design

We begin by constructing contrastive scenario pairs designed to isolate OR as much as possible from incidental surface features. We adapt the standard contrastive logic of representation engineering (e.g., Marks and Tegmark, 2024; Rimsky et al., 2024; Zou et al., 2023) to entrepreneurship construct validation by matching the scenarios on length, topic, register, setting, actor type, and surface framing.

Because OR has no single canonical operational definition (e.g., Hansen et al., 2011; Shane and Venkataraman, 2000; Short et al., 2010), we use the definition introduced in Section 2.4: the cognitive act of noticing a generalisable structure in a scenario that could serve as the basis for a potential value-generating activity, independent of evaluation or action. The definition follows Baron's (2006) pattern-recognition account and Grégoire et al.'s (2010) structural-alignment view, keeps recognition prior to evaluation and exploitation (McMullen and Shepherd, 2006; Shane and Venkataraman, 2000; see also Kuckertz et al., 2017), and addresses Davidsson's (2015) concern about assumed favourability. We therefore treat OR as the recognition of a potential value-generating structure, not as a guaranteed gain, a favourable evaluation, or a decision to act.

The two poles of our contrastive pairs follow from this definition. In OR-present scenarios, the actor interprets a recurring signal as a generalisable structure that could support a potential value-generating activity. Both poles depict concrete work, so the contrast lies in how the actor construes the signal, although the OR-present construal is typically expressed through a broader offering. In OR-absent scenarios, the actor channels the same signal into a local response that does not generalise, for example managing it within existing procedures, absorbing it into an existing routine, or handling it as a one-off.

Contrastive batteries in prior representation engineering range from a few hundred pairs (e.g., Rimsky et al., 2024) to several thousand (e.g., Marks and Tegmark, 2024). Hand-authoring a tightly matched battery at this scale would be difficult to audit consistently. Following the broader synthetic-data tradition (e.g., Wang et al., 2023), we generate candidate pairs with Claude Opus 4.7 (Anthropic, 2026a), a frontier LLM available at the time of data generation (May 2026).

We provide Claude Opus 4.7 with the operational OR definition and instruct it to produce 5,000 pairs of OR-present and OR-absent scenarios (Appendix A.1). We impose rules at three levels (Appendix A.2). First, at the vocabulary level, we ban construct labels (e.g., opportunity, gap, recognize) and role labels (e.g., entrepreneur, founder, CEO). This targets the word-identity shortcut that motivates control-task probing (e.g., Hewitt and Liang, 2019) and removes the single-word cues most likely to let a classifier recover the label. Second, at the pair level, the OR-present and OR-absent scenarios must match on incidental features and fall within a bounded token-overlap range. This excludes both trivial word swaps that a probe could exploit as a shortcut and freely diverging rewrites that reintroduce incidental differences (e.g., Geirhos et al., 2020). Third, at the batch level, failure modes and openings must vary across pairs so that no recurring failure type or sentence template can stand in for the OR contrast (e.g., Chen et al., 2025; Marks and Tegmark, 2024).

We ask the generator to self-check each pair against these rules. We then apply a deterministic validation step that removes pairs that the automated checks flag (e.g., word counts, token-overlap band, banned words, and first-person pronouns). To detect subtler lexical asymmetries, we apply an adversarial filter modelled on AFLite (Le Bras et al., 2020; Appendix A.3). The adversarial filter reduces the validated pool of 5,000 pairs to our final sample of 636 scenario pairs and lowers bag-of-words classifier accuracy from 0.912 to 0.528 (chance is 0.500).

These procedures substantially reduce, though they cannot eliminate, the surface-form confounds common in representation engineering research. Table 2 illustrates the contrast with three example pairs.

*- Please insert Table 2 about here -*

#### 3.1.2 Activation extraction

Activation extraction passes each scenario through an LLM (different from the LLM used to generate our scenarios) and records its internal activations. In transformer-based LLMs, the input text is tokenised into words and word-pieces and processed through a sequence of layers. At each layer and token position, the model produces an activation vector, a high-dimensional numerical state representing the model's intermediate computation at that point. These activations are the central object of study in MI (e.g., Bereska and Gavves, 2024; Mueller et al., 2026).

Representation engineering requires access to an LLM's internal activations, which proprietary models (e.g., ChatGPT, Claude, Gemini) do not expose. We therefore use Meta's open-weight model Llama 3.1 8B-Instruct (Grattafiori et al., 2024), in line with recent representation engineering work (e.g., Arditi et al., 2024; Marks and Tegmark, 2024; Rimsky et al., 2024). At 8 billion parameters, the model is capable of following the experimental prompt format while remaining computationally tractable for repeated extraction and intervention runs. Using TransformerLens (Nanda and Bloom, 2022), we record residual-stream activations at every layer as the model processes each scenario. The residual stream is the running internal state that each layer reads from and writes back to, carrying information forward through the network. Full extraction details are in Appendix B.

We randomly split the 636 contrastive pairs into a training set (509 pairs, 80%) and a held-out test set (127 pairs, 20%). We estimate the OR direction on the training pairs only and use the held-out pairs to test generalisation.

#### 3.1.3 Direction estimation and validation

Direction estimation identifies a candidate axis in the model's activation space that separates OR-present from OR-absent activations, treating the contrast as a single direction in line with the linear representation hypothesis. We do not fix the layer or read position in advance because probe informativeness varies with depth and read position (e.g., Gurnee and Tegmark, 2024). We therefore search over all 32 layers and several read positions (Appendix C), using the training data only, and select the layer and position at which the OR contrast is most strongly recoverable.

We recover the direction using two methods, the difference-in-means (DiM) direction and a linear probe. For DiM, we average the OR-present training activations, average the OR-absent training activations, and take the difference between the two class means, which points from OR-absent toward OR-present (e.g., Marks and Tegmark, 2024). Because DiM is defined by the class means alone, without fitting to maximise separation, it provides a constrained readout of the contrast. The linear probe is a logistic regression (LR) classifier trained to separate OR-present from OR-absent activations (e.g., Alain and Bengio, 2017; Belinkov, 2022).

Part of its accuracy can come from classifier flexibility itself (e.g., Hewitt and Liang, 2019). We therefore use DiM as the primary selection criterion and treat LR as a corroborating readout.

The training-based selection procedure identifies layer 12 of 32. At this layer, the DiM direction classifies held-out scenarios with 0.748 accuracy (95% CI [0.705, 0.791]), while the LR probe reaches 0.839 (95% CI [0.795, 0.878]). Both exceed the 0.500 chance level and the 0.528 bag-of-words leakage baseline (Section 3.1.1). The 127 held-out pairs enter neither direction estimation nor layer and read-position selection, so these are out-of-sample estimates. A within-pair label-permutation test rejects random assignment of the OR-present and OR-absent labels ($p=0.002$, Appendix C). Figure 3 shows held-out test accuracy by layer. DiM accuracy is highest in the early-middle layers, where we select layer 12, and declines in later layers, whereas LR accuracy remains relatively high in the upper layers.

Two further checks support the interpretation of the OR direction. First, a small neural network classifier does not improve performance relative to the LR probe (0.008 lower, 95% CI [-0.043, 0.028]), consistent with the OR contrast being substantially captured by a single linear direction (Gurnee and Tegmark, 2024), though not proving that the representation is exclusively one-dimensional. Second, the direction is not driven by scenario topic. Re-estimating it while holding out each of 15 topic clusters yields almost identical directions across folds (mean fold-to-fold cosine 0.996) and separates scenarios of the held-out cluster at 0.786 accuracy, comparable to the random-cluster control (0.792).

Finally, we use directional ablation as a consistency check on the layer-12 OR direction (e.g., Arditi et al., 2024). Ablation removes the component of each activation along the OR direction, equalising the OR-present and OR-absent class means and removing the DiM signal by construction (Appendix C). The re-trained LR probe therefore returns 0.500 on the held-out pairs, the chance value the construction forces for any direction estimated as a class-mean difference, whereas ablating a random direction leaves performance essentially unchanged (0.835). The informative comparison is the random-direction result, which shows that removing an arbitrary direction does not degrade probe performance. The stronger test of whether the direction is functionally necessary for model behaviour is the behavioural ablation in Section 3.2.

Together, these results support Proposition 1 within the estimation battery, indicating that an OR direction separates held-out scenarios well above chance and above the lexical baseline and survives topic holdouts. Whether the direction also survives changes in length, phrasing, person register, and scenario generator, the generalisation half of Proposition 1, is tested in Section 4.

*- Please insert Figure 3 about here -*

## 3.2 Part 2: Steering the OR direction

### 3.2.1 Activation steering procedure

To test the causal question, we use activation steering to modify the model's internal activations during inference and then measure whether its outputs shift in the predicted direction (e.g., Rimsky et al., 2024; Turner et al., 2023; Zou et al., 2023). We implement steering as an additive hook (Turner et al., 2023) in TransformerLens. At the selected layer (layer 12 of 32, hereafter the operating layer), the hook adds a scaled copy of the OR direction to the model's activation at the final position of each forward pass and leaves every other token position unchanged. In the single-pass forced-choice readout this is one fixed answer position, and in free-text generation it acts on each response token as that token is generated (Appendix D.3). Writing h for that activation and r for the OR direction, the steered activation is

$$\tilde{h} = h + \alpha \cdot r$$

Here, α is the steering coefficient, which sets the magnitude and sign of the intervention. Positive values define OR-up steering (the OR dial turned up), expected to shift the model toward OR-present judgments, and negative values define OR-down steering, expected to shift them toward OR-absent judgments. We provide further details on scaling, position, and calibration in Appendix D.1.

We select α using a coherence criterion based on next-token Kullback-Leibler (KL) divergence, following Arditi et al. (2024). This avoids choosing the coefficient that maximises a behavioural outcome (e.g., Wu et al., 2025). We sweep α over a predefined grid and retain the largest value for which the steered model's next-token distribution remains close to its unsteered baseline on a held-out calibration set. We define this coherence gate as KL≤0.100. KL divergence measures how much two probability distributions differ. This procedure selects α=3.5.

Within the KL-bounded range, the steering effect increases with α and remains directionally consistent across most scenarios, with 0.161 of scenarios showing a negative response slope across the KL-passing coefficients (the anti-steerable fraction of Tan et al., 2024; Appendix D.2). The effect is already present at the next-lower grid point, indicating that it does not depend on selecting a coefficient at the KL boundary. Beyond this range, the mean response continues to rise, but the per-scenario response becomes unstable, with 0.375 of scenarios showing a negative slope across the full grid, and on bands whose baseline sits near the response ceiling, the OR-up shift reverses at the doubled coefficient (Appendix G.1). Figure 4 plots this dose-response pattern together with the coherence cost that motivates the gate. We evaluate the steering effect with two readouts, forced-choice and free-text.

*- Please insert Figure 4 about here -*

#### 3.2.2 Forced-choice readout

The forced-choice readout provides our primary evidence that steering shifts the model's opportunity judgments. We append a yes/no question to each scenario and measure the change, relative to the unsteered baseline, in the log-odds of the model answering "yes" versus "no" at the first answer position (e.g., Arditi et al., 2024; Rimsky et al., 2024) (Appendix D.2). We read next-token probabilities directly, without sampling, so the shifts are deterministic. Log-odds, defined as the logarithm of P(yes) divided by P(no), makes shifts comparable across scenarios with very high or low baseline probabilities, where raw probabilities move only slightly.

The model judges standalone scenarios, generated with the same procedure but without contrastive pairing. The first set contains 300 entrepreneurial scenarios with ambiguous OR status, distributed across weak, medium, and strong opportunity bands. The second set contains 400 everyday scenarios across four bands with no OR anchor, covering natural observation, domestic routine, interpersonal moment, and one-sentence everyday scenes. These scenarios test whether steering imposes OR readings where the construct has no foothold. After excluding the calibration scenarios used to select α (5 per band), 665 scenarios remain. The appended question is "Does the scenario describe an entrepreneurial opportunity? Answer yes or no." Framed in the third person, the question asks whether the scenario presents an opportunity for someone, corresponding to third-person recognition in the sense of McMullen and Shepherd (2006), leaving aside whether the opportunity would be attractive or actionable for the respondent (e.g., Ramoglou and Tsang, 2016).

Table 3 reports changes in log-odds under OR steering across scenario types. On the entrepreneurial scenarios, the model already leans toward yes at baseline (P(yes)=0.891), and steering shifts the judgment in both directions. OR-up raises the mean log-odds of "yes" by 0.401, while OR-down lowers it by 0.716, yielding a total swing of 1.117 log-odds. The effect appears in all three opportunity bands. In probability terms, mean P(yes) moves from 0.891 at baseline to 0.924 under OR-up and 0.853 under OR-down. The modal answer flips for 5 of 285 scenarios under OR-up steering and 4 under OR-down steering, always in the direction of the steer, with most of these scenarios in the weak-opportunity band (Table 3). Because several baselines sit near the response ceiling and the KL criterion caps intervention strength, the effect is graded, not categorical. On the everyday scenarios, the baseline P(yes) is near zero (0.004), placing the log-odds deep in the no region (-7.229). Both steering directions nudge the log-odds upward, by 0.322 under OR-up and 0.412 under OR-down (Table 3). Same-direction movement under both signs, never exceeding a probability change of 0.008 in any everyday band, is consistent with a generic perturbation at a probability floor, without any construct-carrying effect. The everyday panel therefore shows an absence of construct-carrying movement at the floor. It does not show that steering cannot manufacture opportunity readings. The band contrast is consistent with context dependence, although it does not by itself establish direction specificity. We examine this specificity using random-direction comparisons, directional ablation, and a valence-direction comparison.

First, on a band-balanced subset mixing entrepreneurial opportunity and everyday scenarios, 100 norm-matched (equal-length) random directions shift log-odds by 0.050 (SD 0.279) under OR-up steering and 0.012 (SD 0.284) under OR-down steering. On the same subset, the OR direction shifts log-odds by 0.369 under OR-up steering. This comparison is descriptive, however, because 11 of 100 random directions reach the OR-up shift (Appendix D.4). The pooled comparison is less informative for OR-down, where the OR direction shifts log-odds by only -0.073 because the everyday scenarios sit at the probability floor, where OR-down steering moves the log-odds upward, offsetting the decrease on the entrepreneurial scenarios (Table 3). Direction-specific evidence for the negative sign therefore rests primarily on the directional-ablation test below and on the discriminant geometry reported in Section 3.3.2. Second, ablating the OR direction from the residual stream and re-reading the recognition questions (Appendix D.2) lowers their log-odds by 0.556 relative to the unablated model, compared with -0.002 for a matched random ablation. The paired contrast is -0.554 (95% CI [-0.658, -0.451], $p<0.001$). This directional-ablation test follows Arditi et al. (2024) and Elazar et al. (2021) and supports direction-specific necessity for the forced-choice readout. A further check steers a norm-matched positive-valence direction and asks whether each direction moves its own readout more than the other's. This causal dissociation from valence is not supported (double-dissociation index -0.097, 95% CI [-0.280, 0.081]).

A strengthened null comparison draws 1,000 isotropic and 1,000 label-shuffled directions and forms norm-matched and KL-matched families from each (Appendix D.1). The OR-down effect clears both norm-matched families (label-shuffled at the 97.6th percentile, p=0.025, and isotropic at p=0.010) but neither KL-matched family (p=0.052 in both), while the OR-up effect lies at the 91.0th percentile of the label-shuffled norm-matched family (p=0.091) and remains directionally large but less decisive throughout.

Together, these results support Proposition 2, although the evidence for causal specificity is stronger for directional ablation and OR-down steering than for the OR-up and valence comparisons. Adding the direction raises the recognition readout, subtracting it lowers it, and removing it reduces the readout relative to a matched random ablation, the signed pattern the proposition predicts. We therefore treat the OR direction as causally involved in the recognition readout, not as a demonstrably selective cause of it.

*- Please insert Table 3 about here -*

### 3.2.3 Free-text readout

The free-text readout tests whether steering changes how the model explains its opportunity judgment. For each of the 95 scenarios per band in the three entrepreneurial opportunity bands of the held-out evaluation split, which is disjoint by construction from the calibration scenarios, we pose the question from Section 3.2.2 in open form and ask the model to explain its reasoning in about three sentences. We generate explanations under the baseline, OR-up, and OR-down conditions using greedy decoding, which always selects the model's most likely next token (200 token cap). We analyse these explanations qualitatively and then apply an automated LLM judge to the entrepreneurial scenarios (285 scenarios across the three opportunity bands).

Table 4a shows three example scenarios, one per opportunity band, each paired with the model's baseline, OR-up, and OR-down explanations. Under OR-up steering, the model often reframes the same situation as a potential venture by naming a generalisable pattern and a possible next step. Under OR-down steering, it more often keeps the same situation within existing arrangements, though the shift is graded and some OR-down explanations still concede a possible opportunity (Table D.1). Because we select these examples for clarity, they illustrate the contrast qualitatively and do not estimate the average effect across the evaluation set.

We then score the explanations with an LLM judge, a protocol in which a separate model rates target-model generations against a predefined rubric (Chen et al., 2025; Wu et al., 2025; Zheng et al., 2023). For each scenario, the judge sees the scenario and the two steered explanations, OR-up and OR-down, unlabelled and without the baseline, and selects the explanation that more clearly recognises an entrepreneurial opportunity. The rubric defines OR as noticing a generalisable structure that could serve as the basis for a potential value-generating activity, independent of evaluation or action, and instructs the judge to reward noticing the structure itself, without weighing or acting on it, and to disregard length or elaboration. The judge is DeepSeek V4 Pro (DeepSeek-AI, 2026) with reasoning mode enabled, a model family different from both the steered Llama model and the Claude generator, so the judge is independent of the models whose outputs it rates. We report the win rate among decisive pairs and an exact binomial test against the 0.500 chance level. To mitigate position bias, an order-doubled robustness protocol additionally judges each pair in both presentation orders under a conservative swap rule, counting a win only when the same explanation wins in both orders (Zheng et al., 2023; Appendix D.3).

Table 4b reports both protocols. In the primary single-presentation protocol, the judge sees each pair once, with the OR-up response on a randomly assigned side, and makes one decision per scenario. Under this protocol, the judge prefers the OR-up explanation in 0.656 of the 285 entrepreneurial pairs (95% CI [0.599, 0.709], exact binomial $p<0.001$). The weak and medium bands sit clearly above chance (both 0.705), whereas the strong band does not (0.558, $p=0.305$), a pattern consistent with limited headroom where the baseline explanation already recognises the opportunity. The stricter order-doubled protocol yields an OR-up preference of 0.708 among the decisive entrepreneurial pairs (Table 4b). A length check does not support a verbosity account, as OR-up explanations are shorter than OR-down explanations (85.239 against 92.881 model tokens). The intervention therefore changes both the model's forced-choice judgment and the way it articulates the entrepreneurial content of a scenario.

The free-text results therefore provide complementary, but not uniform, support for Proposition 2. Because the preference also depends on the judging rubric, we treat it as convergent evidence that steering changes how the model articulates opportunity-related content, not as a separate confirmatory causal test.

*- Please insert Tables 4a and 4b about here -*

## 3.3 Part 3: Relating the OR direction to adjacent constructs

### 3.3.1 Downstream coupling from OR to evaluation and exploitation

Proposition 3 holds that the OR direction is part of a broader representational organisation. McMullen and Shepherd (2006) place recognition upstream of two evaluative doubts, feasibility (whether the actor can enact the possibility) and desirability (whether attaining the outcome would fulfil the actor's motive), and upstream of the commitment to exploit. If the OR direction is causally coupled to this structure, steering it should shift the model's feasibility, desirability, and exploitation judgments.

We hold the OR direction, operating layer, and steering coefficient of Part 2 fixed and apply the same additive hook to the entrepreneurial scenarios. To each scenario, we append three forced-choice questions: feasibility (whether the person could practically act, regardless of whether it would be worthwhile), desirability (whether acting would be worth the cost and effort), and exploitation (whether the person should actively pursue the situation now). We present the questions across the full evaluation battery and report results for the entrepreneurial scenarios because these downstream judgments are meaningful only in opportunity-relevant contexts. For each question, we record the steering-induced change in log-odds. Because these downstream questions do not enter the estimation of the OR direction, the calibration of the steering coefficient, or layer selection, a shift in these readouts reflects cross-readout transfer and cannot be an artefact of circularity (Appendix E.1).

Under OR-up steering, mean log-odds of "yes" rise for every readout (Table 5), increasing by 0.401 for recognition, 0.728 for feasibility, 0.338 for desirability, and 0.483 for exploitation. Under OR-down steering, all four readouts fall, decreasing by 0.716 for recognition, 0.456 for feasibility, 0.331 for desirability, and 0.421 for exploitation. The exploitation question contains action content that the recognition construct excludes. We interpret its movement as causal coupling with entrepreneurial action as modelled by McMullen and Shepherd (2006), not as evidence that the OR direction is itself an action direction. We also do not compare effect magnitudes across readouts because the questions begin at different baselines. Feasibility, for example, begins near the response ceiling (baseline P(yes)=0.978), so a large log-odds shift corresponds to only a small probability change. We therefore read the pattern as broad coupling across adjacent readouts, not as a graded decay from recognition to exploitation. The four readouts also differ in per-scenario coherence. At the selected coefficient, and computed across all 665 evaluation scenarios rather than the 285 entrepreneurial scenarios in Table 5, the share moving against the intended steer under OR-up is 0.116 for recognition, 0.030 for feasibility, 0.075 for desirability, and 0.492 for exploitation. The exploitation split arises on the everyday scenarios, whose exploitation readout falls under OR-up, not on the entrepreneurial means reported in Table 5.

Ablation further indicates that the OR direction is causally involved in these downstream readouts. Removing the OR direction from the residual stream at the operating layer and re-reading the three downstream questions lowers their mean log-odds by 0.404 (95% CI [0.313, 0.499]). The paired contrast relative to a matched random ablation is -0.400 log-odds (95% CI [-0.496, -0.310], uncorrected $p<0.001$). These downstream readouts are closer to a general evaluative direction than the recognition questions. Under OR-up steering, they exceed the corresponding affect and goodness discriminants by only 0.049 log-odds (95% CI [0.015, 0.082]), which is consistent with their theoretical connection to recognition. Accordingly, the construct-specificity claim rests on the OR-versus-discriminant separation in Appendix D.4, and not solely on the magnitude of downstream effects (Appendix E.1).

We frame these findings as causal coupling rather than mediation. Mediation would require evidence that the OR direction lies on the causal path from recognition to exploitation within the model's computation. Our additive steering and ablation interventions cannot establish this, which would require interchange interventions that replace the recognition value along the relevant computational pathway (e.g., Geiger et al., 2024; Mueller et al., 2026). The results instead indicate that the OR direction is a representational structure on which the feasibility, desirability, and exploitation readouts partly depend.

*- Please insert Table 5 about here -*

### 3.3.2 Discriminant geometry against the evaluation and exploitation constructs

The coupling result raises a question of identity. If steering OR moves feasibility, desirability, and exploitation, the OR direction might merely be a relabelled representation of one of these constructs, undermining its status as a distinct representation causally coupled to them. We address this question geometrically. If each construct is represented, approximately, by a linear direction (e.g., Zou et al., 2023), distinctness can be assessed by the angle between their estimated directions, complementing the behavioural coupling test (e.g., Park et al., 2024). This is analogous to a discriminant-validity test in the tradition of Campbell and Fiske (1959).

Because every direction here comes from the same DiM method in the same model, shared-method variance remains a standing alternative. The cross-generator and cross-model checks of Section 4 address part of this rival without eliminating it, because all batteries derive from a common specification. Before we turn to the geometry, the same identity question has a behavioural answer. The full question battery of Appendix D.4 holds the steering direction and scenarios fixed and varies only the question. Steering moves the oblique OR questions by 0.207 log-odds more than matched pleasantness and goodness questions under OR-up and by -0.182 under OR-down (both Holm-corrected, Appendix E.1), and on the entrepreneurial panel the same margin widens to 0.273 and -0.526 (Table D.2). Because pleasantness and goodness capture exactly the positivity a generic-affirmation account predicts should move, this selectivity is the central evidence that the dial carries OR content and not merely a general positive-valence signal. Appendix D.4 reports the full battery and its boundaries.

We estimate a direction for each downstream construct using the DiM procedure from Part 1 and independently constructed contrastive batteries that follow the same design rules (Appendix E.2). In each comparator battery, the underlying recognised opportunity is held constant across the two poles, so the estimated direction captures the feasibility, desirability, or exploitation contrast without re-estimating recognition. We measure geometric alignment as the cosine between directions and interpret each OR-comparator cosine against two references: OR's split-half reliability, estimated in the same model and layer, and the analytic random-direction scale. That scale, the standard deviation of the cosine of an arbitrary direction in the 4,096-dimensional space around zero, is approximately 0.016.

Figure 5 reports the results. The OR split-half reliability reference is 0.925, whereas the most aligned comparator is exploitation, with a cosine of 0.328 (95% CI [0.284, 0.356]). This modest positive alignment fits two accounts: the theoretical adjacency of noticing and committing to act, and a design account, because the OR-present pole expresses the actor's interpretation through a broader and typically priced offering, which shares content with acting on a possibility. The present batteries cannot separate these accounts, so we treat the exploitation alignment as bounded evidence of adjacency, not as confirmation of stage structure. Feasibility follows at 0.174 (95% CI [0.123, 0.215]), while desirability is near orthogonal to OR at -0.017 (95% CI [-0.048, 0.015]), indistinguishable from the random-direction control. Desirability therefore provides the clearest case of coupling without geometric identity. The ordering is also informative, subject to the same design caveat. The two-stage account groups feasibility and desirability as the evaluative pair that follows recognition and places exploitation downstream of both, whereas in the model's geometry recognition sits closest to exploitation, further from feasibility, and near orthogonal to desirability. All three comparator cosines lie well below the OR split-half reliability reference, and each discriminant gap remains significant after the Holm (1979) correction ($p<0.001$). A variance-adjusted recheck that accounts for uneven stretching of activation spaces leaves the ordering unchanged (Appendix E.2).[5]

A joint-span test further indicates that OR is not well approximated by the three comparator directions together. Projecting the OR direction onto their combined span accounts for only 0.116 of the direction (the $R^2$ of its projection, 95% CI [0.086, 0.138]), leaving the majority of the OR direction (0.884) outside the comparator space. A reliability-adjusted residual analysis confirms this pattern, with 0.993 of reproducible OR alignment remaining after removing the joint comparator span (Appendix E.2). This distinctness is nevertheless bounded. Both the construct batteries and entrepreneurship theory suggest that OR can contain some evaluative content, consistent with Davidsson's (2015) argument that recognition and perceived favourability are often conflated within the opportunity construct.

[5] For calibration, Chen et al. (2025) report cosine alignment of comparable and often greater magnitude among distinct persona directions, which they characterise as moderate. The present alignments are therefore modest by the standards of this method.

A stricter causal specificity check compares steering of the OR, feasibility, desirability, and exploitation directions at the same norm-matched coefficient and includes the commerciality direction from Appendix F.2 as a context control. We then read all four construct questions (Table E.2) under each intervention. The matrix provides two complementary tests. First, no rival direction should move the OR question as strongly as the OR direction itself. Second, each direction should move its corresponding question more than the average of the remaining questions. We read every comparison alongside the realised KL dose for each direction. The first criterion is supported for desirability, exploitation, and commerciality, but not for feasibility. The feasibility direction moves the OR question more strongly than the OR direction itself, despite operating at a lower realised dose (Table E.2, Panel B). The second criterion is also not fully supported. Rival directions generally influence their own question less than expected relative to the other readouts, and only for desirability is that shortfall distinguishable from zero (-0.120, 95% CI [-0.167, -0.075]), while the shortfalls for feasibility and exploitation are not (Table E.2, Panel B).

Together, the geometric differentiation and cross-readout effects support Proposition 3's prediction of representations that are distinguishable in geometry and connected at the level of model output, while the specificity matrix shows that this organisation is not fully modular or perfectly selective. The results weigh against the view that the OR direction is merely a favourability or actionability signal relabelled as recognition, because it is geometrically separable from the feasibility, desirability, and exploitation directions estimated from these comparator batteries. Moreover, their joint span accounts for a minority of its reproducible alignment, and they weigh against a fully isolated-module account because steering OR shifts all three downstream readouts. The sharpest qualification is the construct steer-read matrix: the feasibility direction moves the recognition readout more than the OR direction does at a lower realised dose, consistent with the role of feasibility in the first-person evaluation that follows recognition in McMullen and Shepherd's (2006) account.

*- Please insert Figure 5 about here -*

## 4. Robustness checks

The main analyses provide evidence for the representational existence, causal operativity, and representational integration of the OR direction. We conduct five sets of robustness checks to examine whether these conclusions depend on specific features of the research design. These checks address (1) the surface form of the scenarios, (2) commerciality as a rival construct, (3) the LLM that generated the scenarios, (4) the scale and family of the analysed LLM, and (5) the evaluation of the free-text responses. Table 6 summarises the tests and results, and Appendix F reports full procedures, estimates, and diagnostics.

The robustness checks support the stability of the central pattern while also defining its boundaries. The evidence indicates that the recovered direction is not confined to the wording or framing of the original battery (DiM accuracy 0.781 to 0.860 across three out-of-distribution scenario sets, Table F.1). It is also not reducible to commercial content (LR accuracy falls only from 0.839 to 0.811 after removing linearly decodable commerciality on the fitting sample, a lower bound because the erasure is incomplete on held-out pairs, Appendix F.2), and it does not arise solely from the stylistic tendencies of one scenario generator (cross-generator probe transfer at 0.783 to 0.832, Appendix F.3).

While our primary results come from Llama 3.1 8B-Instruct, we repeat the central analyses across four additional LLMs chosen to vary both model scale and family. Llama 3.1 70B-Instruct provides a within-family scale replication, while Qwen3-8B, Qwen2.5-32B-Instruct, and Gemma-3-12B-it extend the analysis across distinct model families and parameter scales. The core pattern of recovery, signed steering, and geometric separation holds across these models, with held-out DiM accuracy ranging from 0.705 to 0.744 (Table F.3). This cross-model consistency makes it less likely that the result is specific to one architecture, model family, or scale.

In addition, the free-text checks provide complementary evidence concerning the behavioural readout. Judging every pair in both presentation orders under a conservative swap rule does not weaken the preference for OR-up responses (0.708 against 0.656, Table 4b), reducing concern that the result is driven primarily by position bias in the LLM judge. Applying the intervention at every token position produces an even stronger response difference (order-doubled win rate 0.971), but also exceeds the predefined coherence gate (mean next-token KL divergence 0.671 against the 0.100 bound, Appendix F.5). We therefore interpret the all-position result as a robustness ceiling, and the calibrated intervention remains the confirmatory estimate. Overall, the robustness checks indicate that the main findings do not depend on a particular scenario format, commercial framing, scenario generator, or evaluation procedure, and that recovery and signed steering also appear in every additional model tested, though construct selectivity replicates only in part (Appendix F.4).

*- Please insert Table 6 about here -*

## 5. Additional analyses

The robustness checks defend the principal conclusions. The additional analyses address a different question. Given the weight that AI scholars assign to role prompting in shaping LLM behaviour (e.g., Shanahan et al., 2023), a natural objection to the steering results is that internal intervention may be unnecessary. If the goal is to elicit stronger opportunity-recognition responses, one could simply prompt the model to adopt an entrepreneurial role (e.g., Obschonka and Fisch, 2025). We therefore compare activation steering with entrepreneurial role prompting using both the forced-choice readout and a survey profile of published entrepreneurship, personality, and off-construct instruments. Table 7 summarises both designs and results, together with the exploratory feature-level analysis introduced below, all of which are detailed in Appendix G.

The two levers come apart on the properties expected from an internal intervention. The role-prompted persona is the broader intervention. It increases opportunity-related judgments across bands and shifts a wide range of survey scales, including off-construct measures such as the ADHD self-report. Steering is narrower and follows the theoretical structure established earlier. OR and theoretically adjacent constructs such as exploitation and alertness move together (the OR scale moves from 0.652 at baseline to 0.718 under OR-up and 0.625 under OR-down steering), consistent with the downstream coupling identified in Section 3.3.1, while off-construct measures change little. This output pattern mirrors the selectivity observed for the OR direction in Part 2 and Section 3.3.2. Steering is reversible in sign and graded by intervention strength within the coherence gate, and OR-down steering partially offsets the persona effect. Because prompt strength and steering coefficients are not expressed on a common scale, the comparison provides convergent evidence, not a second causal test, and the survey profile should not be interpreted as psychometric validation. The OR dial therefore provides a narrower and sign-reversible intervention than entrepreneurial role prompting, though not a perfectly construct-specific control.

An exploratory sparse autoencoder (SAE) decomposition complements the representation-engineering analyses with a bottom-up MI method, asking whether the OR direction aligns with identifiable features in a pretrained feature dictionary (Llama Scope; He et al., 2024). The strongest single-feature alignment is a decoder cosine of 0.341, compared with a random-direction reference of approximately 0.066. The most strongly aligned feature carries a label unrelated to OR, and the remaining alignment is spread across heterogeneous features whose labels capture only partial aspects of the construct, such as strategic planning and community engagement. Of the ten top-aligned features, only one separates OR-present from OR-absent scenarios above chance on held-out pairs, so no single feature accounts for the OR direction. This pattern suggests that the OR direction reflects information distributed across multiple features, not a single monosemantic feature. Because the autoencoder is trained on the base model and applied to instruct-tuned activations, we interpret the decomposition as illustrative only (Table 7, Appendix G.2).

*- Please insert Table 7 about here -*

## 6. Discussion and conclusion

This study develops a structural theory of artificial entrepreneurial cognition and provides its first empirical demonstration through OR. We show that an entrepreneurship construct can have an identifiable and causally consequential counterpart inside an open-weight LLM: an OR direction that is recoverable across varied inputs, experimentally steerable and ablatable, and systematically organised relative to theoretically adjacent constructs. This shifts the empirical point of entry from observable outputs to internal structure. Rather than inferring an entrepreneurship construct from what an AI system says or does, representation engineering, a top-down approach within mechanistic interpretability, allows researchers to locate a construct-related internal structure, intervene on it directly, and observe how the system's judgments change.

The recovered direction separates OR-present from OR-absent scenarios across held-out and altered inputs. Adding and subtracting it shifts forced-choice and free-text opportunity judgments in the predicted directions, and directional ablation reduces the recognition readout, with the off-construct comparisons and the matched-random ablation contrast reducing concerns about nonspecific perturbation. The direction is also geometrically distinguishable from feasibility, desirability, and exploitation, although intervention on it changes judgments concerning all three. Recognition is therefore separable in representational geometry but connected to evaluation and action in model output, and the recovery, signed steering, and geometric separation underlying this pattern hold across four additional LLMs spanning different model families and scales.

These findings do not imply that LLMs understand opportunities, possess entrepreneurial intentions, or reproduce the human entrepreneurial mind. Nor do they imply that the recovered direction is the models' only or complete representation of OR. The narrower conclusion is that the tested systems contain reproducible internal structures associated with OR that are sufficiently stable and functionally relevant to support direct empirical investigation and causal manipulation.

### 6.1 Theoretical implications

The first implication is that AI in entrepreneurship need no longer be theorised solely as a tool, actor, simulator, or source of outputs. It can also be theorised as a representational system. Existing research has primarily asked what AI enables, generates, predicts, or reproduces (e.g., Chalmers et al., 2021; Dimov, 2026; Obschonka and Fisch, 2025). A theory of artificial entrepreneurial cognition asks a different set of questions: which entrepreneurship-relevant distinctions an AI system internally represents, how those distinctions are organised, and which of them causally shape its responses. This responds to calls to make AI itself a subject of entrepreneurship theorising, not only an external technological condition (e.g., Lévesque et al., 2022; Obschonka et al., 2025).

Consistency across model families and scales indicates that the recoverable OR pattern is not unique to a single model. It does not, however, reveal how that pattern arose during training. Nor does cross-model consistency alone exclude the possibility that some of the recovered alignment reflects features of the shared theory-guided scenario design. Research in mechanistic interpretability offers one plausible account of the pattern's origin: predictive training can compress recurring distinctions into reusable internal representations (e.g., Jiang et al., 2024; Lindsey et al., 2025; Templeton et al., 2024). Our findings are compatible with this account but do not establish it.

The second implication concerns opportunity theory and the long-standing question of whether recognition can be separated from the evaluation that follows it (Davidsson, 2015; McMullen and Shepherd, 2006). In the primary model, recognition and evaluation are separable in representational geometry but connected in model responses. The OR direction remains distinguishable from feasibility, desirability, and exploitation, with most of the direction (0.884 by projection) lying outside their combined span, yet steering it moves the model's assessments of all three on the entrepreneurial scenarios. To our knowledge, the model provides a first setting in which this distinction and the potential bundling highlighted by Davidsson (2015) can be observed and experimentally examined within the same AI system. The ordering of the alignments departs from the stage logic. Recognition aligns most with exploitation and least with desirability, although the evaluative pair sits between them in the two-stage account. A design account, in which the OR-present pole's broader offering shares content with action, and a substantive account, in which noticing that something could be built is more tightly bound in text to acting on it than to wanting it, both fit this pattern, and the present data cannot separate them. We claim this as a property of the AI system, not as confirmation of either account for human entrepreneurs. The novelty is that one system allows us to examine where they separate and where they remain connected.

Two scope conditions are important. First, the scenarios adopt a discovery-style framing in which a potentially value-generating structure is present in the described situation (e.g., Alvarez and Barney, 2007). The results therefore do not adjudicate the discovery-creation debate. Second, the study examines recognition at the opportunity-as-happening level of analysis (e.g., Dimov, 2011) and in the third person, never the first (e.g., McMullen and Shepherd, 2006). It tests whether the model represents a described situation as containing an opportunity-relevant structure, not whether the model reaches a first-person assessment that an opportunity is worth pursuing. The scenarios capture a possible opportunity structure, as opposed to a retrospectively confirmed opportunity (e.g., Ramoglou and Tsang, 2016). The theoretical significance therefore lies in demonstrating that opportunity theory can guide the mapping of an AI system's internal organisation, not in using that system to resolve how human entrepreneurs recognise and evaluate opportunities.

The third implication concerns the division of opportunity work between people and machines. Ramoglou et al. (2026) describe a division in which generative AI expands the space of imaginable ventures while human evaluation contracts that space by eliminating possibilities that cannot be actualised. From this perspective, the discriminating work of separating opportunities from non-opportunities remains primarily human. Our findings complicate this division without overturning it. The distinction between a situation containing the theory-defined opportunity-relevant structure and one that does not has a recoverable internal counterpart in the model and can be strengthened or weakened through intervention. The system therefore encodes more than an undifferentiated capacity to generate possibilities. What our results do not establish is that the model applies this distinction reliably under genuine uncertainty, connects it to the realities of a specific market, or bears responsibility for the consequences of acting on it (Ramoglou et al., 2025; Townsend et al., 2025). On this reading the human contribution lies less in possessing the distinction exclusively and more in applying it through experience, contextual knowledge, and responsibility.

The fourth implication concerns the structural theory itself. Its three requirements (recoverability, causal operativity, and systematic organisation) make the theory falsifiable. A direction that separated OR-present from OR-absent scenarios only in the original battery but collapsed under compression, negation, a change of grammatical person, or scenarios produced by a new generator would indicate surface fitting, not recoverability. Failure to recover a corresponding direction in another model would instead limit the cross-model scope of the theory. A direction that decoded the distinction but left model responses unchanged under intervention would indicate information availability without causal operativity. A direction that could not be distinguished from commerciality, evaluation, or exploitation, or that shifted all responses indiscriminately, would fail the requirement of systematic organisation relative to adjacent constructs.

A corollary concerns the scope of causal inference. Representation engineering does not solve entrepreneurship's endogeneity problems (e.g., Anderson et al., 2022; Åstebro and Hoos, 2021; Tzabbar, 2026), nor does evidence from an LLM establish causal relations among human constructs. It changes the level at which intervention becomes possible in an AI system by allowing a candidate internal structure itself to be added, subtracted, or removed. The resulting causal claim is therefore local and bounded. The recovered direction is causally involved in the model's responses under the tested conditions, not in human cognition, field outcomes, or the complete computational pathway generating those responses.

### 6.2 Practical implications

The findings raise an immediate practical question: does the OR dial provide a hidden lever for turning AI into a more powerful tool for entrepreneurial opportunity search?[6] The answer is not straightforward. Turning the dial up raises the probability that the model answers a recognition question affirmatively, and its explanations name a generalisable, value-relevant pattern more often than the OR-down explanations do. Whether OR-up steering would surface possibilities a user would otherwise miss, or identify opportunities that are more novel, feasible, or ultimately successful, is a further step the present design does not test. Greater opportunity sensitivity is also not necessarily better opportunity judgment. Because intervention on the OR direction shifts feasibility, desirability, and exploitation assessments as well, increasing it could make the model more likely to notice a possibility and more inclined to view it favourably. Future research should therefore test whether calibrated steering improves opportunity search or instead trades fewer missed opportunities for more false positives.

[6] Related MI and representation engineering research already points to important practical implications. By treating internal directions as intervention points rather than merely objects of explanation, studies have used activation steering to improve instruction following and truthfulness and to mitigate or control social bias, toxicity, hallucination, and refusal (e.g., Arditi et al., 2024; Rodriguez et al., 2025; Stolfo et al., 2025; Wang et al., 2025).

The opposing steering directions could also support construct-level sensitivity testing. OR-up could provide a more opportunity-sensitive reading, whereas OR-down could offer a more conservative counter-reading. Comparing the two would reveal which judgments remain stable and which depend strongly on the model's OR-related representation. Steering is narrower in what it moves and reversible in sign, and it does both without the off-construct movement the role prompts produce (Section 5). Because the prompt, input, and task remain unchanged, one theory-defined internal influence can be varied while the resulting judgment is observed. Steering therefore provides a controlled stress test of AI-supported entrepreneurial assessments in models whose internal states are accessible.

More broadly, the mechanistic-interpretability perspective introduced here shifts practical attention from how entrepreneurial AI systems are prompted (e.g., Ferrati et al., 2024; Short and Short, 2023) to how entrepreneurship-relevant tendencies are represented and used internally. If other constructs can be located and causally validated with comparable rigour and precision, they may eventually become design variables that can be examined and selectively adjusted, no longer broad characteristics inferred only from model outputs. The OR dial provides an initial demonstration of this broader possibility.

### 6.3 Limitations and future research

Several limitations define the scope of the study and priorities for future research. First, we examine one focal construct, and only Llama 3.1 8B-Instruct receives the complete causal-selectivity pipeline. The additional runs omit the random-direction control, the strengthened null comparison, and the directional-ablation necessity estimate, and the headline model's clean null-tier separation reappears fully in only two of them (Llama 3.1 70B-Instruct and Qwen2.5-32B-Instruct), with a much weaker gap in Gemma-3-12B-it and a reversal in Qwen3-8B. Effect sizes are also not directly comparable because intervention strength was calibrated separately for each model. Nevertheless, the central pattern holds across four additional models. Future research should apply the complete pipeline across further architectures and test whether OR is represented through multiple directions or higher-dimensional subspaces.

Second, the study relies on synthetic scenarios. The contrastive design, lexical restrictions, adversarial filtering, and cross-generator analyses reduce surface-level confounds, but cannot establish transfer to naturally occurring entrepreneurial material. At the same time, this control is what allows the OR contrast to be isolated from wording, topic, and setting. The present study therefore establishes the direction under controlled construct-validation conditions. Research using pitch transcripts, investor memoranda, and founder interviews can test whether it transfers to noisier and strategically presented information. The OR-present and OR-absent labels also rest on the generation specification and automated validation. Independent human coding of the released battery remains an open task.

Third, the study examines a bounded form of OR. The scenarios concern another actor's situation and therefore capture third-person recognition rather than a first-person opportunity in the sense of McMullen and Shepherd (2006). The second-person analyses demonstrate transfer beyond the original grammatical framing, but do not establish intention or agency. The discriminant analyses are also confined mainly to the downstream boundary between OR, feasibility, desirability, and exploitation. Separation from upstream constructs such as entrepreneurial alertness and prior knowledge remains untested.

Fourth, the causal evidence is not equally strong across all specificity tests. Directional ablation provides the clearest direction-specific evidence, and OR-down steering clears the strengthened null comparison under norm-matching, whereas OR-up steering does not. The causal double dissociation from positive valence is also unsupported. The OR direction should therefore not be interpreted as a perfectly selective or isolated module. Yet adding, subtracting, and removing it produces theoretically coherent changes in opportunity judgments, and ablation sharply distinguishes it from a matched random direction. The downstream effects further indicate coupling with feasibility, desirability, and exploitation, although they do not establish mediation or computational sequence. Interchange interventions and circuit-level analyses could examine these mechanisms (e.g., Geiger et al., 2024; Mueller et al., 2026). The present evidence supports the bounded theoretical claim that an OR-related structure is recoverable, causally operative, and systematically connected to adjacent constructs.

Fifth, the free-text analysis relies primarily on an LLM judge. Judging each pair in both presentation orders reduces concern about position bias, and the verdict-cue check of Appendix F.5 shows the OR-up preference surviving in the 221 pairs whose explanations open with the same verdict, but machine evaluation may still track the model's own framing rather than the substance of the explanation. Expert raters and protocols that remove the opening verdict before evaluating the explanation would provide a stronger test. The free-text results should therefore be viewed as complementary to the forced-choice and ablation evidence. Within that role, they show that steering affects the explanations the model generates as well as its next-token probabilities. A design constant also applies to the forced-choice readout, where every probe ends with the same instruction and names yes before no, so the analysis cannot separate a residual answer-order preference from the construct effect.

These limitations define a broader research programme. The pipeline can be extended to alertness (Tang et al., 2012), entrepreneurial self-efficacy (Chen et al., 1998), effectuation (Sarasvathy, 2001), and other entrepreneurship constructs. Moreover, mapping several constructs within the same models could reveal a representational architecture of artificial entrepreneurial cognition. Comparisons with human experts, interchange interventions, and construct-level audits could further connect the present demonstration to questions of entrepreneurial judgment, transparency, and governance.

### 6.4 Conclusion

In a world in which AI systems increasingly take on entrepreneurial tasks, entrepreneurship research can no longer confine the study of cognition to human actors. Understanding these systems has become an urgent scientific challenge as their capabilities advance faster than our ability to explain and guide them (Horvitz and West, 2026). The OR dial demonstrates that an entrepreneurship-relevant distinction can be located inside an AI system and shown to matter for its judgments. Its broader significance lies in moving the field from studying only what AI produces to examining how entrepreneurship-relevant distinctions are represented and used within it. Entrepreneurship theory, long used to explain human cognition, can now also interrogate the internal organisation of AI systems involved in entrepreneurial decisions. The next frontier is not simply better entrepreneurial AI, but a theory-guided account of the internal distinctions on which its judgments depend.

**Declaration of generative AI and AI-assisted technologies in the manuscript preparation process**

During the preparation of this work the authors used Anthropic Claude and OpenAI Codex to improve the language and readability of the manuscript and to assist with editing. After using these tools, the authors reviewed and edited the content as needed and take full responsibility for the content of the study. The analysis code was developed with the assistance of generative

AI tools and was reviewed, tested, and validated by the authors. The generative models used as research instruments, including the analysed model, the scenario generator, and the LLM judge, are documented with names, versions, and providers in the methods and the technical appendix.

**References**

Alain, G., & Bengio, Y. (2017). Understanding intermediate layers using linear classifier probes. *International Conference on Learning Representations (Workshop Track)*.

Alvarez, S. A., & Barney, J. B. (2007). Discovery and creation: Alternative theories of entrepreneurial action. *Strategic Entrepreneurship Journal, 1*(1–2), 11–26.

Anderson, B. S., Schueler, J., Baum, M., Wales, W. J., & Gupta, V. K. (2022). The chicken or the egg? Causal inference in entrepreneurial orientation–performance research. *Entrepreneurship Theory and Practice, 46*(6), 1569–1596.

Anthropic. (2026a). *Claude Opus 4.7 system card*. https://www.anthropic.com/claude-opus-4-7-system-card

Ardichvili, A., Cardozo, R., & Ray, S. (2003). A theory of entrepreneurial opportunity identification and development. *Journal of Business Venturing, 18*(1), 105–123.

Arditi, A., Obeso, O., Syed, A., Paleka, D., Panickssery, N., Gurnee, W., & Nanda, N. (2024). Refusal in language models is mediated by a single direction. *Advances in Neural Information Processing Systems, 37*.

Åstebro, T., & Hoos, F. (2021). Impact measurement based on repeated randomized control trials: The case of a training program to encourage social entrepreneurship. *Strategic Entrepreneurship Journal, 15*(2), 254–278.

Baron, R. A. (2006). Opportunity recognition as pattern recognition: How entrepreneurs "connect the dots" to identify new business opportunities. *Academy of Management Perspectives, 20*(1), 104–119.

Baron, R. A. (2008). The role of affect in the entrepreneurial process. *Academy of Management Review, 33*(2), 328–340.

Beaglehole, D., Radhakrishnan, A., Boix-Adserà, E., & Belkin, M. (2026). Toward universal steering and monitoring of AI models. *Science*, 391(6787), 787–792.

Belinkov, Y. (2022). Probing classifiers: Promises, shortcomings, and advances. *Computational Linguistics, 48*(1), 207–219.

Belrose, N., Schneider-Joseph, D., Ravfogel, S., Cotterell, R., Raff, E., & Biderman, S. (2023). LEACE: Perfect linear concept erasure in closed form. *Advances in Neural Information Processing Systems, 36*.

Bereska, L. F., & Gavves, E. (2024). Mechanistic interpretability for AI safety: A review. *Transactions on Machine Learning Research*.

Bricken, T., Templeton, A., Batson, J., Chen, B., Jermyn, A., Conerly, T., Turner, N., Anil, C., Denison, C., Askell, A., Lasenby, R., Wu, Y., Kravec, S., Schiefer, N., Maxwell, T., Joseph, N., Hatfield-Dodds, Z., Tamkin, A., Nguyen, K., … Olah, C. (2023). *Towards monosemanticity: Decomposing language models with dictionary learning*. Transformer Circuits Thread. https://transformer-circuits.pub/2023/monosemantic-features/index.html

Campbell, D. T., & Fiske, D. W. (1959). Convergent and discriminant validation by the multitrait-multimethod matrix. *Psychological Bulletin, 56*(2), 81–105.

Cao, X., & Kosinski, M. (2024). Large language models know how the personality of public figures is perceived by the general public. *Scientific Reports, 14*(1), Article 6735.

Chalmers, D., MacKenzie, N. G., & Carter, S. (2021). Artificial intelligence and entrepreneurship: Implications for venture creation in the fourth industrial revolution. *Entrepreneurship Theory and Practice, 45*(5), 1028–1053.

Chen, C. C., Greene, P. G., & Crick, A. (1998). Does entrepreneurial self-efficacy distinguish entrepreneurs from managers? *Journal of Business Venturing, 13*(4), 295–316.

Chen, R., Arditi, A., Sleight, H., Evans, O., & Lindsey, J. (2025). *Persona vectors: Monitoring and controlling character traits in language models*. arXiv. https://arxiv.org/abs/2507.21509

Csaszar, F. A., Ketkar, H., & Kim, H. (2024). Artificial intelligence and strategic decision-making: Evidence from entrepreneurs and investors. *Strategy Science, 9*(4), 322–345.

Davidsson, P. (2015). Entrepreneurial opportunities and the entrepreneurship nexus: A re-conceptualization. *Journal of Business Venturing, 30*(5), 674–695.

DeepSeek-AI. (2026). *DeepSeek-V4: Towards highly efficient million-token context intelligence*. arXiv. https://arxiv.org/abs/2606.19348

Dimov, D. (2011). Grappling with the unbearable elusiveness of entrepreneurial opportunities. *Entrepreneurship Theory and Practice, 35*(1), 57–81.

Dimov, D. (2026). Giving texture to theory: A computational thought experiment on opportunity capital with LLM agents. *Journal of Business Venturing Insights, 25*, Article e00635.

Eckhardt, J. T., & Shane, S. A. (2003). Opportunities and entrepreneurship. *Journal of Management, 29*(3), 333–349.

Elazar, Y., Ravfogel, S., Jacovi, A., & Goldberg, Y. (2021). Amnesic probing: Behavioral explanation with amnesic counterfactuals. *Transactions of the Association for Computational Linguistics, 9*, 160–175.

Elhage, N., Hume, T., Olsson, C., Schiefer, N., Henighan, T., Kravec, S., Hatfield-Dodds, Z., Lasenby, R., Drain, D., Chen, C., Grosse, R., McCandlish, S., Kaplan, J., Amodei, D., Wattenberg, M., & Olah, C. (2022). *Toy models of superposition*. Transformer Circuits Thread. https://transformer-circuits.pub/2022/toy_model/index.html

Ferrati, F., Kim, P. H., & Muffatto, M. (2024). Generative AI in entrepreneurship research: Principles and practical guidance for intelligence augmentation. *Foundations and Trends in Entrepreneurship, 20*(3), 245–383.

Geiger, A., Wu, Z., Potts, C., Icard, T., & Goodman, N. D. (2024). Finding alignments between interpretable causal variables and distributed neural representations. *Proceedings of the Third Conference on Causal Learning and Reasoning, 236*, 160–187.

Geirhos, R., Jacobsen, J.-H., Michaelis, C., Zemel, R., Brendel, W., Bethge, M., & Wichmann, F. A. (2020). Shortcut learning in deep neural networks. *Nature Machine Intelligence, 2*(11), 665–673.

Grattafiori, A., Dubey, A., Jauhri, A., Pandey, A., Kadian, A., Al-Dahle, A., Letman, A., Mathur, A., Schelten, A., Vaughan, A., Yang, A., Fan, A., Goyal, A., Hartshorn, A., Yang, A., Mitra, A., Sravankumar, A., Korenev, A., Hinsvark, A., … Ma, Z. (2024). *The Llama 3 herd of models*. arXiv. https://arxiv.org/abs/2407.21783

Grégoire, D. A., Barr, P. S., & Shepherd, D. A. (2010). Cognitive processes of opportunity recognition: The role of structural alignment. *Organization Science, 21*(2), 413–431.

Gurnee, W., & Tegmark, M. (2024). Language models represent space and time. *International Conference on Learning Representations*.

Gusto. (2026). *2026 new business formation report: How artificial intelligence is reshaping who starts a business in America*. https://gusto.com/resources/gusto-insights/new-business-formation-2026

Hagendorff, T., Fabi, S., & Kosinski, M. (2023). Human-like intuitive behavior and reasoning biases emerged in large language models but disappeared in ChatGPT. *Nature Computational Science, 3*(10), 833–838.

Hansen, D. J., Shrader, R., & Monllor, J. (2011). Defragmenting definitions of entrepreneurial opportunity. *Journal of Small Business Management, 49*(2), 283–304.

He, Z., Shu, W., Ge, X., Chen, L., Wang, J., Zhou, Y., Liu, F., Guo, Q., Huang, X., Wu, Z., Jiang, Y.-G., & Qiu, X. (2024). *Llama Scope: Extracting millions of features from Llama-3.1-8B with sparse autoencoders*. arXiv. https://arxiv.org/abs/2410.20526

Hewitt, J., & Liang, P. (2019). Designing and interpreting probes with control tasks. *Proceedings of the 2019 Conference on Empirical Methods in Natural Language Processing and the 9th International Joint Conference on Natural Language Processing (EMNLP-IJCNLP)*, 2733–2743.

Holm, S. (1979). A simple sequentially rejective multiple test procedure. *Scandinavian Journal of Statistics, 6*(2), 65–70.

Horvitz, E., & West, R. (2026). A narrowing window to understand AI. *Science, 392*(6802), 1003.

Jiang, Y., Rajendran, G., Ravikumar, P., Aragam, B., & Veitch, V. (2024). On the origins of linear representations in large language models. *Proceedings of the 41st International Conference on Machine Learning, 235*, 21879–21911.

Kirzner, I. M. (1997). Entrepreneurial discovery and the competitive market process: An Austrian approach. *Journal of Economic Literature, 35*(1), 60–85.

Kleinert, S., & Urbig, D. (2026). Startup evaluations with generative artificial intelligence: An exploratory study on early-stage investments and survival predictions by large language models. *Entrepreneurship Theory and Practice*. Advance online publication. https://doi.org/10.1177/10422587261430320

Kosinski, M. (2024). Evaluating large language models in theory of mind tasks. *Proceedings of the National Academy of Sciences, 121*(45), Article e2405460121.

Kuckertz, A., Kollmann, T., Krell, P., & Stöckmann, C. (2017). Understanding, differentiating, and measuring opportunity recognition and opportunity exploitation. *International Journal of Entrepreneurial Behavior & Research, 23*(1), 78–97.

Kuratko, D. F., Fisher, G., & Audretsch, D. B. (2021). Unraveling the entrepreneurial mindset. *Small Business Economics, 57*(4), 1681–1691.

Le Bras, R., Swayamdipta, S., Bhagavatula, C., Zellers, R., Peters, M. E., Sabharwal, A., & Choi, Y. (2020). Adversarial filters of dataset biases. *Proceedings of the 37th International Conference on Machine Learning, 119*, 1078–1088.

Lévesque, M., Obschonka, M., & Nambisan, S. (2022). Pursuing impactful entrepreneurship research using artificial intelligence. *Entrepreneurship Theory and Practice, 46*(4), 803–832.

Li, L., Teng, Y., Wang, Y., & Hu, X. (2026). Understanding large language models demands distinguishing human projection from machine cognition. *Communications Psychology*, 4(1), Article 108.

Lindsey, J., Gurnee, W., Ameisen, E., Chen, B., Pearce, A., Turner, N. L., Citro, C., Abrahams, D., Carter, S., Hosmer, B., Marcus, J., Sklar, M., Templeton, A., Bricken, T., McDougall, C., Cunningham, H., Henighan, T., Jermyn, A., Jones, A., … Batson, J. (2025). *On the biology of a large language model*. Transformer Circuits Thread. https://transformer-circuits.pub/2025/attribution-graphs/biology.html

Longpre, S., Akiki, C., Lund, C., Kulkarni, A., Chen, E., Solaiman, I., Ghosh, A., Jernite, Y., & Kaffee, L.-A. (2025). *Economies of Open Intelligence: Tracing power & participation in the model ecosystem*. arXiv. https://arxiv.org/abs/2512.03073

Marks, S., & Tegmark, M. (2024). The geometry of truth: Emergent linear structure in large language model representations of true/false datasets. *Conference on Language Modeling*.

McMullen, J. S., & Shepherd, D. A. (2006). Entrepreneurial action and the role of uncertainty in the theory of the entrepreneur. *Academy of Management Review, 31*(1), 132–152.

Mitchell, R. K., Busenitz, L., Lant, T., McDougall, P. P., Morse, E. A., & Smith, J. B. (2002). Toward a theory of entrepreneurial cognition: Rethinking the people side of entrepreneurship research. *Entrepreneurship Theory and Practice, 27*(2), 93–104.

Mueller, A., Brinkmann, J., Li, M., Marks, S., Pal, K., Prakash, N., Rager, C., Sankaranarayanan, A., Sen Sharma, A., Sun, J., Todd, E., Bau, D., & Belinkov, Y. (2026). The quest for the right mediator: Surveying mechanistic interpretability for NLP through the lens of causal mediation analysis. *Computational Linguistics*, 52(1), 331–378.

Nanda, N., & Bloom, J. (2022). *TransformerLens* [Computer software]. GitHub. https://github.com/TransformerLensOrg/TransformerLens

Obschonka, M., & Fisch, C. (2025). From humans to machines: Researching entrepreneurial AI agents built on large language models. *Journal of Business Venturing Insights, 24*, Article e00581.

Obschonka, M., Grégoire, D. A., Nikolaev, B., Ooms, F., Lévesque, M., Pollack, J. M., & Behrend, T. S. (2025). Artificial intelligence and entrepreneurship: A call for research to prospect and establish the scholarly AI frontiers. *Entrepreneurship Theory and Practice, 49*(3), 620–641.

Ooms, F., Annen, J., Panda, R., Meunier, P., Tshibanda, L., Laureys, S., Pollack, J. M., & Surlemont, B. (2024). Advancing (neuro)entrepreneurship cognition research through resting-state fMRI: A methodological brief. *Entrepreneurship Theory and Practice, 48*(2), 719–741.

Park, K., Choe, Y. J., & Veitch, V. (2024). The linear representation hypothesis and the geometry of large language models. *Proceedings of the 41st International Conference on Machine Learning, 235*, 39643–39666.

Pellert, M., Lechner, C. M., Wagner, C., Rammstedt, B., & Strohmaier, M. (2024). AI psychometrics: Assessing the psychological profiles of large language models through psychometric inventories. *Perspectives on Psychological Science, 19*(5), 808–826.

Ramoglou, S., Chandra, Y., & Jin, S. Q. (2026). Opportunity search in the era of GenAI: Navigating uncertainty in an expanding universe of imaginable but unknowable futures. *Journal of Management Studies, 63*(2), 695–721.

Ramoglou, S., Schaefer, R., Chandra, Y., & McMullen, J. S. (2025). Artificial intelligence forces us to rethink Knightian uncertainty: A commentary on Townsend et al.'s "Are the futures computable?" *Academy of Management Review, 50*(2), 471–473.

Ramoglou, S., & Tsang, E. W. K. (2016). A realist perspective of entrepreneurship: Opportunities as propensities. *Academy of Management Review, 41*(3), 410–434.

Rimsky, N., Gabrieli, N., Schulz, J., Tong, M., Hubinger, E., & Turner, A. M. (2024). Steering Llama 2 via contrastive activation addition. *Proceedings of the 62nd Annual Meeting of the Association for Computational Linguistics (Volume 1: Long Papers)*, 15504–15522.

Rodriguez, P., Blaas, A., Klein, M., Zappella, L., Apostoloff, N., Cuturi, M., & Suau, X. (2025). Controlling language and diffusion models by transporting activations. *International Conference on Learning Representations*.

Sarasvathy, S. D. (2001). Causation and effectuation: Toward a theoretical shift from economic inevitability to entrepreneurial contingency. *Academy of Management Review, 26*(2), 243–263.

Shanahan, M., McDonell, K., & Reynolds, L. (2023). Role play with large language models. *Nature, 623*(7987), 493–498.

Shane, S., & Venkataraman, S. (2000). The promise of entrepreneurship as a field of research. *Academy of Management Review, 25*(1), 217–226.

Shapero, A., & Sokol, L. (1982). The social dimensions of entrepreneurship. In C. A. Kent, D. L. Sexton, & K. H. Vesper (Eds.), *Encyclopedia of entrepreneurship* (pp. 72–90). Prentice-Hall.

Shepherd, D. A., & Majchrzak, A. (2022). Machines augmenting entrepreneurs: Opportunities (and threats) at the nexus of artificial intelligence and entrepreneurship. *Journal of Business Venturing, 37*(4), Article 106227.

Shiffrin, R., & Mitchell, M. (2023). Probing the psychology of AI models. *Proceedings of the National Academy of Sciences, 120*(10), Article e2300963120.

Short, C. E., & Short, J. C. (2023). The artificially intelligent entrepreneur: ChatGPT, prompt engineering, and entrepreneurial rhetoric creation. *Journal of Business Venturing Insights, 19*, Article e00388.

Short, J. C., Ketchen, D. J., Jr., Shook, C. L., & Ireland, R. D. (2010). The concept of "opportunity" in entrepreneurship research: Past accomplishments and future challenges. *Journal of Management, 36*(1), 40–65.

Stolfo, A., Balachandran, V., Yousefi, S., Horvitz, E., & Nushi, B. (2025). Improving instruction-following in language models through activation steering. *International Conference on Learning Representations*.

Suddaby, R. (2010). Editor's comments: Construct clarity in theories of management and organization. *Academy of Management Review, 35*(3), 346–357.

Tan, D., Chanin, D., Lynch, A., Paige, B., Kanoulas, D., Garriga-Alonso, A., & Kirk, R. (2024). Analysing the generalisation and reliability of steering vectors. *Advances in Neural Information Processing Systems, 37*.

Tang, J., Kacmar, K. M., & Busenitz, L. (2012). Entrepreneurial alertness in the pursuit of new opportunities. *Journal of Business Venturing, 27*(1), 77–94.

Templeton, A., Conerly, T., Marcus, J., Lindsey, J., Bricken, T., Chen, B., Pearce, A., Citro, C., Ameisen, E., Jones, A., Cunningham, H., Turner, N. L., McDougall, C., MacDiarmid, M., Tamkin, A., Durmus, E., Hume, T., Mosconi, F., Freeman, C. D., … Henighan, T. (2024). *Scaling monosemanticity: Extracting interpretable features from Claude 3 Sonnet*. Transformer Circuits Thread. https://transformer-circuits.pub/2024/scaling-monosemanticity/

Tigges, C., Hollinsworth, O. J., Geiger, A., & Nanda, N. (2024). Language models linearly represent sentiment. *Proceedings of the 7th BlackboxNLP Workshop: Analyzing and Interpreting Neural Networks for NLP*, 58–87.

Townsend, D. M., & Hunt, R. A. (2019). Entrepreneurial action, creativity, & judgment in the age of artificial intelligence. *Journal of Business Venturing Insights, 11*, Article e00126.

Townsend, D. M., Hunt, R. A., Rady, J., Manocha, P., & Jin, J. H. (2025). Are the futures computable? Knightian uncertainty and artificial intelligence. *Academy of Management Review, 50*(2), 415–440.

Turner, A. M., Thiergart, L., Leech, G., Udell, D., Vazquez, J. J., Mini, U., & MacDiarmid, M. (2023). *Steering language models with activation engineering*. arXiv. https://arxiv.org/abs/2308.10248

Tzabbar, D. (2026). Why theory matters for causal inference? Rethinking endogeneity in entrepreneurship research. *Strategic Entrepreneurship Journal*. Advance online publication. https://doi.org/10.1002/sej.70033

Wang, T., Jiao, X., Zhu, Y., Chen, Z., He, Y., Chu, X., Gao, J., Wang, Y., & Ma, L. (2025). Adaptive activation steering: A tuning-free LLM truthfulness improvement method for diverse hallucinations categories. *Proceedings of the ACM on Web Conference 2025*, 2562–2578.

Wang, Y., Kordi, Y., Mishra, S., Liu, A., Smith, N. A., Khashabi, D., & Hajishirzi, H. (2023). Self-instruct: Aligning language models with self-generated instructions. *Proceedings of the 61st Annual Meeting of the Association for Computational Linguistics (Volume 1: Long Papers)*, 13484–13508.

Wehner, J., Abdelnabi, S., Tan, D., Krueger, D., & Fritz, M. (2025). Taxonomy, opportunities, and challenges of representation engineering for large language models. *Transactions on Machine Learning Research*.

Wood, M. S., & McKelvie, A. (2015). Opportunity evaluation as future focused cognition: Identifying conceptual themes and empirical trends. *International Journal of Management Reviews, 17*(2), 256–277.

Wu, Z., Arora, A., Geiger, A., Wang, Z., Huang, J., Jurafsky, D., Manning, C. D., & Potts, C. (2025). AxBench: Steering LLMs? Even simple baselines outperform sparse autoencoders. *Proceedings of the 42nd International Conference on Machine Learning, 267*, 67035–67080.

Xu, N., Zhang, Q., Du, C., Luo, Q., Qiu, X., Huang, X., & Zhang, M. (2025). Revealing emergent human-like conceptual representations from language prediction. *Proceedings of the National Academy of Sciences*, 122(44), Article e2512514122.

Zheng, L., Chiang, W.-L., Sheng, Y., Zhuang, S., Wu, Z., Zhuang, Y., Lin, Z., Li, Z., Li, D., Xing, E. P., Zhang, H., Gonzalez, J. E., & Stoica, I. (2023). Judging LLM-as-a-judge with MT-Bench and Chatbot Arena. *Advances in Neural Information Processing Systems, 36*.

Zou, A., Phan, L., Chen, S., Campbell, J., Guo, P., Ren, R., Pan, A., Yin, X., Mazeika, M., Dombrowski, A.-K., Goel, S., Li, N., Byun, M. J., Wang, Z., Mallen, A., Basart, S., Koyejo, S., Song, D., Fredrikson, M., … Hendrycks, D. (2023). *Representation engineering: A top-down approach to AI transparency*. arXiv. https://arxiv.org/abs/2310.01405

## Figures

**Figure 1. From external inference to internal intervention: Mechanistic interpretability as a new route to studying entrepreneurial cognition.**

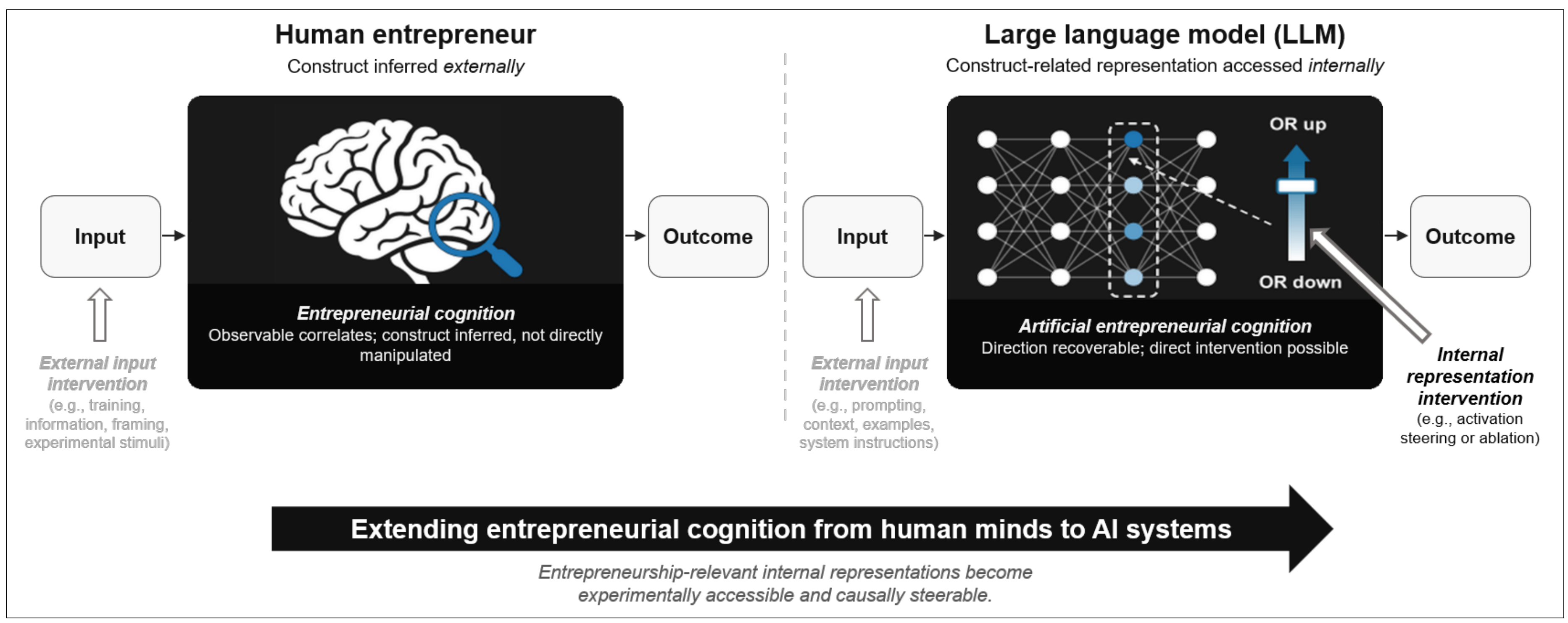

*Notes*: This figure contrasts two routes of empirical access to entrepreneurial cognition, illustrated through opportunity recognition (OR). The left panel depicts the input-side route typically used in research on human entrepreneurs. Because a construct-specific internal state cannot ordinarily be located and manipulated directly, researchers manipulate the information presented to participants, for example through vignettes, conjoint tasks, or framing manipulations. They then observe what participants report or do and infer the cognitive processes underlying those outcomes. The right panel depicts the additional internal route available in open-weight large language models (LLMs). Because internal activations are accessible, researchers can locate an OR direction in the model's activation space (what we refer to as the OR dial), adjust it at a chosen strength during inference through activation steering, or remove it through ablation. They can then examine whether the model's assessments change in theoretically predicted ways. The comparison concerns empirical access only. It does not equate human cognition with AI systems, nor does it imply that the model understands opportunities or possesses entrepreneurial agency.

**Figure 2. Propositions of the structural theory of artificial entrepreneurial cognition.**

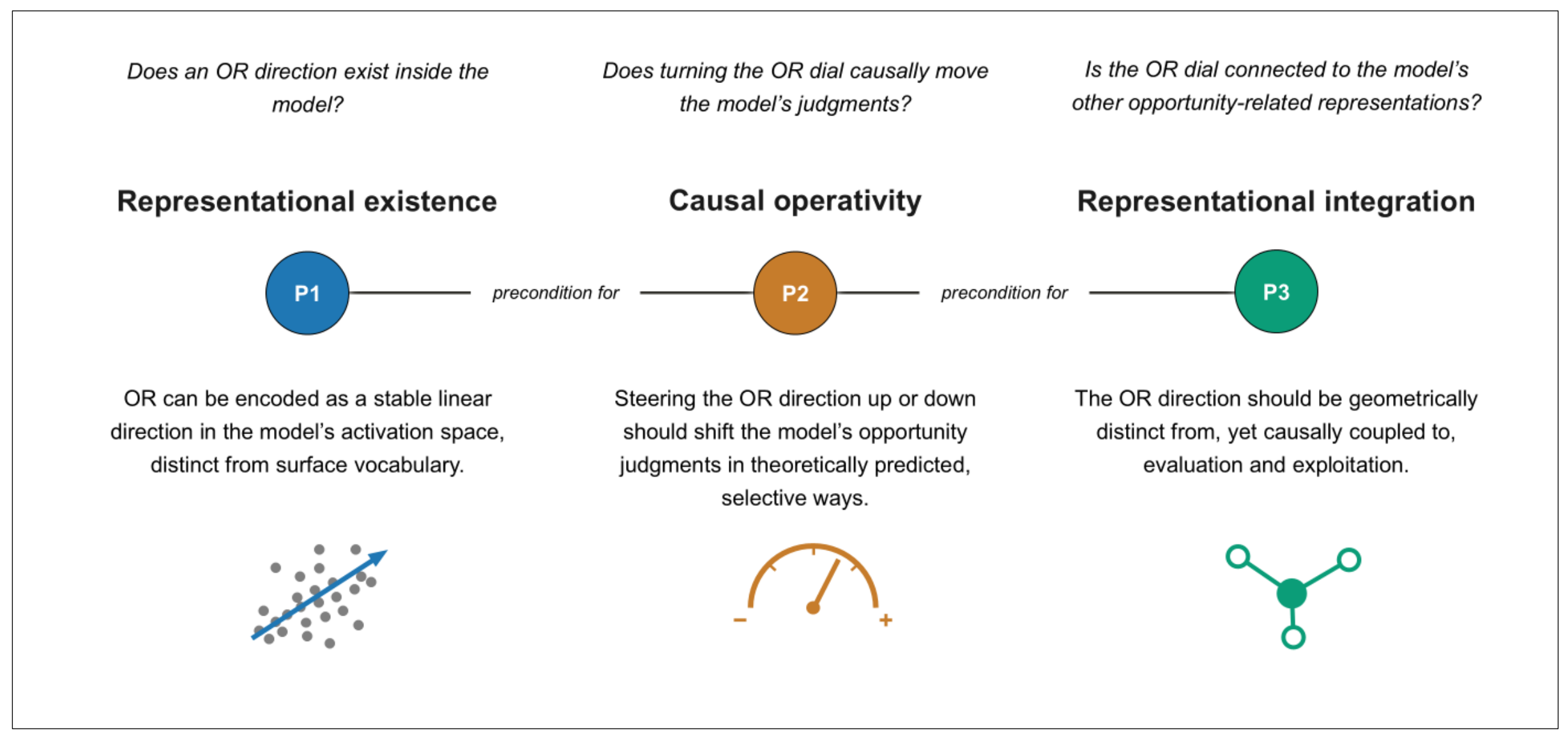

*Notes*: The upper row states each proposition as a question about the opportunity recognition (OR) direction or dial. The lower rows state the proposition labels, their relationship, and their application to OR.

**Figure 3. Held-out probe accuracy by layer.**

LR 0.839
DiM 0.748
Held-out test accuracy
Layer
LR probe
DiM probe

*Notes*: Each point shows held-out accuracy of the difference-in-means (DiM) or logistic regression (LR) probe at a given layer. The layer is selected using 200-split cross-validation on the training pairs only. The test pairs remain a lockbox, a held-out set not used for layer selection or direction estimation. The curve displays lockbox accuracy at every layer for transparency, but only the layer 12 value serves as the headline estimate, and no selection decision uses these values.

**Figure 4. Dose-response of OR steering by signed steering coefficient.**

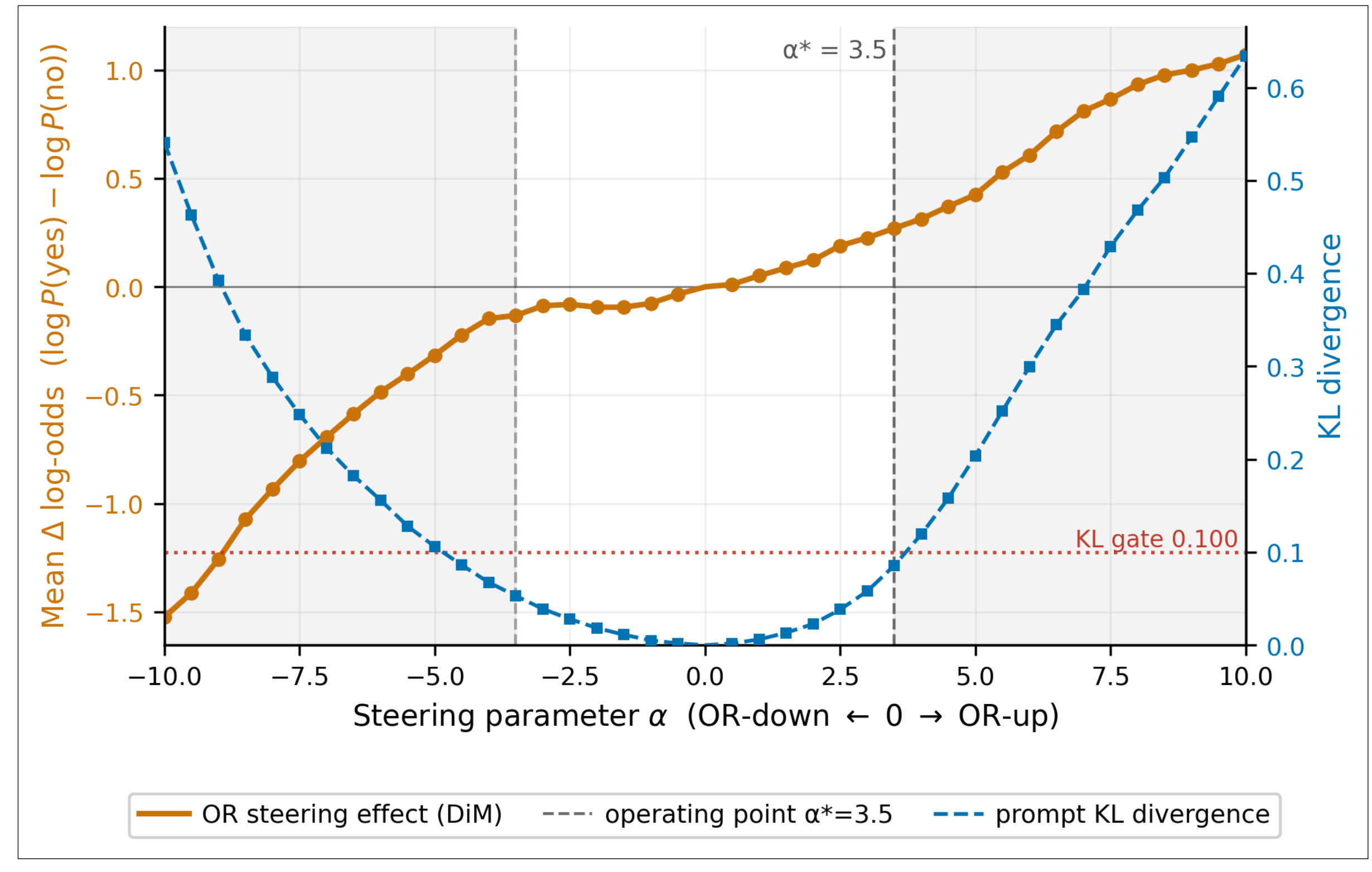

*Notes:* This figure shows the mean change in forced-choice decision log-odds, defined as log P(yes) minus log P(no), as a function of the signed steering coefficient α. Positive α steers toward OR-present responses (OR dial turned up) and negative α steers toward OR-absent responses (OR dial turned down). The dashed curve is the coherence cost, measured as prompt-final KL divergence between the steered and unsteered model, with the predefined 0.100 gate marked. The vertical dashed lines mark the operating coefficient of 3.5, the largest magnitude whose KL stays below the gate, and the shaded regions lie beyond it. The effect and KL curves are computed on 56 band-balanced evaluation scenarios, while the coefficient α=3.5 is selected separately on 35 held-out calibration scenarios (Appendix D.1).

**Figure 5. Cosine alignment of the OR direction with adjacent construct directions.**

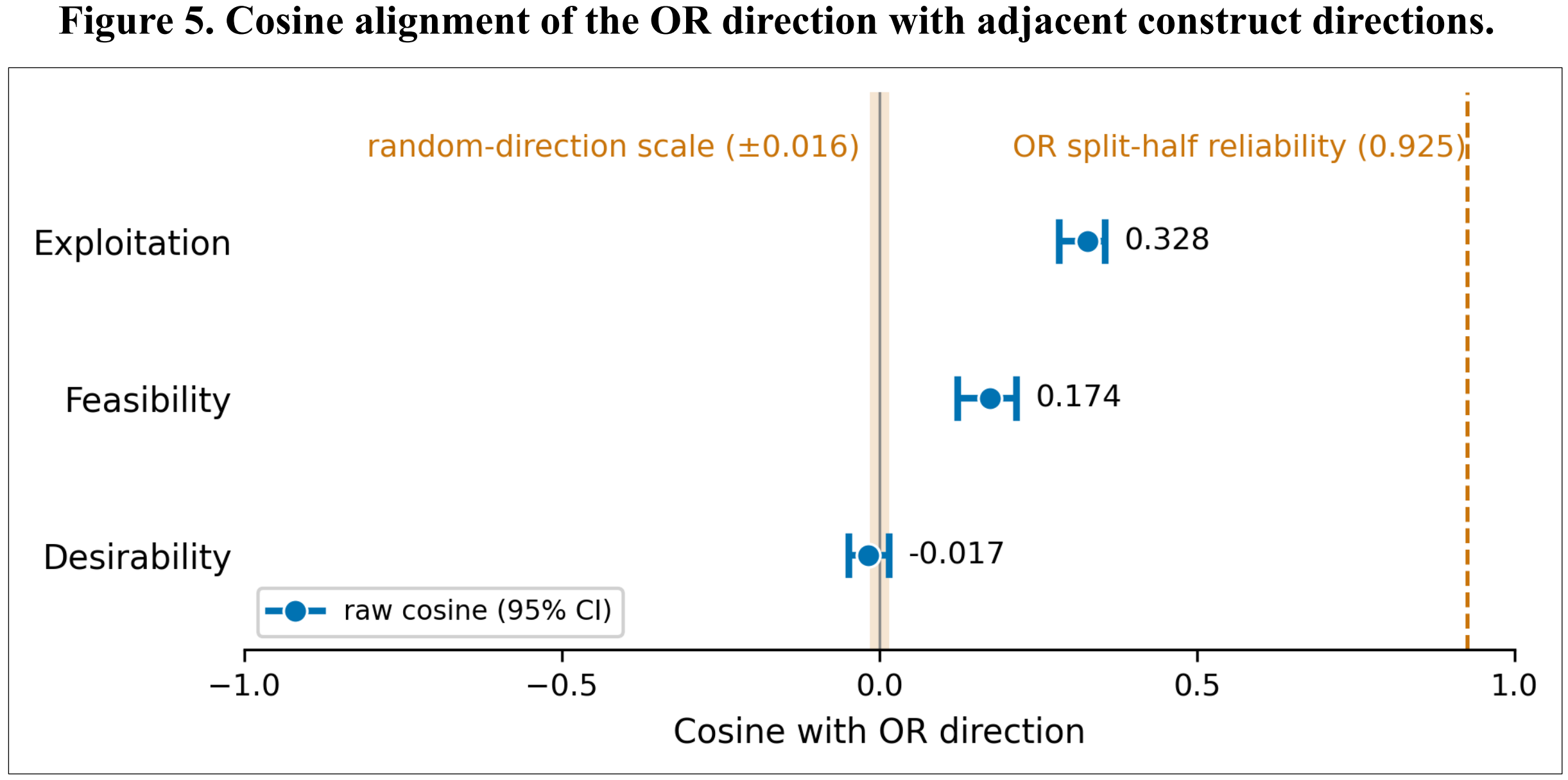

Notes: This figure illustrates the cosine similarity between the OR direction and adjacent construct directions (feasibility, desirability, exploitation), estimated by difference-in-means at layer 12, with 95% bootstrap confidence intervals. The shaded band marks the random-direction scale of plus or minus 0.016, the analytic standard deviation of the cosine of an arbitrary direction in a 4,096-dimensional space, centred on zero. The dashed line marks the OR split-half reference of 0.925.

## Tables

### Table 1. Overview of our pipeline for locating, steering, and relating the OR direction.

| Part | Section | Step | Description | Key result |
|---|---|---|---|---|
| **Part 1: Locating the OR direction**<br><br>*Tests Proposition 1: representational existence* | 3.1.1 | Scenario design | Construct matched OR-present/OR-absent scenario pairs using Claude Opus 4.7, then apply vocabulary-, pair-, and batch-level rules, deterministic validation, and adversarial filtering. | 636 contrastive pairs remain. Bag-of-words classifier accuracy falls from 0.912 to 0.528 (0.500 chance). |
| | 3.1.2 | Activation extraction | Run each scenario through Llama 3.1 8B-Instruct and record residual-stream activations across layers. | Activations extracted across 32 layers for 509 training/127 held-out pairs. |
| | 3.1.3 | Direction estimation and validation | Estimate an OR direction separating OR-present from OR-absent activations. Search layers and read positions on training data, then validate on held-out pairs and ablation controls. | Layer 12, read position -5. Held-out DiM 0.748, LR 0.839. Direction ablation returns the re-trained LR probe to chance. |
| **Part 2: Steering the OR direction**<br><br>*Tests Proposition 2: causal operativity* | 3.2.1 | Activation steering procedure | Add or subtract a scaled copy of the OR direction at the operating layer during inference. Select the coefficient α using a coherence gate. | Steering coefficient $\alpha=3.5$. |
| | 3.2.2 | Forced-choice readout | Append a yes/no recognition question and read steering-induced changes in log-odds. | On entrepreneurial scenarios, OR-up raises the recognition readout by 0.401 log-odds and OR-down lowers it by 0.716. Directional ablation lowers the recognition readout. |
| | 3.2.3 | Free-text readout | Generate open-form explanations, then use an LLM judge to assess which of the OR-up and OR-down explanations more clearly recognises an opportunity (paired comparison). | Judge prefers OR-up in 0.656 of the entrepreneurial pairs (both-order robustness 0.708). |
| **Part 3: Relating the OR direction to adjacent constructs**<br><br>*Tests Proposition 3: representational integration* | 3.3.1 | Downstream coupling | Steer the OR direction and test whether feasibility, desirability, and exploitation readouts shift with it. | All three downstream readouts rise under OR-up and fall under OR-down. |
| | 3.3.2 | Discriminant geometry | Estimate separate directions for feasibility, desirability, and exploitation, then compare their geometric alignment with the OR direction. | OR direction is distinct from adjacent directions. Their joint span accounts for 0.116 of the OR direction. |

*Notes*: OR=opportunity recognition, LLM=large language model, DiM=difference-in-means, LR=logistic regression. The pipeline is organised in three parts, each testing one proposition of the structural theory of artificial entrepreneurial cognition. OR-up and OR-down denote steering the model along and against the OR direction. Log-odds values are the steering-induced change in the model's log-odds of answering "yes" versus "no". Full procedures and additional diagnostics are reported in the corresponding subsections and appendices.

## Table 2. Example contrastive pairs from our main battery of scenarios.

| ID | OR-present scenario | OR-absent scenario | Words | OR-absent interpretation |
|---|---|---|---|---|
| 8 | Every Saturday at a regional farmers' market, a pickle maker tracks recurring comments from visitors about their grandparents' preservation methods. He drafts a short proposal for a monthly tasting evening where attendees could bring family recipes to be jarred on-site, paying a modest fee for ingredients and his guidance. | Every Saturday at a regional farmers' market, a pickle maker tracks recurring comments from visitors about their grandparents' preservation methods. He drafts a small chalk sign above the counter where attendees could read three quotes he has collected, displayed alongside the seasonal varieties and rotated every other month. | 49/48 | Absorbs the recurring signal into static display content rather than a generalisable participatory service. |
| 7 | Across multiple vehicles at a regional bus depot, the same brake-line component fails on cold mornings, all from one model year. The maintenance technician sketches a workshop modification kit using off-the-shelf fittings, drafts a depot-level installation procedure with parts diagrams, and approaches the regional operator about piloting the kit on a small fleet sample. | Across multiple vehicles at a regional bus depot, the same brake-line component fails on cold mornings, all from one model year. The maintenance technician sketches a daily inspection checklist using the existing toolset, drafts a depot-level reporting form with failure-history columns, and approaches the regional operator about quicker spare-part deliveries during winter cold spells. | 54/54 | Treats the recurring failure as a monitoring-and-spares problem rather than a portable fix others could pilot. |
| 19 | During monthly critiques, the club secretary observes members repeatedly struggling to print large-format work at affordable prices. He converts a corner of his garage into a small inkjet station, calibrates it carefully, and circulates a weekend printing schedule that asks five euros per print to cover paper, ink, and a modest reserve for replacement heads. | During monthly critiques, the club secretary observes members repeatedly struggling to print large-format work at affordable prices. He converts a corner of his garage into a small inkjet station, calibrates it carefully, and circulates a weekend printing schedule that lets members reserve free slots whenever the club treasury covers the consumables that month. | 55/53 | Frames the recurring signal as shared club resourcing rather than a generalisable basis for a priced service. |

*Notes*: This table lists three example pairs from the main battery (n=636 pairs), reproduced verbatim. The ID column gives each pair's identifier in the released battery. The column "Words" reports the word counts of the OR-present and OR-absent scenarios, and the column "OR-absent interpretation" paraphrases the failure-mode label stored with that pair in the battery. In each pair, the two scenarios share actor, setting, signal, and activity, but differ in how the actor interprets the recurring signal.

## Table 3. Forced-choice readout under OR steering.

| | | Baseline | OR-up steering | OR-down steering |
|---|---|---|---|---|
| **Scenarios** | **n** | **Log-odds (P(yes))** | **ΔLog-odds [95% CI] (P(yes))** | **ΔLog-odds [95% CI] (P(yes))** |
| **Entrepreneurial scenarios** | **285** | **3.359 (0.891)** | **0.401 [0.363, 0.437] (0.924)** | **-0.716 [-0.752, -0.681] (0.853)** |
| Weak opportunity | 95 | 2.105 (0.814) | 0.530 [0.488, 0.570] (0.867) | -0.542 [-0.589, -0.489] (0.763) |
| Medium opportunity | 95 | 2.621 (0.873) | 0.547 [0.500, 0.591] (0.915) | -0.616 [-0.668, -0.562] (0.826) |
| Strong opportunity | 95 | 5.351 (0.986) | 0.125 [0.060, 0.188] (0.990) | -0.989 [-1.022, -0.953] (0.971) |
| **Everyday scenarios** | **380** | **-7.229 (0.004)** | **0.322 [0.290, 0.356] (0.007)** | **0.412 [0.387, 0.436] (0.005)** |
| Natural observation | 95 | -8.359 (0.000) | 0.290 [0.243, 0.338] (0.000) | 0.521 [0.480, 0.564] (0.000) |
| Domestic routine | 95 | -6.995 (0.002) | 0.178 [0.120, 0.230] (0.003) | 0.402 [0.359, 0.448] (0.003) |
| Interpersonal moment | 95 | -6.694 (0.003) | 0.250 [0.185, 0.315] (0.004) | 0.341 [0.297, 0.385] (0.003) |
| Everyday (one sentence) | 95 | -6.867 (0.011) | 0.570 [0.508, 0.628] (0.020) | 0.384 [0.328, 0.439] (0.012) |

*Notes*: Llama 3.1 8B-Instruct, layer 12, α=3.5. Forced-choice readout on the main OR question ("Does the scenario describe an entrepreneurial opportunity? Answer yes or no."), read deterministically from the model's next-token probabilities. Steering columns report changes in log-odds of "yes" versus "no", with 95% bootstrap confidence intervals (CIs). Parenthetical P(yes) values are the mean probability of "yes" in each condition. Each band contains 95 scenarios after excluding calibration scenarios. At this KL-gated dose, the modal answer flips in the intended direction for 5 of 285 entrepreneurial scenarios under OR-up and 4 under OR-down, and for 0 of 380 everyday scenarios.

## Table 4a. Illustrative free-text explanations under OR steering.

| Band | Scenario | Baseline response | OR-up steering response | OR-down steering response |
|---|---|---|---|---|
| Weak | Invoices arriving at a single-shop florist this season show the wholesale price of locally grown dahlias dropping by twelve percent compared with last summer. The shop sources from three suppliers and sells weekly arrangements priced between twenty and forty euros. Sales volumes have remained flat for the past four months, with no shifts in customer composition reported by staff.<br><br>Does the scenario describe an entrepreneurial opportunity? Explain your reasoning in about 3 sentences. | The scenario describes a potential entrepreneurial opportunity because the drop in wholesale prices of locally grown dahlias creates a cost advantage for the florist, which could be passed on to customers to increase sales or used to improve profit margins. However, the fact that sales volumes have remained flat suggests that the shop may not be effectively utilizing this cost advantage, creating a gap in the market that an entrepreneurial venture could exploit. | The scenario describes an entrepreneurial opportunity because the drop in wholesale prices of locally grown dahlias creates a potential cost savings for the florist, which could be passed on to customers or used to increase profit margins. With sales volumes remaining flat, the florist may be able to capitalize on this opportunity by adjusting their pricing strategy or product offerings to stay competitive and attract new customers. This scenario presents a classic example of a supply-side shock that can create a business opportunity for entrepreneurs who are agile and responsive to market changes. | The scenario does not explicitly describe an entrepreneurial opportunity. However, it hints at a potential opportunity for the florist to capitalize on the price drop of dahlias, as they can now purchase them at a lower wholesale price. This could allow the shop to maintain or even increase profit margins, but it does not necessarily create a new opportunity for growth or innovation. |
| Medium | A bakery owner in a coastal town observes that several customers have asked over the past three weeks whether the shop stocks sourdough loaves made without wheat. The bakery currently sells only standard wheat-based bread, and the owner has not tracked these inquiries formally. Foot traffic has stayed level, and the nearest specialty grocery sits eleven kilometres inland.<br><br>Does the scenario describe an entrepreneurial opportunity? Explain your reasoning in about 3 sentences. | The scenario describes a potential entrepreneurial opportunity because there is a demonstrated customer demand for a specific product (sourdough loaves made without wheat) that the bakery does not currently offer. The fact that customers are inquiring about this product despite the bakery's standard offerings suggests that there is a gap in the market that could be filled by introducing a new product line. The bakery's coastal location and the lack of a nearby specialty grocery store also create a unique opportunity for the bakery to cater to this demand. | The scenario describes a potential entrepreneurial opportunity because the bakery owner has identified a specific customer demand (sourdough loaves made without wheat) that is not currently being met by the existing business. This demand is likely driven by a growing trend towards gluten-free and specialty breads, and the bakery's coastal location may provide a unique advantage in catering to this market. By capitalizing on this opportunity, the bakery owner can differentiate their business and potentially attract new customers. | The scenario describes a potential entrepreneurial opportunity because the bakery owner has noticed a specific customer demand for a product that is not currently offered (sourdough loaves made without wheat). This demand, combined with the fact that the nearest specialty grocery is located eleven kilometers inland, suggests that customers are willing to travel to the bakery to purchase standard wheat-based bread, and may be even more likely to do so if the bakery offered the product they are looking for. However, it's worth noting that the owner has not formally tracked these inquiries, which may indicate that the demand is not significant enough to warrant a new product offering. |
| Strong | At a community swim meet, a coach observes that parents repeatedly ask where to buy the silicone caps the team uses, after struggling to find them locally. Three different parents inquire during a single afternoon session. The team currently orders caps in bulk from an overseas supplier at low per-unit cost, and no nearby sporting-goods retailer stocks the specific brand.<br><br>Does the scenario describe an entrepreneurial opportunity? Explain your reasoning in about 3 sentences. | Yes, the scenario describes an entrepreneurial opportunity. The demand for the silicone caps is evident, as three parents inquired about purchasing them during a single session, and the team's current supplier is overseas, indicating a potential gap in the local market. An entrepreneur could capitalize on this demand by sourcing the caps from a local supplier or even manufacturing them locally, potentially offering a more convenient and accessible option for customers. | Yes, the scenario describes an entrepreneurial opportunity. The coach has identified a demand for a specific product (silicone swim caps) that is not readily available locally, indicating a potential market gap. By filling this gap, the coach could create a business that supplies the caps to the swim team and potentially other customers, such as local swim clubs, schools, or even individual swimmers. | The scenario describes a potential entrepreneurial opportunity because there is a demand for a specific product (the silicone caps) that is not readily available in the local market. The fact that parents are inquiring about where to buy these caps suggests that they are willing to purchase them, but the current lack of availability is preventing them from doing so. However, it's worth noting that the opportunity may be limited to a small geographic area, and the demand may not be significant enough to justify a full-fledged business venture. |

*Notes*: Llama 3.1 8B-Instruct, layer 12, α=3.5. Responses are generated with greedy (deterministic) decoding and reproduced verbatim.

## Table 4b. Free-text readout under OR steering: LLM-judge ratings.

| | Single presentation | | | Order-doubled robustness | | |
|---|---|---|---|---|---|---|
| **Scenarios** | **Win/loss** | **Win rate** | **vs. chance** | **Win/tie/loss** | **Win rate** | **vs. chance** |
| **Entrepreneurial scenarios** | **187/98** | **0.656** | **<0.001** | **143/83/59** | **0.708** | **<0.001** |
| Weak opportunity | 67/28 | 0.705 | <0.001 | 58/23/14 | 0.806 | <0.001 |
| Medium opportunity | 67/28 | 0.705 | <0.001 | 47/34/14 | 0.770 | <0.001 |
| Strong opportunity | 53/42 | 0.558 | 0.305 | 38/26/31 | 0.551 | 0.470 |

*Notes:* An LLM judge (DeepSeek V4 Pro, reasoning enabled) sees each scenario and the paired OR-up and OR-down explanations to the main OR question and picks the one that more clearly recognises an entrepreneurial opportunity (285 scenarios, 95 per band). In the single-presentation protocol, the judge sees each pair once, with the OR-up response randomly assigned to the left or right side. In the order-doubled protocol (Zheng et al., 2023), each pair is judged in both presentation orders. A win counts only when the same explanation wins in both, inconsistent verdicts count as ties (consistency 0.709). Win rates are tested against 0.500 using an exact two-sided binomial test.

## Table 5. Downstream-coupling readout under OR steering for entrepreneurial scenarios.

| | | Baseline | OR-up steering | OR-down steering |
|---|---|---|---|---|
| **Construct** | **Question: Does the scenario describe …** | **Log-odds (P(yes))** | **ΔLog-odds (ΔP(yes))** | **ΔLog-odds (ΔP(yes))** |
| Recognition | … an entrepreneurial opportunity? | 3.359 (0.891) | 0.401 (0.033) | -0.716 (-0.038) |
| Feasibility | … a situation where the person could take some practical action, regardless of whether it would be worthwhile? | 4.114 (0.978) | 0.728 (0.011) | -0.456 (-0.011) |
| Desirability | … something that would be worth the cost and effort for the person to pursue? | 0.809 (0.649) | 0.338 (0.057) | -0.331 (-0.055) |
| Exploitation | … something the person should actively pursue now? | 1.496 (0.748) | 0.483 (0.059) | -0.421 (-0.054) |

*Notes*: Same deterministic forced-choice readout as in Table 3 (layer 12, α=3.5), on the 285 entrepreneurial scenarios. Recognition is the main OR readout from Table 3, and provides the within-construct reference for the three downstream readouts. The downstream questions do not enter OR direction estimation, α calibration, or layer selection. ΔP(yes) is shown because it is interpretable, but when baselines are near the probability ceiling, as for feasibility, log-odds provide the more sensitive readout. At the selected coefficient, and computed across all 665 evaluation scenarios rather than the 285 entrepreneurial scenarios shown here, the share moving against the intended steer under OR-up is 0.116 (recognition), 0.030 (feasibility), 0.075 (desirability), and 0.492 (exploitation). The exploitation share reflects the everyday scenarios, whose exploitation readout falls under OR-up.

## Table 6. Summary of robustness checks.

| # | Concern | Method | Key result | Interpretation and boundary |
|---|---|---|---|---|
| 1 | Surface form and register | Apply directions estimated from the main battery to shortened scenarios, negation-based contrasts, and second-person scenarios (e.g., Marks and Tegmark, 2024; Rimsky et al., 2024; Tigges et al., 2024). | DiM accuracy is 0.781, 0.841, and 0.860 across the three variations. LR accuracy is 0.835, 0.900, and 0.900. Directional ablation returns the re-trained LR probe to 0.500 in each set, by construction. | The direction transfers beyond the surface form and register of the original battery. The variation sets are not AFLite-filtered, and shared semantic confounds remain possible. |
| 2 | Commerciality as a rival explanation | Test recovery on commerce-matched pairs, remove linearly decodable commerciality using LEACE, and estimate an independent commerciality direction (Belrose et al., 2023; Campbell and Fiske, 1959; Davidsson, 2015). | DiM accuracy remains 0.750 when commerciality provides no distinguishing information. Removing linearly decodable commerciality on the fitting sample reduces LR accuracy only from 0.839 to 0.811, a drop that is a lower bound because the erasure is incomplete on held-out pairs. OR and commerciality directions have whitened cosine of 0.000 and produce different causal effects. | The OR direction is not reducible to commerciality. The tests operationalise commerciality at different levels. |
| 3 | Generator-specific wording or style | Generate an independent battery with DeepSeek and test cross-generator probe transfer and direction alignment. | A Claude-trained probe classifies DeepSeek scenarios with 0.832 accuracy, and a DeepSeek-trained probe classifies Claude scenarios with 0.783 accuracy. The directions align at a cosine of 0.799. | Recovery is unlikely to depend on stylistic regularities specific to one generator. Artefacts shared by the common scenario specification remain possible. |
| 4 | Model scale or family | Repeat recovery, ablation, steering, and geometric analyses in Llama 3.1 70B-Instruct, Qwen3-8B, Qwen2.5-32B-Instruct, and Gemma-3-12B-it. | The replication models recover OR directions with held-out DiM accuracy between 0.705 and 0.744. Ablation returns the re-trained LR probe to 0.500 by construction, steering shifts the OR readouts as predicted, and OR remains geometrically distinct from feasibility, desirability, and exploitation. | Core patterns extend across tested scales and model families. The additional runs omit the random-direction control, and the headline model's clean null-tier separation reappears fully only in the two largest additional models, weaker in Gemma-3-12B-it and reversed in Qwen3-8B. |
| 5 | Free-text judge bias and intervention placement | Re-judge all OR-up and OR-down explanations in both presentation orders and repeat steering at every token position. | The order-doubled OR-up win rate is 0.708 on the entrepreneurial scenarios, with judge consistency of 0.709. All-position steering raises the order-doubled win rate to 0.971 but produces mean next-token KL divergence of 0.671. | The free-text result is not explained by simple presentation-order bias. All-position steering provides a robustness ceiling but exceeds the coherence gate and is not interpreted as the calibrated effect. |

*Notes:* LEACE=linear concept erasure (Belrose et al., 2023). Detailed procedures, estimates, confidence intervals, and diagnostic analyses are reported in Appendix F.

## Table 7. Summary of additional analyses.

| Analysis | Purpose and method | Key result | Interpretation and boundary |
|---|---|---|---|
| Steering versus role prompting | Compare OR-up and OR-down steering with entrepreneur-persona prompting on the forced-choice readout. Combine the persona prompt with each steering direction to compare the breadth and reversibility of the two interventions. | The entrepreneur persona produces larger and broader changes across most opportunity bands. Steering produces smaller, band-sensitive shifts. OR-down steering lowers OR readouts and partially offsets the persona effect. | Steering and role prompting are distinct intervention levers. Steering is narrower and reversible in sign, but prompt strength and steering coefficients have no common scale, so the comparison is convergent evidence only. |
| Survey-style construct profile | Compare OR steering and entrepreneur, high-OR, and low-OR persona prompts across measures of OR, exploitation, alertness, self-efficacy, proactive personality, bricolage, personality, and off-construct attributes. | The OR scale rises from 0.652 at baseline to 0.718 under OR-up steering and falls to 0.625 under OR-down steering. Exploitation and theoretically related profiles move in the same direction, while most off-construct measures move little. Persona prompts produce broader changes, including movement in the ADHD self-report. | The profile provides descriptive convergent evidence about the content affected by steering. It is not psychometric validation, and results at twice the selected coefficient are descriptive because that intervention exceeds the coherence gate. |
| Sparse autoencoder (SAE) feature alignment | Rank features from a pretrained sparse autoencoder (Llama Scope) by alignment with the OR direction and inspect the most aligned feature labels. | The strongest single-feature alignment is a decoder cosine of 0.341 at the OR-present pole and -0.287 at the OR-absent pole, above a random-direction reference of approximately 0.066. No single labelled feature accounts for the direction, and alignment is distributed across weakly related features. | The pattern is consistent with a distributed representation, with no single nameable feature accounting for it. Because the autoencoder is trained on the base model and applied to instruct-tuned activations, the decomposition is illustrative only. |

*Notes:* SAE=sparse autoencoder. Survey responses are scored on a normalised 0-to-1 scale and averaged within each instrument. Steering uses the gated coefficient unless otherwise stated. Complete conditions, item batteries, estimates, and statistical comparisons are reported in Appendix G.

## Technical appendix

## A. Scenario design

### A.1 Main scenario generation procedure

We generate the main scenario battery with Claude Opus 4.7 in batches of five. Each request begins with our operational definition of OR and instructs the model to produce contrastive pairs matched on incidental features but differing in how the actor interprets the same recurring signal. Claude drafts both scenarios, checks them against the specification rules, and revises them before returning the batch. The deterministic validation step (Appendix A.2) then re-checks rule compliance, and the adversarial filter (Appendix A.3) removes the pairs a surface classifier can still separate. Each returned pair includes the two scenarios, a domain label, and a short failure-mode label. Failure-mode labels are post hoc metadata and do not enter direction estimation.

### A.2 Validation rules and enforcement

The generation specification defines validation rules at three levels: vocabulary (applied to every scenario), pair (applied to the two members of a contrastive pair), and batch (applied to the set of pairs returned in a single call).

**(1) Vocabulary rules.** We ban words and phrases that could allow a classifier to separate OR-present from OR-absent scenarios by relying on surface vocabulary. First, we ban construct labels that name OR or closely approximate it (e.g., opportunity, gap, recognize). Second, we ban role labels (e.g., entrepreneur, founder, CEO), which could confound OR with persona activation. The validator enforces these bans mechanically, matching whole words and their morphological variants against 29 banned stems or phrases.

**(2) Pair-level rules.** These rules make the OR-present and OR-absent scenarios in each pair match on incidental features while differing primarily in interpretation. Both scenarios use the same actor, setting, domain, and activity, maintain similar length, fall within a fixed lexical-overlap band (Jaccard between 0.350 and 0.750), adopt a neutral descriptive register, third person in every format, and end with a single period. The Jaccard range excludes both very distant pairs, where broad surface differences could dominate the contrast, and near-identical pairs, where a trivial word substitution could become a shortcut. Two pair-level rules are especially important for construct validity. First, both poles must remain active. The OR-absent scenario does something concrete with the recurring signal and never merely ignores, stalls, or refuses. Second, the intended contrast is interpretative rather than behavioural. The two scenarios differ in how the actor interprets the same recurring signal, not in whether the actor acts.

**(3) Batch-level rules.** Across each batch of pairs, the OR-absent scenarios draw on different failure modes, and the OR-present scenarios open with different clauses and sentence structures. These constraints reduce repeated templates that a probe could learn across the battery. A repeated failure mode or opening structure would function as the machine-generated analogue of an annotation artefact in supervised datasets (Gururangan et al., 2018).

**Rule enforcement.** We enforce these rules in two stages. First, the generation prompt instructs Claude to check every rule and revise each pair before returning a batch. Second, a deterministic validation step re-checks each returned pair in code, verifying all rules that reduce to a mechanical test. These include per-scenario length band and the within-pair length difference, the Jaccard overlap band, the forbidden-word bans, the absence of first-person pronouns, and the terminal period. Pairs that fail these checks are discarded and replaced, so only compliant pairs enter the battery. The remaining rules rely on the generator and on manual review (e.g., surface parallelism, the matched main verb and modality, the requirement that both poles stay active, batch-level diversity).

### A.3 AFLite filtering against lexical leak

We run an adaptation of AFLite (Le Bras et al., 2020) as an additional guard against lexical leakage. The original filter scores predictability with probes on sentence embeddings. We substitute a TF-IDF logistic regression because the target here is a surface-vocabulary shortcut, and an embedding-based filter would also remove semantic content the construct requires. AFLite assigns each pair a predictability score, defined as the fraction of held-out predictions in which a TF-IDF logistic-regression classifier correctly recovers a scenario's OR-present or OR-absent label. It then iteratively removes pairs above the threshold $\tau$=0.750. This prunes the main candidate pool from 5,000 pairs to 636 over 449 iterations, lowering mean pair predictability from 0.909 to 0.509. To rule out the possibility that this reduction merely reflects a smaller battery, we compare the filtered battery with a random 636-pair subsample of the original pool.

We then re-audit the surviving battery with a freshly trained bag-of-words classifier over ten random group-aware splits that were not used during filtering. Grouping occurs at the pair level, so the OR-present and OR-absent members of a pair always fall in the same fold. Surface accuracy falls from 0.912 on the input pool to 0.528 on the filtered battery, close to the 0.500 chance baseline. By contrast, accuracy remains 0.867 on the size-matched random subsample and reaches 0.950 on the removed pairs. The reduction therefore reflects the filtering procedure, not sample size, and the removed pairs are the most separable part of the pool. Filtering approximately preserves the battery's domain composition, with the 636 kept pairs spanning 526 of the pool's 2,963 fine-grained domain labels and the largest domain moving from 0.019 to 0.020 of the battery. The result is not specific to logistic regression. A support-vector machine ranks pairs by predictability similarly (Spearman 0.811). These results indicate that any separability the direction-estimation probe recovers is not readily explained by the lexical cues captured by these surface classifiers.

### A.4 Resulting scenarios

The pipeline uses six scenario formats: the main contrastive pairs, three further pair formats (short, negation, and second-person), and two standalone formats (entrepreneurial and everyday) used in causal steering. Each format follows a written specification (Appendix A.1). The everyday format uses two specifications, one for the multi-sentence bands and one for the one-sentence band. The pair formats share a common base specification with a small format-specific overlay. The standalone formats add vocabulary bans, for example removing business framing from everyday scenarios and value- or risk-laden wording from both standalone formats. Table A.1 summarises the composition and role of each format. Table 2 in the main text reproduces example pairs verbatim, and the released battery contains each scenario.

**Table A.1. Scenario battery composition.**

| Format | Description | Structure | Count | Role in pipeline |
|---|---|---|---|---|
| Main | Contrastive pairs whose intended construct difference is the OR-present versus OR-absent interpretation. | OR-present/OR-absent pair | 636 | Direction estimation and probe evaluation |
| Short | Main contrast compressed to a single clause per pole, used to test robustness to shortened wording. | OR-present/OR-absent pair | 100 | Robustness check |
| Negation | Main contrast realised through explicit syntactic negators rather than verb substitutions. | OR-present/OR-absent pair | 100 | Robustness check |
| Second-person | Main contrast rewritten with the reader as the actor, to test robustness to person register. | OR-present/OR-absent pair | 100 | Robustness check |
| Entrepreneurial (standalone) | Single scenarios with an ambiguous signal a model could frame as OR-present or OR-absent. | Single scenario, stratified in 3 bands | 300 | Causal steering |
| Everyday (standalone) | Single scenarios in non-entrepreneurial contexts (nature, domestic, interpersonal) with no OR anchor, in a full-length and a one-sentence register. | Single scenario, stratified in 4 bands | 400 | Causal steering |

*Notes*: All scenarios are generated with Claude Opus 4.7. The main contrastive battery is filtered with AFLite.

## B. Activation extraction

We analyse Meta's Llama 3.1 8B-Instruct[7]. We record activations with TransformerLens version 3.2.1 (Nanda and Bloom, 2022), which exposes an LLM's internal activations through named hooks (intercept points attached to specific layers) and caches them during the forward pass. For each scenario, we apply the model's standard chat template and wrap the scenario as a user message (Arditi et al., 2024). We then run a single forward pass and record the residual stream after each of the 32 layers.

[7] Available at: https://huggingface.co/meta-llama/Llama-3.1-8B-Instruct, released July 23, 2024.

From each scenario's token sequence, we record six read positions per layer. Four are single-token positions at offsets -1, -2, -3, and -5 from the end of the tokenised prompt. The remaining two are pooled positions, one averaging over all tokens in the templated prompt and one over the scenario's own content tokens. We read several positions because, under the chat template, the final tokens close the user turn and need not carry scenario content, and OR-related information may be spread across the scenario without concentrating at any single token. The content-token mean excludes the chat-template wrapper by pooling only tokens between the shared template prefix and suffix. This yields one activation vector per scenario, layer, and read position.

## C. Direction estimation

From the activation vectors described in Appendix B, we estimate the OR direction with two linear readouts. The first is the difference-in-means (DiM) direction. At a given layer and read position, we average the OR-present training activations, average the OR-absent training activations, and take the difference between the two class means (Marks and Tegmark, 2024). To classify a scenario, we project its activation onto the DiM direction and compare the resulting score with the midpoint between the two training-class projection means. The second readout is an L2-regularised logistic regression probe (LR, C=1), trained on standardised activations to separate the two poles (e.g., Alain and Bengio, 2017; Belinkov, 2022). The standardisation parameters are estimated on the training split only.

We select the read position and layer jointly by cross-validation. For each of the six read positions and each of the 32 layers, we average DiM validation accuracy over 200 resampled 80/20 pair-level splits of the training pairs (seed 42), keeping both members of a pair on the same side of each split. We then select the layer-position combination with the highest mean validation accuracy. As a control for selection artefacts, we repeat the procedure with shuffled training labels and report the gap between true-label and shuffled-label accuracy (Hewitt and Liang, 2019). At the selected setting the shuffled-label control sits at 0.501, so the gap is 0.289. The selected setting is layer 12 at the -5 read position, the fifth token from the end of the tokenised prompt. At this setting, DiM validation accuracy is 0.789, compared with held-out test accuracy of 0.748, consistent with the expected validation-test gap after selecting the setting that maximises validation performance. After selection, we fit the final OR direction on the training pairs only and evaluate it on the held-out test pairs, reporting pair-level bootstrap 95% confidence intervals on test accuracy from 10,000 resamples.

We then run two further checks. First, we conduct a label-permutation test distinct from the shuffled-label control used during layer and read-position selection. We re-estimate the direction on the training pairs under 1,000 random within-pair label flips and record how often a permuted direction matches or exceeds the true held-out test accuracy. This gives a p-value against the null that the within-pair labels carry no separating information. Second, we conduct a direction-ablation check in which we remove the component of each activation along the OR direction and re-fit the logistic-regression probe on the ablated activations. Ablation equalises the class-conditional means by construction, removing the mean-difference signal a linear predictor exploits (cf. Belrose et al., 2023). The informative comparison is the matched random-direction ablation, which leaves probe accuracy essentially unchanged (0.835). A probe retrained in the ablated space stays at chance at every layer, an outcome the construction forces for any direction estimated as a class-mean difference, so we read this check as verifying the projection itself.

## D. Activation steering

### D.1 Steering procedure and parameter selection

The steering hook adds the raw DiM OR vector $r$ estimated in Appendix C. We use this vector at its native scale, without rescaling, so that the single coefficient α determines the size of the intervention and the same calibration can be used for the survey readout in Appendix G.1 (Turner et al., 2023). Unit-normalised directions are used for ablation and cosine analyses. We estimate $r$ at the -5 read position, a read-position choice also used by Arditi et al. (2024), and apply steering at the final position of each forward pass, which is the final token in the single-pass forced-choice readout. This implements a within-layer transfer from the read position used to estimate the direction to the token position used for intervention. At the selected strength the added vector $\alpha \cdot r$ remains small relative to the residual stream it perturbs, consistent with the in-distribution magnitude the coherence gate enforces.

**Calibration set.** We select α on a held-out calibration set of 35 standalone scenarios, five from each of the seven bands described in Section 3.2.2 (see also Appendix A.4).

**Operating point selection.** We sweep $\alpha$ over the grid {0.5, 1.0, …, 10.0} and select the operating point by a coherence gate. For two next-token distributions $p$ and $q$ over the vocabulary, KL divergence is defined as:

$$\mathrm{KL}(p \parallel q) = \Sigma\, p(v) \log[p(v) / q(v)],$$

Here, $p$ is the unsteered next-token distribution and $q$ is the steered next-token distribution at the same prompt position. For each candidate $\alpha$, we compute prompt-final KL divergence on the calibration set, using the final position of each calibration scenario, with no probe question appended. We retain the largest $\alpha$ for which mean KL remains at or below 0.100. This adapts the KL gate of Arditi et al. (2024) to coefficient selection under activation steering. The selected coefficient is α=3.5, with calibration-set KL=0.086 and evaluation-set mean KL of 0.080 (Appendix D.3). Sensitivity checks using alternative calibration subsets select coefficients between $\alpha$=2.5 and $\alpha$=5.5, placing the headline value $\alpha$=3.5 within the expected range.

**Competence check.** As a stress test, we add $r$ at every token position and measure per-token cross-entropy on held-out neutral text (Wehner et al., 2025). Mean cross-entropy remains close to the baseline value of 3.988 under OR-up steering (-0.045, 95% CI [-0.068, -0.023]) and modestly increases under OR-down steering (+0.169, 95% CI [+0.148, +0.190]). This indicates that even stronger all-position intervention leaves generic next-token predictive behaviour largely intact, although OR-down steering produces a small degradation.

**Random-direction control and necessity ablation.** The random-direction control draws 100 isotropic Gaussian vectors in residual-stream space, rescales each to the norm of the OR direction, and applies the same final-position hook at the operating point. We report the mean and standard deviation of the resulting log-odds shifts across the 100 directions. This null contextualises the observed steering shifts but is not the primary basis for direction specificity. Direction specificity rests principally on the necessity ablation, with the strengthened control below providing an additional check. The strengthened control draws 1,000 isotropic and 1,000 label-shuffled directions, then matches them on norm and realised prompt KL.

The necessity ablation projects out the identified OR direction from the residual stream at all positions:

$$\tilde{h} = h - (h \cdot \hat{r})\, \hat{r},$$

where $\hat{r}$ is the unit-normalised OR direction. We apply this projection to 60 scenarios across the three recognition questions, which are the main opportunity question and the two oblique questions (Appendix E.1) and compare it against the same projection using 100 matched random unit directions. We report the paired contrast (OR-ablation minus random-ablation) with a bootstrap confidence interval and a two-sided p-value (e.g., Elazar et al., 2021).

We also illustrate what happens when the steering coefficient is pushed past the coherence gate. Doubling the coefficient to α=7.0, we regenerate the bakery scenario from Table 4a in Table D.1. Mean next-token KL divergence remains within the gate at the selected coefficient (0.080 OR-up, 0.044 OR-down) but rises to 0.385 OR-up and 0.194 OR-down at the doubled coefficient, well above the gate. The generated text also begins to hedge. In particular, the OR-down explanation denies the opportunity before conceding that the scenario hints at one. We therefore treat the doubled-coefficient examples as an illustration of the strength trade-off.

**Table D.1. Steering-strength illustration: free-text explanations for an entrepreneurial scenario at the doubled steering coefficient (α=7.0).**

| Scenario | Condition | Response |
|---|---|---|
| **Bakery scenario of Table 4a (medium opportunity).** | **OR-up steering (α=7.0)** | The scenario describes an entrepreneurial opportunity because it identifies a customer need or demand that is not currently being met by the bakery. The owner's observation suggests that there is a market for gluten-free or wheat-free bread, which could be a growing trend driven by consumer demand for healthier and more specialized products. This opportunity is further supported by the proximity of the bakery to a potential customer base, as well as the existing infrastructure and supply chain for specialty breads. |
| | **OR-down steering (α=7.0)** | The scenario does not explicitly describe an entrepreneurial opportunity. However, it hints at a potential opportunity by mentioning that several customers have inquired about sourdough loaves made without wheat, which suggests that there may be a demand for this type of product. The fact that the bakery has not tracked these inquiries formally and that foot traffic has stayed level also implies that the bakery may not be aware of the full extent of this demand. |

*Notes:* Llama 3.1 8B-Instruct, layer 12. The bakery scenario of Table 4a is continued under OR-up and OR-down steering at the doubled coefficient (α=7.0), which exceeds the 0.100 coherence gate. The baseline and calibrated α=3.5 responses appear in Table 4a. The explanations are generated using greedy deterministic decoding and reproduced verbatim. The table is illustrative and is not used as a calibrated readout.

### D.2 Forced-choice question battery

Each probe is a yes/no question appended to a scenario. Every question refers directly to the scenario and ends with the identical suffix "Answer yes or no." The battery uses this single uniform frame with no separate lead-in, so the probes differ only in the question they pose. The fixed suffix places the model's answer at a common final token. For each scenario and question, we run a single forward pass and read the next-token distribution at that final position. The readout is therefore deterministic. We score the answer by pooling over surface forms. Each of "yes" and "no" resolves to six single-token variants, covering lower-case, title-case, and upper-case forms with and without a leading space. The decision log-odds is the log-sum-exp of the yes-variant logits minus the log-sum-exp of the no-variant logits. P(yes) is the corresponding softmax mass on the yes variants over the full vocabulary. Reading a steered model's response from answer-token log-odds at a fixed answer position, without sampling text, follows the readout used in activation-steering studies (Arditi et al., 2024; Rimsky et al., 2024). Steering enters this forward pass through the layer-12 hook described in Appendix D.1. The baseline applies no hook, while the OR-up and OR-down conditions apply the hook at $+\alpha$ and $-\alpha$ respectively.

We read every probe under three conditions: baseline, OR-up steering, and OR-down steering. For each of the 665 evaluation scenarios, we report the change from baseline in both log-odds and P(yes), averaged within band (95 scenarios) and within domain (285 entrepreneurial, 380 everyday). The anti-steerable fraction reported in Section 3.2.1 is computed on the 56-scenario dose subset and is the share of scenarios whose slope of the OR-up response across the KL-gated coefficient grid is negative. At the selected coefficient itself, the share of evaluation scenarios whose per-scenario shift opposes the intended steer is 0.116 for the recognition question, with the downstream values reported in Table 5. The reported baseline log-odds is the mean of the per-item log-odds. This differs from the log-odds of the mean P(yes), especially near the probability ceiling.

The selectivity margin in Table D.2 is defined in log-odds as the mean of the two oblique questions minus the mean of the two evaluative discriminants (the pleasantness and goodness probes), under each steering sign. The explicit OR probe is excluded from this margin. The primary specificity test compares the oblique mean against the discriminant mean and against the null mean as paired bootstrap contrasts under OR-up and OR-down. These four contrasts enter the same Holm-corrected primary family as the downstream contrasts defined in Appendix E.1, while the per-band margins are descriptive. The steering coefficient $\alpha$ is fixed on the held-out calibration set by the coherence gate described in Appendix D.1, independently of these readout probes. The reported selectivity tests are therefore not used in the tuning step.

### D.3 LLM judge implementation for free-text readout

The judge reads the steered free-text explanations from the run's generation log and scores them through DeepSeek's chat-completions interface. The free-text step elicits explanations with a single question, so the generated and judged sets are identical. This set contains the full held-out evaluation split of the three entrepreneurial bands, 95 scenarios per band, and is disjoint by construction from the 35 α-calibration scenarios. The judge therefore never scores a scenario that informs the choice of steering coefficient. For each scenario, the script fills the rubric of Box D.1 with the scenario text and the two steered explanations. The judge sees the full response, including any leading yes or no. The rubric, reproduced in Box D.1 except for a one-line JSON answer instruction and one dash rendered as a colon, asks which response more clearly reflects the fixed OR definition and instructs the judge to assess fit to that definition and to disregard length or elaboration. The judge runs with reasoning mode enabled and must choose one response.

We run the judge on two free-text sets that differ only in how the OR direction enters the generation. The headline set is generated with the OR direction added at the final position of every forward pass, so it acts on the last token of the prompt as the model reads the prompt and then on each response token as that token is generated. Averaged over the full 665-scenario evaluation set, its next-token KL divergence against the unsteered model is 0.080, inside the 0.100 coherence gate that selects the coefficient. On the 285 entrepreneurial scenarios that form this judged set, the realised divergence is 0.151 OR-up and 0.084 OR-down, so the judged corpus itself sits above the gate in the OR-up direction. The gate averages over all seven bands, and the entrepreneurial scenarios respond more strongly to the same coefficient than the everyday scenarios. The robustness set is generated using the same direction but adds it at every token position. This deliberately stronger intervention has next-token KL divergence of 0.671.

We report the all-position set as a labelled robustness ceiling in Appendix F.5 and Table F.4. The two corpora do not show a length artefact, with all-position and final-token generations averaging 86.660 and 89.077 tokens, respectively.

For each scenario, the judge compares the OR-up and OR-down free-text explanations as a paired choice, with the baseline response withheld, and selects the explanation that more clearly reflects an entrepreneurial opportunity according to our definition of OR. The headline protocol presents each pair once, with answer order assigned by seeded randomisation balanced within each band and panel, yielding one decision per scenario. As a robustness check against position bias, we also use an order-doubled protocol following Zheng et al. (2023), in which each pair is judged in both presentation orders. Under the conservative swap rule, a win counts only when the same explanation is preferred in both orders, and order-dependent flips are treated as inconsistencies. We test win rates against the 0.500 chance level using exact two-sided binomial tests.

**Box D.1. LLM-judge rubric for the free-text readout.**

Task: Read the scenario and the two responses to it below. Which response more clearly reflects this definition of opportunity recognition, Response A or Response B?

Definition (your reference for judging both responses): Entrepreneurial opportunity recognition is the cognitive act of noticing a generalisable structure in a scenario that could serve as the basis for a potential value-generating activity, independent of evaluation or action.

How to judge: Pick the response that more clearly recognises the opportunity itself, as defined: noticing the underlying structure that could become the basis for a new value-creating activity, rather than weighing, qualifying, or acting on it. A response is not a better match just because it is longer or more elaborated; judge only the fit to the definition.
Always commit to one of A or B; do not mark them as tied.

## D.4 Construct selectivity of the steering effect: OR content versus generic affirmation

Part 2 shows that steering the OR direction shifts the model's forced-choice recognition responses. A distinct question is whether this shift is specific to OR-related content or instead reflects a generic tendency to answer "yes", a sycophancy bias documented in models finetuned on human feedback (e.g., Sharma et al., 2024). We hold the steering direction and scenarios fixed and vary only the question. A construct-specific effect should shift OR-related questions more than evaluative discriminants and structural nulls.

The selectivity comparison uses nine of the battery's questions, ordered by intended relatedness to OR, with the exact wording shown in Table D.2. The set includes the main opportunity question (a), two explicit variants (b, c), two oblique questions that avoid the word "opportunity" (d, e) (cf. Geirhos et al., 2020), two evaluative discriminants capturing affect and positive evaluation (f, g), and two structural nulls (h, i). Comparing steering effects across target, discriminant, and null questions extends the control logic of representation-steering research, which verifies that a steering effect does not spill onto off-target material (e.g., Arditi et al., 2024; Rimsky et al., 2024).

The two control classes discipline this result in different ways. The evaluative discriminants are the sharper control. Pooled across domains, the oblique OR questions exceed the pleasantness and goodness questions by 0.207 log-odds under OR-up (95% CI [0.181, 0.234]). Under OR-down steering, the same contrast is -0.182 (95% CI [-0.218, -0.147]). Both contrasts remain significant after Holm correction (Appendix E.1). Because pleasantness and goodness capture the kind of positivity that a generic-affirmation account predicts should move, their smaller response helps separate OR-related content from generic positivity. The structural nulls provide weaker controls. The oblique OR questions exceed them by 0.207 log-odds under OR-up steering (95% CI [0.165, 0.250]), but some null questions begin near the response ceiling. On entrepreneurial scenarios, for example, the multi-person null has a baseline of P(yes)=0.986 and shifts by 0.688 log-odds. At this ceiling, even a broad affirmative shift can produce large log-odds changes. The entrepreneurial-specificity evidence therefore rests primarily on the pleasantness and goodness comparison. Several explicit OR probes also begin near the ceiling, with per-band baseline P(yes) between 0.814 (weak band) and 0.998 (strong band), so some off-target movement remains possible. The relevant criterion is systematically larger movement of OR-related questions relative to controls, not the absence of any off-target movement (e.g., Durmus et al., 2024; Tan et al., 2024).

To test whether the effect is specific to OR or merely reflects generic positive valence, we steer a norm-matched valence direction estimated with the same DiM procedure from a matched valence battery. We then compare the OR and goodness readouts, requiring each direction to move its own readout more than the other's. The sign criterion is not met. The double-dissociation index is -0.097, 95% CI [-0.280, 0.081]. The corresponding confirmatory test is an intersection-union member of the two-member causal family whose other member is the necessity-ablation contrast. It takes the maximum of its two limb p-values and is far from significance (Holm-adjusted p=1.000, raw two-sided p for the index alone 0.269). This test therefore does not support causal dissociation from valence. However, the raw cosine between the OR and positive-valence directions is small (0.125), providing geometric evidence that the two directions are not identical.

**Table D.2. Forced-choice readout under OR steering, full question battery, entrepreneurial scenarios.**

| | Baseline | OR-up steering | OR-down steering |
|---|---|---|---|
| **Questions** | **Log-odds (P(yes))** | **ΔLog-odds (ΔP(yes))** | **ΔLog-odds (ΔP(yes))** |
| **Entrepreneurial scenarios (n=285)** | | | |
| (a) Does the scenario describe an entrepreneurial opportunity? | 3.359 (0.891) | 0.401 (0.033) | -0.716 (-0.038) |
| (b) Is there a genuine entrepreneurial opportunity in this scenario? | 3.397 (0.918) | 0.519 (0.031) | -0.613 (-0.034) |
| (c) Is there an entrepreneurial opportunity in this scenario? | 5.379 (0.986) | 0.238 (0.006) | -0.629 (-0.009) |
| (d) Does the scenario describe something someone could build into something worthwhile? | 5.489 (0.992) | 0.295 (0.003) | -0.603 (-0.006) |
| (e) Does the scenario describe something that could also work in many other places? | 2.766 (0.917) | 0.392 (0.023) | -0.434 (-0.030) |
| (f) Does the scenario describe a pleasant situation? | -2.575 (0.309) | 0.063 (0.018) | 0.101 (-0.019) |
| (g) Does the scenario describe a good situation for the person? | -1.579 (0.315) | 0.078 (0.024) | -0.087 (-0.028) |
| (h) Does the scenario involve more than one person? | 4.991 (0.986) | 0.688 (0.006) | -0.194 (-0.003) |
| (i) Is the scenario written in the past tense? | 0.066 (0.524) | 0.018 (0.004) | 0.162 (0.029) |
| Selectivity margin (oblique minus evaluative discriminants) | | 0.273 | -0.526 |

*Notes*: Llama 3.1 8B-Instruct, layer 12, α=3.5. Entrepreneurial scenarios n=285 (three bands of 95), all held-out evaluation scenarios. The readout follows Appendix D.2. Baseline columns report log-odds with P(yes) in parentheses. Steering columns report changes. The selectivity margin (log-odds) is the mean of the two oblique OR questions (d, e) minus the mean of the two evaluative discriminants (f, g). A pooled random direction comparison on the 150 band-balanced evaluation scenarios (drawn round-robin across bands, 21 or 22 per band) gives a contextual null. In that subset, 11 of 100 random directions reach the OR-up shift. Under OR-down, the OR direction shifts log-odds by only -0.073 on this subset and 41 of 100 random directions reach that shift (empirical p=0.416).

## E. Representational integration

### E.1 Downstream coupling

The downstream battery adds three yes/no probes tracking the two evaluative doubts and the subsequent entrepreneurial action in McMullen and Shepherd's (2006) framework, with exact wording shown in Table 5. Every probe is appended after the scenario with the shared suffix ("Answer yes or no."). The readout is the next-token log-odds of "yes" versus "no". None of the downstream or discriminant questions enters OR-direction estimation, α calibration, or layer selection. A steering shift on these probes therefore reflects transfer from the OR intervention to adjacent readouts, with no circular reuse of the same questions (Arditi et al., 2024).

We assess downstream specificity using paired bootstrap contrasts pooled across the entrepreneurial and everyday evaluation scenarios. All contrasts are computed in log-odds with shared resample indices across the two arms of each contrast, so the same scenarios are drawn for both. The downstream mean averages the three downstream questions, the discriminant mean averages the pleasantness and goodness questions, and the null mean averages the two structural-null questions. Under OR-up steering, the downstream questions move more than the discriminant questions (downstream mean minus discriminant mean=0.049, 95% CI [0.015, 0.082]) and more than the structural-null questions (downstream mean minus null mean=0.048, 95% CI [0.009, 0.085]). Each contrast uses a paired bootstrap, with 1,000 resamples for the mean difference and its 95% percentile interval and 10,000 resamples for the two-sided p-value. These downstream contrasts join the four oblique contrasts of Appendix D.4 in a single eight-contrast primary family, which is Holm-corrected (Holm, 1979). The family includes downstream mean minus discriminant mean and downstream mean minus null mean under each steering sign. Seven of the eight contrasts pass. The exception is downstream mean minus discriminant mean under OR-down steering, whose estimate is 0.137 and differs from zero in the direction opposite to prediction, because the downstream readouts fall less than the pleasantness and goodness discriminants under OR-down.

We test sufficiency by adding the OR direction at the operating layer and final-token position. We test necessity by directional ablation, which projects the OR direction out of the residual-stream activation at every token position in the operating layer and re-reads the same probes (Arditi et al., 2024; Elazar et al., 2021). This is a single-layer variant of the directional ablation of Arditi et al. (2024), which removes the direction at every layer. We localise the ablation to the operating layer, where the steering coefficient is calibrated. As a matched control, we ablate 100 freshly drawn random unit directions one at a time, using the same protocol as the random-direction steering null. The necessity test runs on a subset of 60 evaluation scenarios stratified across the seven bands and pools the three downstream questions. In the recognition-question pool, we exclude the two explicit near-duplicate questions so that near-paraphrases do not pseudo-replicate the bootstrap. Each downstream construct is read with a single forced-choice question, trading construct coverage for a clean common scale.

### E.2 Discriminant geometry

Following the scenario generation logic of Appendix A, we build three additional batteries of 500 matched pairs. These batteries are generated with Claude Opus 4.8 (Anthropic, 2026b), the then-current successor of the Appendix A generator, and differ only on feasibility (enactable versus blocked), desirability (fulfilling the actor's guiding motive versus not), or exploitation (acted on versus left unused). Each comparator direction is estimated by DiM at the same layer and read position as the OR direction, so all directions occupy the same activation coordinates. We then report each comparator direction's cosine with the OR direction (Table E.1). As a robustness check, we also apply a de-anisotropy variant akin to the causal inner product of Park et al. (2024), here estimated as a diagonal class-centred rescaling of activation coordinates. This variant leaves the ordering unchanged.

We read each cosine against two references measured in the same model. The first is OR split-half reliability, defined as the mean cosine of two independent half-sample OR estimates over 100 splits (0.925). This provides a within-construct reference for reproducible OR alignment. The second is the analytic random-direction scale, defined as one over the square root of the model width (0.016), which gives the standard deviation of a random direction's cosine around zero. Because estimation noise on either side could mechanically lower OR-comparator cosines, we also report a disattenuated alignment that divides each cosine by the geometric mean of the OR direction's and the comparator construct's Spearman-Brown-corrected split-half reliabilities (Table E.1). These reliability estimates capture sampling noise within one battery style, not cross-battery style variance, which remains a limitation.

For each comparator construct, we test whether its cosine with OR falls below OR's split-half reliability reference. We use a joint paired bootstrap with 10,000 resamples and Holm-correct within the three-construct family. Every discriminant gap is significant after Holm correction. We then test the three comparator directions jointly by projecting the OR direction onto their combined span. The projection remains exact despite correlation among comparator directions, including the feasibility-exploitation cosine of 0.381. The three comparator directions together explain only 0.116 of the OR direction (a projection $R^2$, 95% CI [0.086, 0.138]). The residual remains strongly aligned with OR, with a surviving cosine of 0.940. A reliability-corrected residual alignment, computed as the residual cross-half cosine 0.918 divided by the OR split-half reliability reference of 0.925, is 0.993.

**Table E.1. The OR direction versus each comparator construct.**

| Construct | Cosine with OR | 95% CI | Own reliability | Disattenuated |
|---|---|---|---|---|
| Feasibility | 0.174 | [0.123, 0.215] | 0.981 | 0.178 |
| Desirability | -0.017 | [-0.048, 0.015] | 0.951 | -0.018 |
| Exploitation | 0.328 | [0.284, 0.356] | 0.984 | 0.335 |

*Notes*: Cosines are between DiM directions at layer 12. Confidence intervals are 95% bootstrap intervals (10,000 resamples). Reliability is each construct's split-half reliability (the mean cosine of two half-sample estimates over 100 splits). The disattenuated cosine divides each OR-comparator cosine by the geometric mean of the OR direction's and that comparator's Spearman-Brown-corrected split-half reliabilities, a deliberately generous adjustment for estimation noise on both sides. The reliability column reports the uncorrected half-sample value. All three cosines fall below the OR's split-half reliability reference of 0.925, and each discriminant gap remains significant after Holm correction within the three-construct family ($p<0.001$). The analytic random-direction scale in the 4,096-dimensional activation space is approximately 0.016.

We complement the cosine analysis with a descriptive steer-read matrix. Each row applies one norm-matched direction at the common selected coefficient, and each column reads one construct question on the same scenarios. Commerciality, estimated from the 500-pair battery described in Appendix F.2, is included as a context row. We read it in two ways. First, we examine whether rival directions move the OR question less than the OR direction does. Second, we examine whether each rival direction moves its own question more than the mean of the other construct questions. Both readings are interpreted alongside the realised prompt-KL dose for each row.

**Table E.2. Construct steer-read matrix for causal specificity.**

**Panel A. Mean change in log-odds by steered direction and readout**

| Steered direction | OR question | Feasibility | Desirability | Exploitation |
|---|---|---|---|---|
| OR | 0.317 [0.234, 0.397] | 0.551 [0.482, 0.618] | 0.255 [0.205, 0.303] | -0.001 [-0.128, 0.128] |
| Feasibility | 0.616 [0.518, 0.712] | 0.393 [0.356, 0.430] | 0.201 [0.147, 0.257] | 0.551 [0.420, 0.683] |
| Desirability | 0.150 [0.104, 0.198] | 0.369 [0.315, 0.419] | 0.065 [0.011, 0.119] | 0.036 [-0.041, 0.112] |
| Exploitation | 0.020 [-0.034, 0.072] | 0.218 [0.162, 0.275] | -0.143 [-0.191, -0.095] | -0.059 [-0.168, 0.049] |
| Commerciality | -0.515 [-0.596, -0.440] | -0.177 [-0.233, -0.118] | -0.531 [-0.632, -0.433] | -0.682 [-0.887, -0.479] |

**Panel B. Descriptive contrasts and dose diagnostics**

| Steered direction | OR-column gap | Own-minus-other | Raw cosine with OR | Mean prompt KL |
|---|---|---|---|---|
| OR | Reference | Not interpreted | 1.000 | 0.078 |
| Feasibility | -0.299 [-0.424, -0.172], $p<0.001$ | -0.063 [-0.151, 0.025], $p=0.159$ | 0.174 | 0.031 |
| Desirability | 0.167 [0.078, 0.257], $p<0.001$ | -0.120 [-0.167, -0.075], $p<0.001$ | -0.017 | 0.045 |
| Exploitation | 0.297 [0.208, 0.385], $p<0.001$ | -0.091 [-0.193, 0.009], $p=0.079$ | 0.328 | 0.053 |
| Commerciality | 0.832 [0.757, 0.911], $p<0.001$ | Not applicable | -0.007 | 0.071 |

*Notes:* Estimates use 60 evaluation scenarios per cell with scenario-level bootstrap confidence intervals. Each cell reports mean change in log-odds from the common unsteered baseline. Each direction is estimated independently and norm-matched to the OR direction, then steered at

the common coefficient and read on the same scenarios. Panel B reports two diagnostic contrasts. The OR readout gap compares the OR-row shift with each rival direction's shift on the OR question. Positive values indicate that OR steering moves the OR question more than the rival direction. Own-minus-other compares each rival's own readout with the mean of the other readouts. Mean prompt KL reports realised intervention dose. The exploitation column uses the action-oriented exploitation question from Table 5. Confidence intervals and two-sided p-values accompany the filled estimates.

# F. Robustness checks

## F.1 Robustness to scenario length, phrasing, and person register

A standard robustness check in representation engineering tests whether a recovered direction generalises to inputs unlike those used to estimate it. A direction tied to the battery's surface form should degrade when scenario length or contrast framing changes (e.g., Marks and Tegmark, 2024; Tigges et al., 2024). We test this in three ways. First, we shorten the scenarios to 20–30 words to check whether the OR direction depends on wording that accumulates in the longer main scenarios. Second, we realise the contrast through explicit negation. Both members of each pair contain a matched number of negation words, within one, with the negation scope swapped across the poles, a framing under which linear probes are known to struggle (Marks and Tegmark, 2024). Because the OR direction is estimated without negation framing, transfer to this set indicates that the direction tracks the interpretative contrast, not the negation wording. Third, we rewrite 100 otherwise matched pairs in the second person, placing the reader in the actor's position, to test whether the direction transfers beyond an observer-framed scenario.

For each variation, we re-estimate the DiM direction on resampled subsets of the 509 training pairs, apply each estimate to the new evaluation set, and report mean transfer accuracy with a 95% bootstrap interval across resampled directions (e.g., Li et al., 2023; Marks and Tegmark, 2024). The layer and read position remain fixed at the values selected in Section 3. The robustness sets are therefore used only for evaluation, not for layer selection or direction estimation.

The OR direction transfers across all three variations (Table F.1), with second-person transfer the strongest of the three. We interpret this as register transfer, not as evidence that the model itself occupies a first-person state of recognition. Re-training the probe after projecting out the OR direction returns accuracy to chance in every set by construction, and the re-estimated per-variation directions align substantially with the main direction without being identical to it (Table F.1). These results make it less likely that the main recovery result depends on the surface form of the original battery, while not ruling out lexical or semantic confounding more broadly. One reading caveat applies.

**Table F.1. Robustness to scenario length, phrasing, and person register.**

| | | DiM | | LR | | | |
|---|---|---|---|---|---|---|---|
| **Scenario set** | **n** | **Mean** | **[95% CI]** | **Mean** | **[95% CI]** | **LR ablated** | **Similarity** |
| Main scenarios (cross-validation stability band) | 509/127 | 0.789 | [0.740, 0.838] | 0.839 | [0.795, 0.878] | 0.500 | - |
| Short scenarios | 100 | 0.781 | [0.765, 0.800] | 0.835 | [0.785, 0.880] | 0.500 | 0.746 |
| Negation scenarios | 100 | 0.841 | [0.825, 0.855] | 0.900 | [0.860, 0.935] | 0.500 | 0.681 |
| Second-person scenarios | 100 | 0.860 | [0.830, 0.885] | 0.900 | [0.860, 0.940] | 0.500 | 0.752 |

*Notes*: This table reports out-of-sample performance of the layer-12 OR direction on scenario sets that differ from the main battery in length (short), phrasing (negation), and person register (second-person). DiM accuracy is averaged over 200 resampled directions estimated from the training pairs, so the confidence interval reflects stability across resampled directions, not a population interval over items. For the main scenario reference row, the DiM mean is the average cross-validation accuracy across training resamples and is evaluated on the held-out folds of the 509 training pairs rather than on the 127 lockbox pairs, so it is higher than the lockbox estimate reported in Section 3.1.3. In that row, n lists the DiM basis before the LR basis. LR reports the single probe from Section 3.1.3 applied without re-fitting, with 95% bootstrap confidence intervals over held-out pairs. LR after ablation reports accuracy after projecting the OR direction out of the activations and re-training the probe. Similarity is the cosine between the OR direction re-estimated on the row's own scenario set and the main OR direction.

### F.2 Robustness to a broader commercial-venture axis

AFLite addresses lexical shortcuts, but it cannot rule out a semantic rival built into the scenario contrast. OR-present scenarios may more often frame the recognised structure as something sold to paying customers, whereas OR-absent scenarios may remain non-commercial. This rival is demanding because market exchange is part of influential definitions of entrepreneurial opportunity (Shane and Venkataraman, 2000), while our operational definition aims to isolate recognition from evaluation and action. Construct clarity therefore requires testing whether the recovered direction reflects OR or a broader commercial-venture axis (Campbell and Fiske, 1959; Davidsson, 2015; Suddaby, 2010). Our first analysis controls for commerciality within the 636-pair OR battery. A blind coder (DeepSeek V4 Pro), from a model family different from the scenario generator, rates each scenario independently, without access to its paired scenario or the OR label. The rating captures the extent to which value arises from selling to paying customers, as opposed to being provided free to a community. The ratings confirm a real imbalance: 0.511 of OR-present scenarios are rated commercial, compared with 0.250 of OR-absent scenarios, yielding a correlation with the OR label of 0.270 (95% CI [0.233, 0.310]).

If the OR probe reads recognition and not commercial framing, its accuracy should hold in pairs where commerciality carries no distinguishing signal, and it does (Table F.2). On commerce-matched pairs, where both scenarios fall on the same side of the commercial cut, DiM accuracy on the held-out pairs is 0.750 against a 0.500 chance level. Accuracy on mixed pairs is 0.765 against 0.750 on commerce-matched pairs (difference 0.015, 95% CI [-0.083, 0.110], permutation p=0.840, with LR corroborating at 0.060, p=0.272). With 34 mixed pairs the interval cannot rule out a moderate commercial-alignment advantage, so we read this as descriptive convergence, not as a test of no difference. A probe that reads commerciality would still be expected to fall to chance on the commerce-matched pairs, and it does not. As a stronger test, we erase linearly decodable commerciality from the activations using LEACE (Belrose et al., 2023), whose guarantee applies to the fitting sample (post-erasure commerce probe AUC 0.504 there), and re-estimate the OR probe. On held-out pairs a refit commerce probe still reaches an AUC of 0.724, down from 0.859, so the erasure is incomplete out of sample and the resulting drop in OR accuracy is a lower bound on commerce dependence. LR accuracy falls by only 0.028 (0.839 to 0.811), a drop larger than any produced by the 300 matched random erasures but small enough to leave the OR signal largely intact.

Our second analysis estimates an independent commerciality direction from 500 matched pairs that hold recognition constant and vary only whether value flows through a paying or a free channel. The OR and commerciality directions are close to orthogonal. Their whitened cosine, the alignment after adjusting for the shared covariance of the activation space, is 0.000 (95% CI [-0.022, 0.021]). The commerciality direction is the most reliable of our comparator directions, with split-half reliability of 0.986, so the near-zero alignment is not a noise artefact. The separation also holds causally. Steering the commerciality direction at the same norm and coefficient moves the OR readout in the opposite direction to OR steering (paired gap 0.832 log-odds, 95% CI [0.757, 0.911], $p<0.001$) at comparable doses (mean prompt KL 0.078 against 0.071). The gap therefore does not reflect under-dosing of the commerciality direction. Table E.2 reports the same gap with a scenario-level bootstrap interval that agrees to three decimals. The two analyses measure commerciality at different levels by design. The independent battery isolates the narrow question of where a recognised gain ends up, paying from free value channels while holding recognition constant. The blind coder, by contrast, assesses the broader commercial content of the whole scene. The geometric and causal separation therefore speaks to the narrow value-destination contrast, while the broader-content conclusion rests on the first analysis, where the OR probe survives exactly where a commerce shortcut would be expected to fail.

**Table F.2. Robustness to commercial venture framing.**

| Pair stratum | DiM accuracy | LR accuracy | Held-out pairs |
|---|---|---|---|
| Commerce-matched pairs | 0.750 | 0.822 | 90 |
| Mixed pairs, commercial cue aligned with the OR label | 0.765 | 0.882 | 34 |

*Notes:* A blind coder (DeepSeek V4 Pro) rates each scenario independently on a 0-to-3 scale for the extent to which value arises from selling to paying customers. Pairs are commerce-matched when both scenarios fall on the same side of the commercial cut, where a rating of 2 or 3 counts as commercial, and mixed otherwise. The table reports OR probe accuracy on held-out pairs within each stratum. Chance is 0.500.

### F.3 Robustness to the scenario generator: DeepSeek instead of Claude

We generate the main contrastive battery with Claude Opus 4.7, so a recovered direction could partly reflect label-correlated stylistic regularities of that generator, not the intended OR contrast (e.g., Wang et al., 2023). We therefore construct an independent battery with DeepSeek V4 Pro (DeepSeek-AI, 2026) under the same specification, yielding 632 training pairs and 158 held-out pairs. We then test cross-generator transfer by training an LR probe on pairs generated by one model and evaluating it on held-out pairs generated by the other. The Claude-trained probe classifies DeepSeek-generated scenarios with 0.832 accuracy, while the DeepSeek-trained probe classifies Claude-generated scenarios with 0.783 accuracy. For comparison, the corresponding within-generator accuracies are 0.839 (Claude) and 0.864 (DeepSeek). The two OR directions align at a cosine of 0.799, a value that none of 1,000 per-pair sign-flip permutations reaches (p=0.001). Cross-generator transfer therefore weakens the concern that the OR direction rests on generator-specific surface form, though artefacts shared by the common scenario specification remain possible.

### F.4 Robustness to analysed LLM: varying scale and model family

Our main analysis tests the OR direction in a single open-weight model, Llama 3.1 8B-Instruct (Grattafiori et al., 2024), leaving open whether the recovered direction reflects an idiosyncrasy of that model's architecture, scale, or training data. The Claude-versus-DeepSeek convergence reported in Appendix F.3 addresses the input-data side of this concern. To address the model side, we repeat the core analyses on additional open-weight LLMs and ask whether an OR direction can be recovered at comparable held-out accuracy, whether its ablation removes the probe signal, whether steering shifts the recognition readout in the predicted direction, and whether the direction remains geometrically distinct from adjacent evaluation and exploitation constructs.

The replication varies two dimensions. First, we vary scale and rerun the pipeline on Llama 3.1 70B-Instruct (Grattafiori et al., 2024), which shares the Llama 3 architecture with our headline model but is roughly an order of magnitude larger. If OR is not idiosyncratic to the 8B model, an OR direction should also be recoverable in the larger network, although its most informative layer need not occur at the same relative depth. Second, we vary architecture and training corpus, re-running the pipeline on three open-weight models from two non-Llama families, Qwen3-8B (Qwen Team, 2025), Qwen2.5-32B-Instruct (Qwen Team, 2024), and Gemma-3-12B-it (Gemma Team, 2025). These models differ from Llama in architecture and training corpus while remaining instruction-tuned. The two Qwen models also span an 8B to 32B range within one developer family, although they belong to different model generations, so scale and training-recipe differences are not fully separable.

For every model, Table F.3 reports the selected layer and relative depth, held-out DiM accuracy for recovery (Part 1), OR-up and OR-down shifts on the main OR question for steering (Part 2), and cosines with the model's own feasibility, desirability, and exploitation directions for geometry (Part 3). Llama 3.1 70B-Instruct tests an order-of-magnitude scale step within the headline family, while Qwen3-8B, Qwen2.5-32B-Instruct, and Gemma-3-12B-it test transfer beyond the Llama family. Every model recovers an OR direction with held-out accuracy close to the headline estimate of 0.748 and above the bag-of-words leakage reference from Section 3.1.1. Ablating the recovered direction returns the re-trained LR probe to chance (0.500) throughout, by construction, and every model shifts the main OR question in the expected directions under OR-up and OR-down steering (Table F.3). The steering coefficient is calibrated separately for each model under the same KL gate, so the sign and presence of the shift replicate, while effect magnitudes are not directly comparable across model families. The geometry shows the same separability in every replication model. Each model's OR direction aligns only modestly with its own feasibility, desirability, and exploitation directions, and these comparator cosines remain far below that model's OR split-half reliability.

Because a direction estimated in one model lives in that model's own activation space, it is not directly portable across models. We therefore re-estimate the OR direction separately in each model using the procedure of Section 3.1.3 and compare models at the data level. The question is whether each model's own DiM readout separates the same held-out scenario pairs. This provides the model-side analogue of the generator comparison in Appendix F.3.

The cross-family evidence is mixed. Recovery and signed bidirectional shifts appear in all three non-Llama models, but these lean cross-family runs omit the random-direction control. Construct selectivity replicates only in part. The null-tier questions move with the OR questions in all three models, and the OR-versus-null separation observed in the headline model replicates in Qwen2.5-32B-Instruct (+1.978 log-odds, 95% CI [+1.839, +2.127]) but is much weaker relative to its own steering effect in Gemma-3-12B-it (+0.175, 95% CI [+0.095, +0.257]) and reverses in Qwen3-8B (-0.079, 95% CI [-0.135, -0.023]). Gemma-3-12B-it also sits near the yes-ceiling in probability terms, with a baseline P(yes) of 0.989 and four probes starting at or above P(yes)=0.950, three of them at 1.000, which is why Table F.3 reports steering in log-odds, not probabilities. These log-odds magnitudes are not comparable across families in absolute terms. The reported value is the raw gap between the pooled yes and no logits, meaning the model's pre-probability scores for the answer tokens. A model that places the two answer tokens farther apart can show both a larger baseline and a larger shift for the same underlying probability-scale change. This reflects output-scale differences, not stronger encoding of opportunity. Qwen2.5-32B-Instruct, for example, is less certain than Llama 3.1 8B-Instruct in probability terms that the scenarios describe opportunities (0.872 against 0.891), yet its baseline log-odds is roughly five times as large (16.596 against 3.359). The same output-scale inflation also appears in its null questions. The cross-family comparison therefore rests on the sign and presence of the shift, while direction-specific selectivity is assessed only where the required controls are run. In the 70B replication, 0.520 of scenarios move against the intended steer on the recognition probe, so its large mean shift also averages a directionally split sample.

**Table F.3. Robustness of the OR direction across open-weight LLMs.**

| Model | Part 1: Representational existence | | Part 2: Causal operativity | | | | Part 3: Representational integration | | |
|---|---|---|---|---|---|---|---|---|---|
| | Layer (depth) | DiM | Baseline log-odds | Baseline P(yes) | ΔLog-odds (OR-up) | ΔLog-odds (OR-down) | Cosine OR and feasibility | Cosine OR and desirability | Cosine OR and exploitation |
| Llama 3.1 8B-Instruct | 12/32 (0.375) | 0.748 | 3.359 | 0.891 | 0.401 | -0.716 | 0.174 | -0.017 | 0.328 |
| Llama 3.1 70B-Instruct | 37/80 (0.463) | 0.728 | 7.157 | 0.921 | 1.935 | -1.979 | 0.267 | 0.144 | 0.463 |
| Qwen3-8B | 24/36 (0.667) | 0.705 | 9.417 | 0.905 | 1.859 | -1.823 | 0.392 | 0.224 | 0.405 |
| Qwen2.5-32B-Instruct | 44/64 (0.688) | 0.744 | 16.596 | 0.872 | 3.543 | -4.855 | 0.305 | 0.139 | 0.487 |
| Gemma-3-12B-it | 27/48 (0.562) | 0.717 | 18.701 | 0.989 | 1.682 | -1.574 | 0.310 | 0.218 | 0.305 |

*Notes*: Each row uses an OR direction independently re-estimated in the corresponding model on the shared post-AFLite battery (636 pairs, 509 train, 127 held-out test), with the layer selected separately for each model using the 200-split cross-validation procedure from Section 3.1.3. DiM reports held-out (lockbox) accuracy of the difference-in-means probe. Baseline log-odds and baseline P(yes) refer to the main OR question on the entrepreneurial evaluation set. OR-up and OR-down columns report steering-induced changes in log-odds. Steering coefficients are calibrated separately for each model under the same KL gate, so magnitudes are not directly comparable across models. The selected read position is -5 for both Llama models and Qwen2.5-32B-Instruct, and -1 for Qwen3-8B and Gemma-3-12B-it. The selected coefficients are $\alpha$=3.5 for both Llama models, 1.5 for Qwen3-8B, 2.0 for Qwen2.5-32B-Instruct, and 2.5 for Gemma-3-12B-it. Cosine columns report alignment between each model's OR direction and its own feasibility, desirability, and exploitation directions.

## F.5 Robustness of the free-text judge readout

Section 3.2.3 scores each free-text pair once in a single-presentation protocol, with the OR-up response randomly assigned to one side. We check this readout in two ways. First, we test sensitivity to the judge's position bias by re-judging every pair in both presentation orders. Second, we test sensitivity to intervention placement by adding the OR direction at every token position, so the scenario tokens are steered as well as the response tokens. Table 4b reports the headline readout together with the order-swap check, and Table F.4 reports the all-position check.

The first check re-judges the headline corpus in both presentation orders. Under the conservative swap rule of Zheng et al. (2023), a win counts only when the same explanation is preferred in both orders, so every position-driven flip becomes an inconsistency. This stricter protocol raises the OR-up win rate from 0.656 in the single-presentation protocol to 0.708 among decisive entrepreneurial pairs (Table 4b). Judge consistency, defined as the share of pairs whose verdict survives the presentation-order swap, is 0.709. This is a conservative floor because the forced binary choice gives the judge no tie option. The inconsistency is entirely directional, with 0.291 of pairs flipping toward the first-presented response and 0.000 toward the second. The single-presentation protocol assigns order by seeded balance within band and panel, which prevents systematic favouring of either condition, although position bias can still affect the realised estimate, and roughly three in ten decisions are position-driven. A verdict-cue check addresses the further concern that the win rate merely restates the categorical openings of the explanations, since nearly all explanations open with a verdict-like clause. Of the 285 judged pairs, the two openings carry the same verdict in 221, involve a hedged opening in 33 pairs, and disagree in the steered direction in 31 pairs (0.109), never in the opposite direction. Within the 221 pairs whose openings carry the same verdict, so that the opening cannot decide the comparison, the judge still prefers the OR-up explanation in 0.747 of pairs in the first presentation order and 0.636 under the swap rule. The judged difference is therefore not reducible to the opening verdict and reflects substantive difference in the explanations.

The second check adds the OR direction at every token position, so the intervention affects the scenario tokens as well as the response tokens throughout the forward pass. This raises the single-presentation win rate to 0.951 and the order-doubled rate to 0.971 (Table F.4), with every opportunity band significant on its own. The all-position intervention is also substantially more disruptive, with mean next-token KL divergence of 0.671 well above the 0.080 final-token value.

**Table F.4. Free-text readout under all-position OR steering: LLM-judge ratings.**

| | Single presentation | | | Order-doubled robustness | | |
|---|---|---|---|---|---|---|
| **Scenarios** | **Win/loss** | **Win rate** | **vs. chance** | **Win/tie/loss** | **Win rate** | **vs. chance** |
| **Entrepreneurial scenarios** | **271/14** | **0.951** | **<0.001** | **268/9/8** | **0.971** | **<0.001** |
| Weak opportunity | 95/0 | 1.000 | <0.001 | 94/1/0 | 1.000 | <0.001 |
| Medium opportunity | 94/1 | 0.989 | <0.001 | 94/1/0 | 1.000 | <0.001 |
| Strong opportunity | 82/13 | 0.863 | <0.001 | 80/7/8 | 0.909 | <0.001 |

*Notes:* Llama 3.1 8B-Instruct, layer 12, α=3.5, with the OR direction added at every token position as an all-position robustness intervention (mean next-token KL divergence 0.671, beyond the 0.100 coherence gate by design). The judge, protocols, and swap rule are the same as in Table 4b. The all-position corpus contains 285 entrepreneurial scenarios, 95 per band, and judge consistency is 0.968. Win rate is the share of decisive pairs preferring the OR-up explanation, and p-values are exact two-sided binomial tests against 0.500.

# G. Additional analyses

## G.1 Steering versus prompting an entrepreneurial role

### G.1.1 Persona conditions and the persona-versus-steering factorial

A natural objection to the steering results is that prompting the model into an entrepreneurial role could achieve the same effect (Obschonka and Fisch, 2025; Shanahan et al., 2023). We therefore ask what steering adds over role prompting. Two cautions govern the comparison. First, a written instruction and a steering coefficient share no common scale, so the comparison cannot establish which lever is stronger in an absolute sense. Second, a prompt and internal direction that shift the same readout need not share a common cause, so the persona arm provides convergent evidence rather than a second causal test (e.g., Geiger et al., 2021), with prompting and steering routinely benchmarked as distinct intervention families on a shared readout (e.g., Wu et al., 2025). The comparison therefore concerns breadth and reversibility, the properties on which the two levers differ, and only steering includes a signed down arm.

The persona condition uses the entrepreneur-role instruction from Obschonka and Fisch (2025): "In this conversation, please assume the role of an entrepreneur throughout." The instruction is prepended to the scenario, with no activation steering. We cross this persona condition with the forced-choice readout described in Appendix D.2 in a six-condition design for each scenario: neutral baseline, OR-up steering, OR-down steering, persona alone, and persona combined with each steering direction. Every condition is scored using the same answer-token log-odds at the final position. Each cell is reported as a change from the common neutral baseline in log-odds, averaged within band. Table G.1 reports the per-band point estimates. Confidence intervals for the calibrated alpha=3.5 steering effects are reported in Table 3.

The clearest difference is breadth. The entrepreneur persona increases OR responses much more broadly than gated steering, raising the readout in every band, including the strong-opportunity band that already sits near the response ceiling. On weak-opportunity scenarios, for example, the persona shifts log-odds by 1.917, compared with 0.530 under OR-up steering. Gated steering moves the same readout more modestly and in a band-sensitive manner.

Steering is also reversible in a way the persona is not. OR-down steering supplies a signed downward intervention, lowering the OR readout by up to 0.989 log-odds on strong-opportunity scenarios. Pairing the persona with OR-down steering reduces the persona effect in every band, for example from 1.917 to 0.613 on weak-opportunity scenarios. This partial reduction is consistent with the persona operating partly through the same direction, but it does not establish that mechanism. The stricter necessity evidence remains the ablation result reported in Section 3.2.2.

The pipeline also includes an instruction-level counterpart to the persona condition, using a one-sentence alertness framing prepended to 60 evaluation scenarios without steering (Wu et al., 2025). Under steering, the oblique OR probes shift by 0.429 log-odds, against 0.205 and 0.267 for the two control tiers. The alertness instruction increases the same oblique probes less (0.190) and pushes both control tiers strongly downward (-0.450 and -0.882). The difference-in-differences comparisons do not show the vector to be more selective than this instruction (steering minus prompting of the oblique-minus-off-tier margin is -0.417, 95% CI [-0.978, 0.178], $p=0.158$ against the discriminant tier, and -0.911, 95% CI [-1.136, -0.684], $p<0.001$ against the null tier). The two levers instead differ in the direction of off-target movement, with steering nudging the control tiers slightly upward while the instruction pushes them sharply downward.

A doubled-coefficient diagnostic reads the headline question at twice the selected coefficient, starting from the neutral baseline with no persona. Once the intervention moves beyond the coherence gate, the readout is no longer monotonic. In bands whose baseline already sits near the yes-ceiling, the OR-up shift reverses, matching the over-steering reversal described in Section 3.2.1. Table G.1 reports the per-band values at both coefficients.

**Table G.1. Forced-choice readout: representation-level steering versus an entrepreneur persona across the entrepreneurial neutral bands.**

| | | | Steering (ΔLog-odds from baseline) | | | | Persona conditions (ΔLog-odds from baseline) | | |
|---|---|---|---|---|---|---|---|---|---|
| **Scenario set** | **n** | **Baseline P(yes)** | **OR-up (α=3.5)** | **OR-down (α=3.5)** | **OR-up (α=7.0)** | **OR-down (α=7.0)** | **Entrepre-neur** | **Entrepre-neur + OR-up** | **Entrepre-neur + OR-down** |
| **Entrepreneurial scenarios** | **285** | **0.891** | **0.401** | **-0.716** | **-0.332** | **-2.258** | **1.563** | **1.585** | **0.293** |
| Weak opportunity | 95 | 0.814 | 0.530 | -0.542 | 0.116 | -2.087 | 1.917 | 2.130 | 0.613 |
| Medium opportunity | 95 | 0.873 | 0.547 | -0.616 | -0.012 | -2.114 | 2.150 | 2.223 | 0.746 |
| Strong opportunity | 95 | 0.986 | 0.125 | -0.989 | -1.099 | -2.572 | 0.623 | 0.403 | -0.479 |

*Notes:* Llama 3.1 8B-Instruct, layer 12. The forced-choice readout uses the main OR question and deterministic next-token probabilities. Baseline P(yes) reports the neutral baseline probability of "yes." All other cells report changes in log-odds from the band's neutral baseline. Calibrated steering uses α = 3.5. The α = 7.0 columns are a doubled-coefficient diagnostic without persona prompting and exceed the coherence gate by design. Persona columns use the entrepreneur-role instruction from Appendix G.1, alone or combined with calibrated OR-up or OR-down steering. All values are descriptive and sit outside the multiplicity-corrected families.

### G.1.2 Survey-style construct profile

We benchmark the two levers a second way, using a survey-style profile of items from published entrepreneurship, personality, and control instruments. This analysis is a descriptive convergent-and-discriminant pattern check, not a psychometric validation. If the OR direction captures opportunity-recognition content, OR-up steering should raise OR-relevant profiles, OR-down steering should lower them, and both effects should concentrate on OR-relevant instruments and stay limited for off-construct measures. We compare these interventions with four role prompts. The first is the entrepreneur persona. The second is a negated persona ("assume the role of someone who is not an entrepreneur") which mirrors the entrepreneur persona in the same sentence frame. The third is a high-OR persona ("assume the role of someone who is very high in opportunity recognition", followed by the operational definition verbatim), and the fourth is a low-OR mirror, with "very low" replacing "very high". The latter two name the disposition, not an occupation, and anchor the role in the operational OR definition from Section 3.1.1.

The reported battery contains 10 established instruments: one measure of OR, one opportunity exploitation measure, four theoretically related entrepreneurial instruments (alertness, self-efficacy, proactive personality, and bricolage), the Big Five personality traits, and three off-construct measures (risk propensity, fear of failure, and ADHD). Table G.2 lists each instrument with its items, subscales, response format, and source. We present each item to the model in its native response format and read out the rating deterministically as the probability-weighted average of the scale points at the answer position. Scores are normalised to a 0-to-1 range and averaged across items within each instrument. Negatively worded items are reverse-keyed, and the item wording follows the published instruments cited in Table G.2. Several instruments presuppose a personal history the model does not have, most directly the exploitation items, which ask about past founding behaviour (Kuckertz et al., 2017), so baseline columns serve as reference points for within-model contrasts, not interpretable levels. The focal opportunity scales contain no reverse-keyed items, so a within-scale acquiescence check is not available. Two patterns nevertheless argue against a generic agreement drift. Forward-keyed off-construct instruments on five-point response formats barely move in raw terms under OR-up steering (ADHD -0.038 and fear of failure +0.001 raw scale points, against +0.293 for the forward-keyed bricolage scale), and within the Big Five, where reverse-keyed items exist, forward- and reverse-keyed items move in the same raw direction, a local agreement shift that cancels after reverse-keying and therefore does not inflate the reported Big Five scores. A uniform yes-bias would lift all forward-keyed items alike, which is not what we observe. The readout renormalises the answer distribution over the rating digits. A companion diagnostic tracks how many survey items exceed the coherence gate under the steering arms, because the gate is calibrated on neutral scenarios, not on survey prompts. At the selected coefficient, 0.190 of items under OR-up and 0.201 under OR-down exceed the 0.100 gate. At twice the coefficient, 0.974 and 0.781 of items do, with median item KL of 0.765 and 0.311.

We report eight conditions from the survey readout. Three come from the main descriptive comparisons against the unsteered baseline: OR-up steering, OR-down steering, and the entrepreneur persona. Each is evaluated as a paired comparison of scale-level scores, with Holm correction applied across the three contrasts as multiplicity adjustment within this descriptive analysis (Holm, 1979). The remaining five conditions stand outside any multiple-comparison family. OR-up and OR-down at twice the selected coefficient exceed the 0.100 coherence gate from Appendix D.1 by design and are interpreted as dose-escalation diagnostics. The high-OR, low-OR, and negated-persona conditions are descriptive comparison prompts. Table G.3 reports steering at the gated coefficient and at twice that coefficient, alongside the entrepreneur, high-OR, and low-OR persona conditions. The negated-persona values are reported in the text below. The survey readout scores and retains every scale and subscale of the battery in full. The ten tabled instruments span both sides of the prediction, from the largest movers to the off-construct anchors that barely move, and the omitted instruments show broadly intermediate shifts.

Table G.3 reports a directional and selective pattern. OR rises from 0.652 at the baseline to 0.718 under OR-up and falls to 0.625 under OR-down, exploitation and the related entrepreneurial instruments move the same way, and doubling the coefficient widens the OR shift in both directions (Table G.3), although 0.974 of items then exceed the coherence gate, so we read the doubled-coefficient columns as dose escalation past the calibrated regime, not as an extension of the gated dose response. The persona offers no analogue of a signed, graded intervention. Two boundaries apply. The alertness scale's evaluation subscale (Tang et al., 2012) overlaps the evaluation content our OR definition excludes, and the exploitation scale measures the action stage the same definition excludes, so movement on both reads as coupling with adjacent constructs, not as convergent validity. The off-construct profiles move little. The Big Five change modestly. Openness is the largest mover (0.583, up from 0.517), which we report descriptively and do not interpret as a change in the model's personality, since the openness items

share content with the opportunity-relevant items the intervention targets. Risk propensity, fear of failure, and ADHD each shift by 0.019 or less.

The internal structure of self-efficacy sharpens this reading. The scale's facets span the venture process from noticing to execution (McGee et al., 2009). Under OR-up steering the largest increase appears on the recognition-facing Searching facet (0.767, up 0.140) and the smallest on the execution-facing Implementing-financial facet (0.178, up 0.035). This is a descriptive within-model gradient, not a facet-level test, but it concentrates exactly where an intervention on recognition, not execution, should (McMullen and Shepherd, 2006).

The four role prompts produce a larger and broader response. The entrepreneur, high-OR, and low-OR personas move the entrepreneurial scales far more than steering does (Table G.3), and the negated persona mirrors the entrepreneur persona downward (opportunity recognition 0.296, self-efficacy 0.149, exploitation 0.211, from baselines of 0.652, 0.339, and 0.609). One off-target effect qualifies all four personas. The ADHD self-report rises under every persona, and the negated persona raises it more than the entrepreneur persona (from 0.066 at baseline to 0.315 under the entrepreneur persona, 0.359 under the negated persona, 0.353 under the high-OR persona, and 0.280 under the low-OR persona), a pattern that role polarity does not explain. OR steering, by contrast, leaves ADHD nearly unchanged (0.057 under OR-up, 0.052 under OR-down).

The three primary contrasts are distinguishable from the baseline when each scored scale is treated as a paired observation (13 scales, the 10 reported instruments plus entrepreneurial passion, individual entrepreneurial orientation, and effectuation from the same battery): OR-up raises the mean (Holm-corrected $p=0.001$), OR-down lowers it ($p<0.001$), and the entrepreneur persona raises it ($p<0.001$). We treat these as descriptive because the instruments are theoretically curated and correlated. The $\alpha=7.0$ columns of Table G.3 are an uncalibrated over-gate exhibit (Section 3.2.1).

Across both readouts, the persona is the larger and broader lever on the explicit questions and on the survey scales, whereas steering moves less off target and reverses in sign. We do not claim that steering reads opportunity more objectively than prompting, and the difference-in-differences contrasts in Appendix G.1.1 do not show steering to be more selective than the alertness instruction. That instruction opens a wider target-versus-control margin only because it pushes the control tiers sharply downward, and its own effect on the oblique questions is smaller than the vector's, 0.190 against 0.429. The comparison supports a narrower conclusion. The OR dial is a distinct intervention on opportunity-recognition content. It moves the off-target tiers less than the instruction does in absolute size and in the opposite direction, and one signed coefficient under a measured coherence gate reverses the opportunity-recognition readout while leaving the off-construct scales near their unsteered level. None of the role prompts does both.

**Table G.2. Survey battery scales, response formats, and sources.**

| Scale | Items | Subscales (items) | Response format | Source |
|---|---|---|---|---|
| Opportunity recognition | 5 | - | 7-point agreement | Kuckertz et al. (2017) |
| Opportunity exploitation | 4 | - | 7-point agreement | Kuckertz et al. (2017) |
| Entrepreneurial alertness | 13 | Scanning (6), association (3), evaluation (4) | 5-point agreement | Tang et al. (2012) |
| Entrepreneurial self-efficacy | 19 | Searching (3), planning (4), marshalling (3), implementing/people (6), implementing/financial (3) | 5-point confidence | McGee et al. (2009) |
| Proactive personality | 17 | - | 7-point agreement | Bateman and Crant (1993) |
| Entrepreneurial bricolage | 9 | - | 5-point frequency | Davidsson et al. (2017) |
| Big Five personality traits (BFI-44) | 44 | Extraversion (8), agreeableness (9), conscientiousness (9), neuroticism (8), openness (10) | 5-point agreement | John and Srivastava (1999) |
| Risk propensity (RPS-7) | 7 | - | 9-point agreement | Meertens and Lion (2008) |
| Entrepreneurial fear of failure | 18 | Funding (3), idea potential (3), social esteem (3), opportunity cost (3), personal ability (3), financial security (3) | 5-point agreement | Cacciotti et al. (2020) |
| ADHD self-report (ASRS-18, v1.1) | 18 | Inattention (9), hyperactivity/impulsivity (9) | 5-point frequency | Kessler et al. (2005) |

*Notes*: We present each item to the model in its native response format, including the item statement and numbered response scale (for example, 1=strongly disagree to 5=strongly agree), and read the rating as the probability-weighted average of the scale points at the answer position. Scores are normalised to a 0–1 range before aggregation. Every scale is scored and retained in full, negatively worded items are reverse-keyed, and the item wording follows the published instruments cited in the final column.

**Table G.3. Survey readout under OR steering, benchmarked against entrepreneur and OR-disposition role prompts.**

| Scale (items) | Baseline | Steering (Δ score from baseline) | | | | Persona (Δ score from baseline) | | |
|---|---|---|---|---|---|---|---|---|
| | | OR-up (α=3.5) | OR-down (α=3.5) | OR-up (α=7.0) | OR-down (α=7.0) | Entrepre-neur | High OR | Low OR |
| Opportunity recognition (5) | 0.652 | 0.066 | -0.027 | 0.260 | -0.228 | 0.161 | 0.303 | -0.505 |
| Opportunity exploitation (4) | 0.609 | 0.035 | -0.023 | 0.218 | -0.160 | 0.193 | 0.283 | -0.446 |
| Entrepreneurial alertness (13) | 0.582 | 0.055 | -0.017 | 0.269 | -0.101 | 0.123 | 0.261 | -0.336 |
| Entrepreneurial self-efficacy (19) | 0.339 | 0.092 | -0.020 | 0.492 | -0.089 | 0.495 | 0.582 | -0.251 |
| Proactive personality (17) | 0.614 | 0.059 | -0.017 | 0.218 | -0.196 | 0.174 | 0.259 | -0.355 |
| Entrepreneurial bricolage (9) | 0.479 | 0.073 | -0.008 | 0.274 | -0.079 | 0.249 | 0.319 | -0.253 |
| Big Five: Openness (10) | 0.517 | 0.066 | -0.040 | 0.248 | -0.083 | 0.165 | 0.287 | -0.226 |
| Big Five: Conscientiousness (9) | 0.562 | 0.009 | -0.009 | 0.076 | -0.028 | 0.217 | 0.020 | -0.162 |
| Big Five: Extraversion (8) | 0.487 | 0.013 | -0.019 | 0.092 | -0.022 | 0.217 | 0.161 | -0.112 |
| Big Five: Agreeableness (9) | 0.612 | -0.005 | -0.017 | 0.029 | -0.071 | 0.124 | 0.002 | -0.110 |
| Big Five: Neuroticism (8) | 0.440 | 0.023 | -0.002 | 0.036 | -0.006 | -0.085 | -0.051 | 0.097 |
| Risk propensity (7) | 0.344 | 0.019 | -0.001 | 0.088 | 0.054 | 0.252 | 0.302 | -0.011 |
| Fear of failure (18) | 0.511 | 0.001 | -0.002 | 0.212 | -0.061 | -0.070 | 0.046 | -0.193 |
| ADHD self-report (18) | 0.066 | -0.009 | -0.014 | 0.020 | -0.044 | 0.249 | 0.287 | 0.214 |

*Notes*: Values are Llama 3.1 8B-Instruct's survey responses, scored on a normalised 0-to-1 range and averaged across items within each scale (battery in Table G.2). The baseline column reports the mean score of the unsteered model. All remaining columns report changes from this baseline when the OR direction is added or subtracted at layer 12, or when the model is prompted with the entrepreneur, high-OR, and low-OR role prompts. The α=7.0 columns are descriptive double-coefficient diagnostics that exceed the 0.100 coherence gate by construction (Section 3.2.1). They are not interpreted as calibrated readouts. Indented rows are the Big Five personality traits subscales. Paired t-tests with Holm correction cover the OR-up, OR-down, and entrepreneur-persona contrasts only (Appendix G.1).

## G.2 Sparse autoencoder (SAE) decomposition of the OR direction

We examine how the OR direction relates to features in a sparse autoencoder (SAE). An SAE re-expresses activations at a layer as a sparse combination of features drawn from a learned dictionary that contains more candidate features than the activation space has dimensions. Individual features carry tentative natural-language labels (e.g., Bricken et al., 2023; Cunningham et al., 2023). This analysis provides an illustrative feature-level view of the OR direction, not a robustness test or evidence of feature-level mediation.

We compare the OR direction with the decoder vectors from Llama Scope (He et al., 2024), a publicly released set of sparse autoencoders trained on the base Llama 3.1 8B model. We rank the 32,768 decoder vectors of the layer-12 residual-stream SAE by cosine similarity with the unit-normalised OR direction and report the most aligned features at each pole (cf.

Chalnev et al., 2024). The ranking uses the DiM OR direction read at the content anchor, defined as the mean over the scenario's own tokens, in place of the Part 1 read position at -5. We use the content anchor because it is in distribution for the autoencoder, whereas the selected -5 anchor can fall on an outlier-norm special token of the kind base-trained autoencoders reconstruct poorly (Kissane et al., 2024).

The alignments are weak. The top decoder cosine with the OR direction is 0.341, compared with a random-direction reference of approximately 0.066, and the top feature's label has no direct relation to OR. We therefore read the ranking as geometric overlap, not as evidence that any single feature uniquely encodes OR (Table G.4). Among the ten most aligned features on the OR-present pole, three carry labels related to markets, expansion, and commerce. The battery bans construct and role labels (Appendix A.2) but not commercial vocabulary in general, so this alignment may reflect the commercial framing the OR-present pole carries as much as OR content. This pattern is compatible with an opportunity-related semantic component, but does not establish one. Consistent with this caution, only one of the ten top-aligned OR-present features separates OR-present from OR-absent scenarios above chance on the held-out pairs, and only weakly (feature-level AUROC 0.500 for nine of the ten and 0.571 for the remaining one), so the label pattern reflects geometric alignment of decoder directions, not an identified OR feature. The top-ranked features are not simply splits of a single feature, as the maximum pairwise absolute decoder cosine is 0.306. The weak and distributed alignments are consistent with the OR contrast being represented across a higher-dimensional subspace, not localised in a single feature (Elhage et al., 2022).

The SAE reconstructs the Llama 3.1 8B-Instruct activations only partially. At the content position used for the ranking, the reconstruction residual is 0.646 of the activation norm, so the reconstruction accounts for 0.582 of the squared activation norm. The cosine ranking itself is computed against the fixed decoder directions and therefore does not depend on reconstructing the full activation accurately. However, the partial reconstruction limits the interpretation of feature labels and alignments. The feature labels in Table G.4 are automatically generated Neuronpedia descriptions (https://www.neuronpedia.org) for the same layer-12 SAE. Because Llama Scope is trained on base-model activations and applied here to instruction-tuned ones, we treat this decomposition as illustrative only (Kissane et al., 2024).

**Table G.4. SAE features most aligned with each pole of the OR direction.**

| Rank | Similarity | OR-present SAE feature | Similarity | OR-absent SAE feature |
|---|---|---|---|---|
| 1 | 0.341 | Technical specifications and data-structure or programming concepts | -0.287 | Health conditions and epidemiological data |
| 2 | 0.282 | Communication and community engagement | -0.265 | Technical XML schema information |
| 3 | 0.259 | Urban development and infrastructure projects | -0.164 | Personal conflicts and interpersonal relationships |
| 4 | 0.236 | Strategic planning or mechanisms | -0.154 | Service staff and their interactions |
| 5 | 0.232 | Environmental sustainability and construction | -0.149 | Appreciation and gratitude |
| 6 | 0.228 | Structured writing or articles | -0.147 | Emotional connection and interpersonal relationships |
| 7 | 0.226 | **Market potential and growth opportunities** | -0.146 | Cleaning and maintenance services |
| 8 | 0.222 | **Expansion and growth in business contexts** | -0.146 | Numerical data relating to finances |
| 9 | 0.206 | **Commercial production and industrial applications** | -0.144 | Communication and expression of thoughts and feelings |
| 10 | 0.195 | Competitive events or awards | -0.144 | Service issues and problem resolution |

*Notes*: Similarity is the cosine similarity between the layer-12 OR direction and each SAE feature's decoder vector. The table shows the ten most aligned of the 20 ranked features per pole. Alignments are geometric, and feature identity is not verified on our scenarios. Positive values indicate alignment with the OR-present pole, negative values with the OR-absent pole. Labels are auto-generated Neuronpedia descriptions, lightly trimmed for readability. Bold marks the three OR-present features whose labels mention markets, expansion, or commerce.

**References (Appendix)**

Alain, G., & Bengio, Y. (2017). Understanding intermediate layers using linear classifier probes. *International Conference on Learning Representations (Workshop Track)*.

Anthropic. (2026b). *Claude Opus 4.8 system card*. https://www.anthropic.com/claude-opus-4-8-system-card

Arditi, A., Obeso, O., Syed, A., Paleka, D., Panickssery, N., Gurnee, W., & Nanda, N. (2024). Refusal in language models is mediated by a single direction. *Advances in Neural Information Processing Systems, 37*.

Bateman, T. S., & Crant, J. M. (1993). The proactive component of organizational behavior: A measure and correlates. *Journal of Organizational Behavior, 14*(2), 103–118.

Belinkov, Y. (2022). Probing classifiers: Promises, shortcomings, and advances. *Computational Linguistics, 48*(1), 207–219.

Belrose, N., Schneider-Joseph, D., Ravfogel, S., Cotterell, R., Raff, E., & Biderman, S. (2023). LEACE: Perfect linear concept erasure in closed form. *Advances in Neural Information Processing Systems, 36*.

Bricken, T., Templeton, A., Batson, J., Chen, B., Jermyn, A., Conerly, T., Turner, N., Anil, C., Denison, C., Askell, A., Lasenby, R., Wu, Y., Kravec, S., Schiefer, N., Maxwell, T., Joseph, N., Hatfield-Dodds, Z., Tamkin, A., Nguyen, K., … Olah, C. (2023). *Towards monosemanticity: Decomposing language models with dictionary learning*. Transformer Circuits Thread. https://transformer-circuits.pub/2023/monosemantic-features/index.html

Cacciotti, G., Hayton, J. C., Mitchell, J. R., & Allen, D. G. (2020). Entrepreneurial fear of failure: Scale development and validation. *Journal of Business Venturing, 35*(5), Article 106041.

Campbell, D. T., & Fiske, D. W. (1959). Convergent and discriminant validation by the multitrait-multimethod matrix. *Psychological Bulletin, 56*(2), 81–105.

Chalnev, S., Siu, M., & Conmy, A. (2024). *Improving steering vectors by targeting sparse autoencoder features*. arXiv. https://arxiv.org/abs/2411.02193

Cunningham, H., Ewart, A., Riggs, L., Huben, R., & Sharkey, L. (2023). *Sparse autoencoders find highly interpretable features in language models*. arXiv. https://arxiv.org/abs/2309.08600

Davidsson, P. (2015). Entrepreneurial opportunities and the entrepreneurship nexus: A re-conceptualization. *Journal of Business Venturing, 30*(5), 674–695.

Davidsson, P., Baker, T., & Senyard, J. M. (2017). A measure of entrepreneurial bricolage behavior. *International Journal of Entrepreneurial Behavior & Research, 23*(1), 114–135.

DeepSeek-AI. (2026). *DeepSeek-V4: Towards highly efficient million-token context intelligence*. arXiv. https://arxiv.org/abs/2606.19348

Durmus, E., Tamkin, A., Clark, J., Wei, J., Marcus, J., Batson, J., Handa, K., Lovitt, L., Tong, M., McCain, M., Rausch, O., Huang, S., Bowman, S., Ritchie, S., Henighan, T., & Ganguli, D. (2024). *Evaluating feature steering: A case study in mitigating social biases*. Anthropic. https://www.anthropic.com/research/evaluating-feature-steering

Elazar, Y., Ravfogel, S., Jacovi, A., & Goldberg, Y. (2021). Amnesic probing: Behavioral explanation with amnesic counterfactuals. *Transactions of the Association for Computational Linguistics, 9*, 160–175.

Elhage, N., Hume, T., Olsson, C., Schiefer, N., Henighan, T., Kravec, S., Hatfield-Dodds, Z., Lasenby, R., Drain, D., Chen, C., Grosse, R., McCandlish, S., Kaplan, J., Amodei, D., Wattenberg, M., & Olah, C. (2022). *Toy models of superposition*. Transformer Circuits Thread. https://transformer-circuits.pub/2022/toy_model/index.html

Geiger, A., Lu, H., Icard, T., & Potts, C. (2021). Causal abstractions of neural networks. *Advances in Neural Information Processing Systems, 34*.

Geirhos, R., Jacobsen, J.-H., Michaelis, C., Zemel, R., Brendel, W., Bethge, M., & Wichmann, F. A. (2020). Shortcut learning in deep neural networks. *Nature Machine Intelligence, 2*(11), 665–673.

Gemma Team. (2025). *Gemma 3 technical report*. arXiv. https://arxiv.org/abs/2503.19786

Grattafiori, A., Dubey, A., Jauhri, A., Pandey, A., Kadian, A., Al-Dahle, A., Letman, A., Mathur, A., Schelten, A., Vaughan, A., Yang, A., Fan, A., Goyal, A., Hartshorn, A., Yang, A., Mitra, A., Sravankumar, A., Korenev, A., Hinsvark, A., … Ma, Z. (2024). *The Llama 3 herd of models*. arXiv. https://arxiv.org/abs/2407.21783

Gururangan, S., Swayamdipta, S., Levy, O., Schwartz, R., Bowman, S. R., & Smith, N. A. (2018). Annotation artifacts in natural language inference data. *Proceedings of the 2018 Conference of the North American Chapter of the Association for Computational Linguistics: Human Language Technologies, Volume 2 (Short Papers)*, 107–112.

He, Z., Shu, W., Ge, X., Chen, L., Wang, J., Zhou, Y., Liu, F., Guo, Q., Huang, X., Wu, Z., Jiang, Y.-G., & Qiu, X. (2024). *Llama Scope: Extracting millions of features from Llama-3.1-8B with sparse autoencoders*. arXiv. https://arxiv.org/abs/2410.20526

Hewitt, J., & Liang, P. (2019). Designing and interpreting probes with control tasks. *Proceedings of the 2019 Conference on Empirical Methods in Natural Language Processing and the 9th International Joint Conference on Natural Language Processing (EMNLP-IJCNLP)*, 2733–2743.

Holm, S. (1979). A simple sequentially rejective multiple test procedure. *Scandinavian Journal of Statistics, 6*(2), 65–70.

John, O. P., & Srivastava, S. (1999). The Big Five trait taxonomy: History, measurement, and theoretical perspectives. In L. A. Pervin & O. P. John (Eds.), *Handbook of personality: Theory and research* (2nd ed., pp. 102–138). Guilford Press.

Kessler, R. C., Adler, L., Ames, M., Demler, O., Faraone, S., Hiripi, E., Howes, M. J., Jin, R., Secnik, K., Spencer, T., Ustun, T. B., & Walters, E. E. (2005). The World Health Organization Adult ADHD Self-Report Scale (ASRS): A short screening scale for use in the general population. *Psychological Medicine, 35*(2), 245–256.

Kissane, C., Krzyzanowski, R., Conmy, A., & Nanda, N. (2024). *SAEs (usually) transfer between base and chat models*. AI Alignment Forum. https://www.alignmentforum.org/posts/fmwk6qxrpW8d4jvbd/saes-usually-transfer-between-base-and-chat-models

Kuckertz, A., Kollmann, T., Krell, P., & Stöckmann, C. (2017). Understanding, differentiating, and measuring opportunity recognition and opportunity exploitation. *International Journal of Entrepreneurial Behavior & Research, 23*(1), 78–97.

Le Bras, R., Swayamdipta, S., Bhagavatula, C., Zellers, R., Peters, M. E., Sabharwal, A., & Choi, Y. (2020). Adversarial filters of dataset biases. *Proceedings of the 37th International Conference on Machine Learning, 119*, 1078–1088.

Li, K., Patel, O., Viégas, F., Pfister, H., & Wattenberg, M. (2023). Inference-time intervention: Eliciting truthful answers from a language model. *Advances in Neural Information Processing Systems, 36*.

Marks, S., & Tegmark, M. (2024). The geometry of truth: Emergent linear structure in large language model representations of true/false datasets. *Conference on Language Modeling*.

McGee, J. E., Peterson, M., Mueller, S. L., & Sequeira, J. M. (2009). Entrepreneurial self-efficacy: Refining the measure. *Entrepreneurship Theory and Practice, 33*(4), 965–988.

McMullen, J. S., & Shepherd, D. A. (2006). Entrepreneurial action and the role of uncertainty in the theory of the entrepreneur. *Academy of Management Review, 31*(1), 132–152.

Meertens, R. M., & Lion, R. (2008). Measuring an individual's tendency to take risks: The risk propensity scale. *Journal of Applied Social Psychology, 38*(6), 1506–1520.

Nanda, N., & Bloom, J. (2022). *TransformerLens* [Computer software]. GitHub. https://github.com/TransformerLensOrg/TransformerLens

Obschonka, M., & Fisch, C. (2025). From humans to machines: Researching entrepreneurial AI agents built on large language models. *Journal of Business Venturing Insights, 24*, Article e00581.

Park, K., Choe, Y. J., & Veitch, V. (2024). The linear representation hypothesis and the geometry of large language models. *Proceedings of the 41st International Conference on Machine Learning, 235*, 39643–39666.

Qwen Team. (2024). *Qwen2.5 technical report*. arXiv. https://arxiv.org/abs/2412.15115

Qwen Team. (2025). *Qwen3 technical report*. arXiv. https://arxiv.org/abs/2505.09388

Rimsky, N., Gabrieli, N., Schulz, J., Tong, M., Hubinger, E., & Turner, A. M. (2024). Steering Llama 2 via contrastive activation addition. *Proceedings of the 62nd Annual Meeting of the Association for Computational Linguistics (Volume 1: Long Papers)*, 15504–15522.

Shanahan, M., McDonell, K., & Reynolds, L. (2023). Role play with large language models. *Nature, 623*(7987), 493–498.

Shane, S., & Venkataraman, S. (2000). The promise of entrepreneurship as a field of research. *Academy of Management Review, 25*(1), 217–226.

Sharma, M., Tong, M., Korbak, T., Duvenaud, D., Askell, A., Bowman, S. R., Cheng, N., Durmus, E., Hatfield-Dodds, Z., Johnston, S. R., Kravec, S., Maxwell, T., McCandlish, S., Ndousse, K., Rausch, O., Schiefer, N., Yan, D., Zhang, M., & Perez, E. (2024). Towards understanding sycophancy in language models. *International Conference on Learning Representations*.

Suddaby, R. (2010). Editor's comments: Construct clarity in theories of management and organization. *Academy of Management Review, 35*(3), 346–357.

Tan, D., Chanin, D., Lynch, A., Paige, B., Kanoulas, D., Garriga-Alonso, A., & Kirk, R. (2024). Analysing the generalisation and reliability of steering vectors. *Advances in Neural Information Processing Systems, 37*.

Tang, J., Kacmar, K. M., & Busenitz, L. (2012). Entrepreneurial alertness in the pursuit of new opportunities. *Journal of Business Venturing, 27*(1), 77–94.

Tigges, C., Hollinsworth, O. J., Geiger, A., & Nanda, N. (2024). Language models linearly represent sentiment. *Proceedings of the 7th BlackboxNLP Workshop: Analyzing and Interpreting Neural Networks for NLP*, 58–87.

Turner, A. M., Thiergart, L., Leech, G., Udell, D., Vazquez, J. J., Mini, U., & MacDiarmid, M. (2023). *Steering language models with activation engineering*. arXiv. https://arxiv.org/abs/2308.10248

Wang, Y., Kordi, Y., Mishra, S., Liu, A., Smith, N. A., Khashabi, D., & Hajishirzi, H. (2023). Self-instruct: Aligning language models with self-generated instructions. *Proceedings of the 61st Annual Meeting of the Association for Computational Linguistics (Volume 1: Long Papers)*, 13484–13508.

Wehner, J., Abdelnabi, S., Tan, D., Krueger, D., & Fritz, M. (2025). Taxonomy, opportunities, and challenges of representation engineering for large language models. *Transactions on Machine Learning Research*.

Wu, Z., Arora, A., Geiger, A., Wang, Z., Huang, J., Jurafsky, D., Manning, C. D., & Potts, C. (2025). AxBench: Steering LLMs? Even simple baselines outperform sparse autoencoders. *Proceedings of the 42nd International Conference on Machine Learning, 267*, 67035–67080.

Zheng, L., Chiang, W.-L., Sheng, Y., Zhuang, S., Wu, Z., Zhuang, Y., Lin, Z., Li, Z., Li, D., Xing, E. P., Zhang, H., Gonzalez, J. E., & Stoica, I. (2023). Judging LLM-as-a-judge with MT-Bench and Chatbot Arena. *Advances in Neural Information Processing Systems, 36*.